\documentclass[pdflatex,sn-mathphys-num]{sn-jnl}
\usepackage{typearea}
\KOMAoptions{twoside=false}
\usepackage{graphicx}%
\usepackage{multirow}%
\usepackage{amsmath,amssymb,amsfonts}%
\usepackage{amsthm}%
\usepackage{mathrsfs}%
\usepackage[title]{appendix}%
\usepackage{textcomp}%
\usepackage{manyfoot}%
\usepackage{booktabs}%
\usepackage{algorithm}%
\usepackage{algorithmicx}%
\usepackage{algpseudocode}%
\usepackage{listings}%
\usepackage{hyperref} 
\usepackage{breakurl}
\usepackage{url} 
\usepackage{multirow}
\usepackage{subcaption}
\usepackage{tablefootnote}
\NewDocumentCommand{\malakia}{v}{#1}
\newcommand{\xazomara}[1]{#1}
\usepackage[table]{xcolor} 

\theoremstyle{thmstyleone}%
\theoremstyle{thmstyletwo}%

\theoremstyle{thmstylethree}%

\begin{document}

\title[A Review on Discriminative Self-supervised Learning Methods in Computer Vision]{A Review on Discriminative Self-supervised Learning Methods in Computer Vision}


\author*[1]{\fnm{Nikolaos} \sur{Giakoumoglou}}\email{nikos@imperial.ac.uk}
\author[1]{\fnm{Tania} \sur{Stathaki}}\email{t.stathaki@imperial.ac.uk}
\author[1]{\fnm{Athanasios} \sur{Gkelias}}\email{a.gkelias@imperial.ac.uk}
\affil*[1]{\orgdiv{Department of Electrical and Electronic Engineering}, \orgname{Imperial College London}, \orgaddress{\city{London}, \country{UK}}}


\abstract{Self-supervised learning (SSL) has rapidly emerged as a transformative approach in computer vision, enabling the extraction of rich feature representations from vast amounts of unlabeled data and reducing reliance on costly manual annotations. This review presents a comprehensive analysis of discriminative SSL methods, which focus on learning representations by solving pretext tasks that do not require human labels. The paper systematically categorizes discriminative SSL approaches into five main groups: contrastive methods, clustering methods, self-distillation methods, knowledge distillation methods, and feature decorrelation methods. For each category, the review details the underlying principles, architectural components, loss functions, and representative algorithms, highlighting their unique mechanisms and contributions to the field. Extensive comparative evaluations are provided, including linear and semi-supervised protocols on standard benchmarks such as ImageNet, as well as transfer learning performance across diverse downstream tasks. The review also discusses theoretical foundations, scalability, efficiency, and practical challenges, such as computational demands and accessibility. By synthesizing recent advancements and identifying key trends, open challenges, and future research directions, this work serves as a valuable resource for researchers and practitioners aiming to leverage discriminative SSL for robust and generalizable computer vision models.}




\maketitle

\section{Introduction}
\label{sec:introduction}

Over the past decade, artificial intelligence (AI) has made tremendous strides with the advent of deep learning, primarily through the development of convolutional neural networks (CNNs) \cite{krizhevsky2012alexnet} and, more recently, transformers \cite{dosovitskiy2021vit,touvron2021deit}. These architectures have become the dominant approaches for a large variety of tasks in computer vision, such as image classification, object detection, and image segmentation, and natural language processing (NLP), including sentence classification, language modeling, and machine translation. By using hierarchical feature extraction to efficiently process images, CNNs have long been the cornerstone of computer vision. However, the recent introduction of Vision Transformers (ViTs) \cite{dosovitskiy2021vit,touvron2021deit}, by directly applying transformer architectures \cite{vaswani2023attention} --originally conceived for NLP-- to image analysis, has led to substantial advancements in the field. ViTs allow the modeling of global dependencies and contextual information within images in a much more flexible and adaptive way for visual understanding. These deep learning models have been successful because they can discern complex patterns from the large datasets available today (i.e., CNNs excel at hierarchical feature learning, while ViTs use self-attention mechanisms to process whole images holistically), leading to state-of-the-art performance across various AI tasks.

The progress of AI systems has been fueled by large scale datasets that are carefully annotated by humans. While supervised learning has established itself as a robust method for training models on specific tasks, its application faces significant limitations, especially as we confront an era of data abundance. The rapid expansion of available data has outpaced our capacity to label it comprehensively. Deep learning models are particularly data-hungry, requiring vast amounts of labeled data to perform well \cite{alzubaidi2023survey}. Not exploring such large amounts of unlabeled data would not be only inefficient but also a waste of potential insights that could be gained from them. However, obtaining enough labeled data to train specialized models across various tasks can often be challenging and resource-intensive. In response to these challenges, the focus has shifted toward unsupervised learning approaches, particularly representation learning \cite{bengio2014representationlearningreviewnew}. This shift is driven by the need to use the rich, unlabeled datasets by learning useful representations directly from the data itself. Representation learning aims to discover and learn the inherent structures and patterns within the data, often complex and subtle, without the need for explicit labeling. This approach is crucial in making AI models more adaptable and efficient, capable of handling a broader range of tasks with less reliance on expensive and labor-intensive data labeling processes.

Self-supervised learning (SSL) has rapidly emerged as a significant approach in AI, often referred to as \textit{"the dark matter of intelligence"} \cite{balestriero2023cookbook}. This method diverges from traditional supervised learning, which relies on labeled datasets \cite{ohri2021review}, by utilizing the abundant unlabeled data that are often underutilized, allowing machines to achieve a deeper understanding of the world without being limited by labeled datasets. SSL distinguishes itself from unsupervised learning by the way it leverages data. While both utilize unlabeled data to learn underlying representations, SSL is explicitly trained with supervisory signals that are generated from the data itself (self-supervision). This is achieved by leveraging the inherent structure of the data through various so-called pretext tasks \cite{balestriero2023cookbook}. These tasks, such as predicting the next frame in a video sequence \cite{sermanet2018time} or determining the jumbled order of a set of image patches \cite{noroozi2017unsupervised}, generate pseudo-labels which the model uses to learn features. This represents a fundamental shift towards learning versatile representations that can be applied across multiple tasks.

The concept of SSL originated in NLP, where it achieved significant success before being applied to computer vision. Models such as BERT \cite{lan2020albert,devlin2018bert} and GPT \cite{brown2020language} were among the first to show that self-supervised models could outperform supervised models, reaching top results in their field. These models are trained on large volumes of unlabeled text, showcasing the power of SSL to utilize unstructured data \cite{chen2017reading,hamilton2017advances}. Nevertheless, the success of these models can also be attributed to the distinct nature of language data, which is discrete and structured, compared to the continuous and unstructured nature of image data, making it a unique challenge for computer vision \cite{he2020moco}.

In computer vision, SSL has emerged as a method to extract robust features from unlabeled data using properties from the images themselves. It has overcome previous limitations regarding data size and quality, and models such as SEER \cite{goyal2021seer,goyal2022seer}, trained on over of a billion images, illustrate the scale at which SSL can operate, pushing the boundaries of data handling and processing in unprecedented ways. The ability of SSL to match --and in some cases surpass-- the performance of models trained on meticulously labeled data is a testament to its potential. Especially in competitive benchmarks like ImageNet ILSVRC-2012 \cite{deng2009imagenet, russakovsky2015imagenet}, SSL methods have demonstrated their prowess, challenging the traditional paradigms of deep learning. The wide applicability and effectiveness of SSL across various modalities underscore its significance as a cornerstone in the future development of AI technologies.

SSL has also been applied successfully in other areas such as video, audio, and time series analysis \cite{oord2019cpc,zhang2024selfsupervised,liu2022audio,sermanet2018time, schiappa2023self}. These applications demonstrate its versatility in harnessing the intrinsic patterns of diverse data types to improve model performance and understanding.

SSL is generally categorized into two distinct approaches: {discriminative} and {generative}. The generative approach focuses on creating models capable of generating or reconstructing data to learn the overall data distribution. This method often employs models such as autoencoders (AEs) \cite{kingma2013autoencoding,vincent2008extracting} and generative adversarial networks (GANs) \cite{goodfellow2014generative}, which attempt to either reproduce the original data or generate new samples that are statistically similar to the input data. Despite its potential, generative SSL is noted for its high computational demands, particularly with high-resolution images, and questions remain about its utility and efficiency in learning meaningful representations \cite{chen2020simclr,grill2020byol}. Conversely, discriminative SSL aims to develop good data representations by engaging in specific pretext tasks that don't require manual labels. These tasks are designed to enable the model to identify and differentiate between various features of the data, helping to enhance its capability to generalize these learned patterns to new, unseen data. This strategy evaluates the quality of the learned representations through a scoring function, akin to the methods used in traditional supervised learning \cite{doersch2015unsupervised}.

Previous surveys on SSL have covered a wide range of aspects within this evolving field. However, many of these reviews tend to focus on particular areas of SSL or its specific applications within distinct domains \cite{rishnan2022self,wang2022self,shurrab2022self, chowdhury2021applying}. One review highlights various SSL methodologies specifically in computer vision \cite{ozbulak2023know}, while another offers an overview focusing on pretext and downstream tasks along with some applications of SSL \cite{rani2023self}. Another resource provides a comprehensive guide for training SSL methods, though it not formally a review article \cite{balestriero2023cookbook}. Additional reviews discuss SSL pretext tasks and architectures, offering a high-level overview of SSL applications in computer vision \cite{jaiswal2020survey,jing2019selfsupervised}. Furthermore, a review that focuses on contrastive SSL methods also discusses the broad applications of SSL in both NLP and computer vision \cite{kumar2022contrastive}.

Several surveys have laid the groundwork for understanding SSL. One of the earliest and most comprehensive \cite{jing2019selfsupervised} provided an excellent initial overview of SSL techniques, covering image, video, and 3D vision. While a valuable initial resource, its coverage extends up to early 2020, preceding significant advancements in the field. Another important early work \cite{jaiswal2020survey} focused specifically on contrastive SSL methods for both vision and natural language processing, offering a clear explanation of the contrastive learning pipeline. While these surveys were foundational, the rapid advancements in SSL, particularly after 2020, mean that their coverage of state-of-the-art techniques is now limited.

More recent surveys have attempted to capture the progress in the field. \cite{kumar2022contrastive} offered a comprehensive review of contrastive learning methodologies but had a broader scope beyond computer vision, also including NLP, speech and text recognition and prediction. \cite{khan2022survey} provided a shorter survey focusing on comparing a limited number of only six discriminative SSL methods. \cite{liu2023survey} surveyed SSL methods for various domains, including computer vision, NLP, and graph learning, categorizing them by their objectives, i.e., generative, contrastive and adversarial. Similarly, \cite{rani2023self} provided a general overview of the SSL pipeline and its applications across different domains, with less focus on detailed method analysis. \cite{hu2024survey} offered an in-depth look at the principles and components of contrastive learning but did not thoroughly analyze individual methods. \cite{ozbulak2023know} provided a historical view of image-oriented SSL, covering both generative and discriminative approaches. Finally, \cite{balestriero2023cookbook} presented a "{cookbook}" of SSL, outlining fundamental techniques with a unified vocabulary and highlighting key concepts like loss terms and training objectives. It further introduced practical implementation considerations, discussed common training recipes and evaluation methods, and shared practical insights from leading researchers. However, this paper serves more as a tutorial on the foundations and practical aspects of SSL rather than a comprehensive survey of methods. 

In contrast to these broader surveys, several works have focused on specific application domains of SSL. \cite{chowdhury2021applying} and \cite{rishnan2022self} specifically addressed the application of SSL in medicine and healthcare. \cite{shurrab2022self} concentrated on the use of SSL in medical imaging analysis.  \cite{wang2022self} explored SSL in remote sensing, considering different modalities. While these surveys provide valuable insights into domain-specific applications, our survey distinguishes itself by focusing specifically on discriminative SSL within the domain of computer vision. Furthermore, our work provides a more granular categorization of discriminative SSL methods, including {contrastive, clustering, self-distillation, knowledge distillation,} and {feature decorrelation} techniques. Significantly, our survey covers a substantially larger number of discriminative SSL methods and encompasses works up to early 2025, offering a more up-to-date and focused perspective on this rapidly evolving area within computer vision.

The principal \textbf{contributions} of this review paper are summarized as follows:

\begin{itemize}
\item We provide a comprehensive introduction to the discriminative SSL pipeline, detailing its key components including pretext tasks, architectural choices, and loss objectives. This foundational material serves as essential background for understanding the subsequent in-depth review.
\item We present a comprehensive and up-to-date review of discriminative SSL methods in computer vision, covering over 90 papers published from 2017 to 2025. We systematically categorize these methods into {five} primary groups: {contrastive, clustering, self-distillation, knowledge distillation}, and {feature decorrelation}, offering a thorough comparison of their pipeline and architectural designs.
\item We detail the most commonly used evaluation protocols and datasets in discriminative SSL. We compile, analyze, and summarize a wide range of performance results reported across numerous methods using these standard protocols.
\item We trace the architectural evolution of discriminative SSL methods, examining shifts in base networks, the progression of pretext tasks, and the development of training objectives and loss functions. We discuss how each of these advancements has contributed to the field's progress.
\item We analyze the primary challenges faced by discriminative SSL, including an in-depth exploration of the inherent trade-offs within these methods.
\item We discuss critical considerations regarding the evaluation and practical application of discriminative SSL techniques.
\end{itemize}

The rest of the review paper is structured as follows. Section~\ref{sec:background} provides a background overview of SSL, introducing the pipeline, pretext tasks, downstream tasks, architectural components, and loss objectives essential for understanding the field. Section~\ref{sec:selfsupervisedmethods} comprehensively examines the five categories of discriminative SSL methods: {contrastive, clustering, self-distillation, knowledge distillation}, and {feature decorrelation} approaches. For each category, we analyze key representative methods, their architectural designs, and their unique contributions to the field. Section~\ref{sec:benchmarking} presents a thorough evaluation framework, comparing method performance across standard benchmarks and transfer learning tasks. Section~\ref{sec:discussion} offers critical insights into the evolution of architectural frameworks, fundamental challenges, inherent trade-offs, evaluation considerations, and theoretical foundations of discriminative SSL. Finally, Section~\ref{sec:conclusion} concludes the review by summarizing the key findings and suggesting promising directions for future research in this rapidly advancing field.

\section{Background}\label{sec:background}

\begin{figure*}[!t]
\centering
\includegraphics[width=0.9\textwidth]{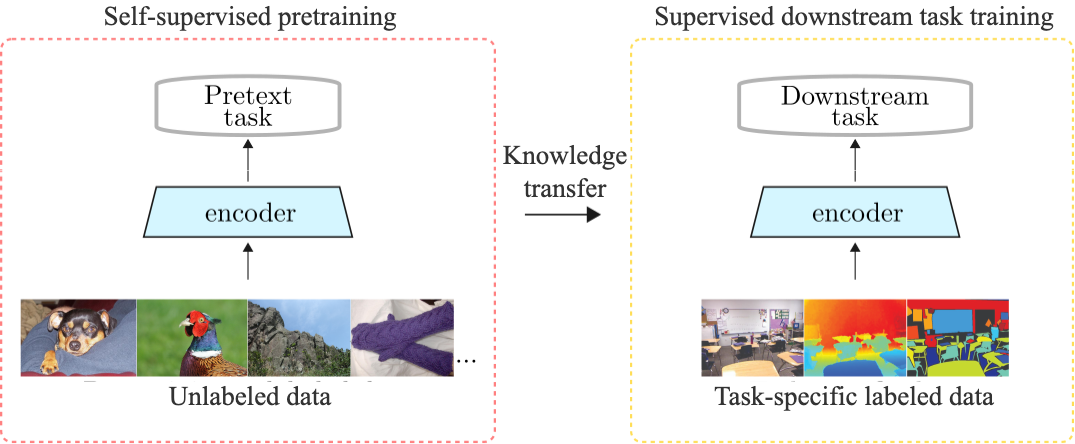}
\caption{Typical SSL pipeline. Once the training on the pretext task is complete, the learned parameters are adapted as a pre-trained model for various downstream computer vision tasks through fine-tuning. Image adjusted from \cite{jing2019selfsupervised}.}
\label{fig:fig1}
\end{figure*}

The SSL pipeline typically consists of two main stages, the self-supervised pretraining and the supervised downstream task training stage, respectively, as depicted in Figure \ref{fig:fig1}. In the pretraining stage, an unlabeled dataset is used to train a neural network (also referred to as the encoder, detailed in Section~\ref{sec:architecturalcomponents}) on predefined pretext tasks (Section~\ref{sec:pretexttasks}), where pseudo labels are generated automatically from certain data attributes. These tasks are designed to encourage the encoder to learn useful representations from the data itself by developing filters that capture both basic and intricate features, which are beneficial for subsequent downstream tasks (Section \ref{sec:downstreamtasks}), without the need for explicit labels. More specifically, the earlier layers of the encoder focus on general low-level features like corners, edges, and textures, while the deeper layers target more complex, high-level features such as objects, scenes, and parts of objects. The learning process is guided by carefully designed loss objectives (Section~\ref{sec:lossobjectives}) that optimize feature discrimination, while data augmentation techniques (Section~\ref{sec:dataaugmentation}) generate diverse training views, enhancing robustness. Once the self-supervised training is complete, the learned visual features can be utilized as a pre-trained model for downstream tasks (Section \ref{sec:downstreamtasks}), typically retaining only the initial layers of the encoder for fine-tuning with limited labeled data.

\subsection{Pretext Tasks}
\label{sec:pretexttasks}

A pretext task in SSL for computer vision is a carefully designed task that allows a model to learn useful visual representations from unlabeled data. These tasks leverage inherent properties of images to create pseudo-labels, eliminating the need for manual annotation. Early pretext tasks have included methods such as colorization (i.e., enhance the ability to understand textures, patterns, and object identities by colorizing grayscale images) \cite{zhang2016colorization}, inpainting (i.e., use context prediction to fill in missing parts of an image) \cite{pathak2016context}, jigsaw puzzles (i.e., learn to solve jigsaw puzzles, using the relative positions of image patches as supervision) \cite{noroozi2017unsupervised, noroozi2018boosting}, geometric transformations (i.e., enhance feature learning for various orientations by predicting the rotation angle applied to an image) \cite{gidaris2018rotnet}, and others.

More recent approaches have focused on contrastive learning via instance discrimination, a technique that trains models to distinguish between similar and dissimilar instances without using labeled data. Instance discrimination operates by maximising the similarity of positive pairs (typically augmented versions of the same image) while minimizing the similarity of negative pairs (different images). Data augmentation, a key component of contrastive learning, involves applying various transformations to images, such as random cropping, resizing, colour distortion, or Gaussian blurring, to create different views of the same instance. In contrastive learning, data augmentation serves a dual purpose: it generates the positive pairs needed for the learning process and challenges the model to learn invariant representations across different transformations.

Data augmentation differs from conventional pretext tasks, which are specific, often handcrafted objectives designed to guide the model in learning useful representations by solving a predefined problem. While conventional pretext tasks focus on specific challenges, contrastive learning with data augmentation concentrates on recognizing different views of the same image, potentially leading to more general and transferable representations. For instance, as a pretext task, colorization involves training a model to predict colors for grayscale images as a learning objective. Besides, in data augmentation, color-related augmentations (e.g., color jittering and random grayscale conversions) are often used to create different views of the same image, serving the learning objective of contrastive learning: to distinguish between these augmented versions of the same image and different images.

Recently, Masked Image Modeling (MIM) was proposed as an effective pretext task for SSL in computer vision. MIM originated from NLP and gained prominence in computer vision with works like BEiT \cite{bao2022beit}. This task involves masking portions of an input image and training models to reconstruct the missing content, challenging them to understand visual context and structure. While effective across various architectures, including CNNs, MIM has shown particular success with ViTs. Unlike traditional pretext tasks, MIM focuses on reconstructing missing information rather than solving specific transformations. However, as a primarily generative method, MIM falls outside the scope of this review, which focuses on discriminative SSL techniques in computer vision.

Among the aforementioned pretext tasks, instance discrimination has become the dominant pretext task for discriminative SSL in computer vision and is widely adopted by most SSL methods. On the other hand, MIM is gaining ground as a powerful pretext task that enables generative SSL to match or even surpass instance discrimination.

\subsection{Downstream Tasks}
\label{sec:downstreamtasks}

Downstream or target tasks are the primary objective of SSL, designed to leverage feature representations acquired during pretext training to tackle more complex problems, such as classification, object detection, semantic segmentation, and depth estimation. They directly assess the model's ability to understand and effectively utilize the learned features, frequently with limited labeled data, by employing techniques like fine-tuning or linear classification. Performance on these downstream tasks (e.g., image classification or object detection), serves as the primary measure of SSL success, quantifying the transferability and robustness of the learned representations and demonstrating the advantage of achieving high performance even with minimal supervision.

\subsection{Architectural Components}
\label{sec:architecturalcomponents}

SSL uses different architectural components to effectively train models that can understand and represent complex visual data without labeled examples. The core of these architectures is the backbone (also referred to as the encoder), which acts as the primary feature extractor. This backbone transforms raw input images into a rich set of features. Initially, early SSL methods used CNNs like AlexNet \cite{krizhevsky2012alexnet} and VGG-16 \cite{simonyan2015vgg}, but more recent approaches often employ ResNets \cite{he2015resnet}, their variants \cite{zagoruyko2016wide}, or ViTs \cite{dosovitskiy2021vit,touvron2021deit, liu2021swin, elnouby2021xcit}. 

Attached to the backbone is usually a projection network, which enhances the quality of representations produced by the encoder. While the backbone/encoder learns general features, the projector helps in optimizing the SSL objective without directly affecting the learned representations. This network is usually a multi-layer perceptron (MLP) that maps the high-dimensional features from the backbone into a space where similar examples are clustered together, aiding the learning of the images' semantic content. The projection network typically includes one to three layers and often features batch normalization to help stabilize the learning process. Some SSL models also include an additional prediction network, another MLP, to introduce asymmetry into the SSL architecture \cite{grill2020byol}. When transferring to downstream tasks, typically only the backbone is retained, while the projector is removed.

\begin{figure*}[!t]
\centering
\includegraphics[width=0.9\textwidth]{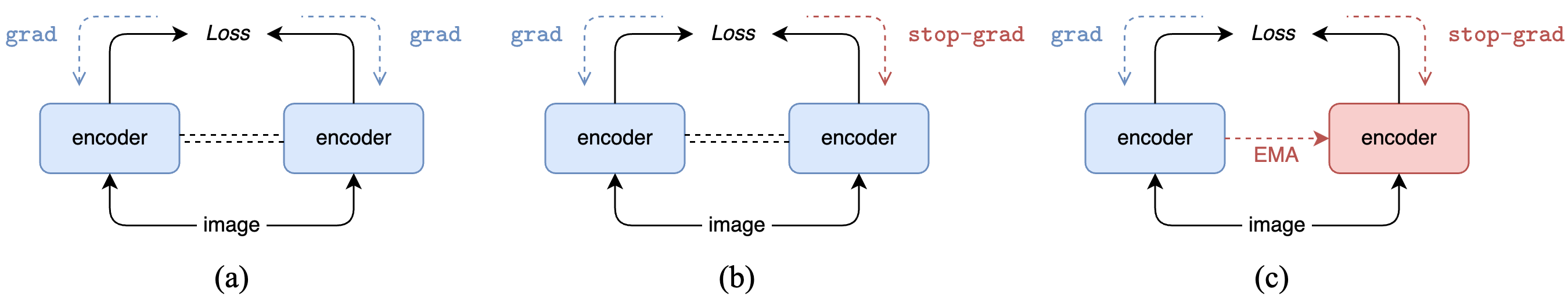}
\caption{Common architectural variations of Siamese networks used in self-supervised learning, each demonstrating different gradient flow mechanisms. (a) Siamese architecture with gradients (\texttt{grad}) flowing from both branches. (b) Siamese architecture where gradients flow from the first branch only. (c) Siamese architecture with gradient flow from the first branch and a stop-gradient (\texttt{stop-grad}) operation on the second branch, which is updated using an EMA.}
\label{fig:fig3}
\end{figure*}

Modern SSL methods frequently employ dual backbone architectures, such as Siamese architectures (see Figure~\ref{fig:fig3}a), which consist of two identical neural networks that share the same weights. The fundamental premise of a Siamese architecture is to process two separate inputs simultaneously and compare their feature representations. This setup ensures consistent representations across different augmentations of the same image, improving the model's ability to generalize from unlabeled data. 

In dual backbone architectures, the updating process can vary. Often, both backbones are updated through standard back-propagation. However, to prevent one branch of the Siamese network from affecting the learning of the other, a stop-gradient operation is sometimes employed \cite{chen2020simsiam} (see Figure~\ref{fig:fig3}b). This technique halts the gradient flow from one branch, effectively isolating the updates of each network and helping to stabilize the training process by reducing interference between the branches and preventing the model from collapsing into trivial solutions.

Furthermore, in some cases, a momentum-based update method such as an exponential moving average (EMA) is employed for one of the backbones to help stabilize learning updates \cite{he2020moco} (see Figure~\ref{fig:fig3}c). This approach can smooth out abrupt changes in model parameters over training iterations, potentially leading to faster convergence and better generalization, depending on the specific SSL method being used. 

All these architectural components will be revisited in the remaining of this review when the different SSL techniques will be presented in detail. 

\subsection{Loss Objectives in Self-supervised Learning}
\label{sec:lossobjectives}

In SSL, selecting appropriate loss objectives is crucial for effectively training models to process and interpret complex, unlabeled visual data. These loss functions are specifically designed to align with the goals of various SSL methods and have played a crucial role in the SSL evolution.

Cross entropy (CE) loss is commonly used in self-distillation and clustering methods to measure the difference between the predicted probabilities and the target distribution. It is particularly effective for models where outputs can be viewed as class probabilities. The formula for CE loss, when \( z_1 \) and \( z_2 \) are outputs from Siamese networks is given by:

\begin{equation}
\text{CE}(z_1, z_2) = - \sum_{i=1}^{N} z_{2,i} \cdot \log(z_{1,i})
\end{equation}

Mean squared error (MSE) loss and mean absolute error (MAE) loss are favored for regression tasks and methods that aim to closely match feature distributions. MSE calculates the average of the squares of the differences between predicted values and targets, ideal for embedding similarity, as shown by:

\begin{equation}
\text{MSE}(z_1, z_2) = \frac{1}{N} \sum_{i=1}^{N} (z_{11,i} - z_{21,i})^2
\end{equation}

MAE, offering robustness against outliers, computes the average magnitude of differences between pairs of observations:

\begin{equation}
\text{MAE}(z_1, z_2) = \frac{1}{N} \sum_{i=1}^{N} |z_{11,i} - z_{21,i}|
\end{equation}

Negative cosine similarity (NCS) loss is prevalent in self-distillation, focusing on the angle between predictions rather than their magnitude. For \( \ell_2 \)-normalized vectors \( z_1 \) and \( z_2 \), NCS is calculated as:

\begin{equation}
\text{NCS}(z_1, z_2) = -\frac{z_1 \cdot z_2}{\|z_1\| \cdot \|z_2\|}
\end{equation}

Kullback-Leibler (KL) divergence is utilized to measure the divergence of one probability distribution from another, which is crucial in predicting probability distributions. This measure is particularly useful in self-distillation methods that involve matching the output distributions of different network branches:

\begin{equation}
\text{KL}(P || Q) = \sum P \cdot \log \frac{P}{Q}
\end{equation}

where \( P \) and \( Q \) are the true and predicted distributions, respectively. In this context, $P=softmax\left(z_2\right)$ and $Q=softmax\left(z_1\right)$.

A common approach in contrastive frameworks to extract meaningful representations from data without explicit labels, involves contrasting representations from augmented versions of the same data instance. The objective is to minimize the distance between these embeddings in the feature space, thus ensuring that the learned representations are invariant to the applied augmentations while being discriminative enough to distinguish among different instances. 
The initial loss function in this paradigm, known as contrastive loss, was initially introduced in \cite{bromley1998signature, chopra2005learning}. It aims to reduce the distance between positive pairs (i.e., different augmentations of the same instance) while enforcing a margin between negative pairs (i.e., augmentations of different instances). 
Building upon the contrastive loss, the triplet loss \cite{weinberg2009distance, chechik2009learning} introduces a relative comparison between an anchor, a positive, and a negative sample, further refining the ability to capture differences between embeddings. 
To extend the comparison to multiple negative samples, the $(N+1)$-tuple loss \cite{sohn2016improved} contrasts one positive pair against several negatives within the same batch. 
The Noise-Contrastive Estimation (NCE) loss \cite{wu2018instdis} introduces a temperature parameter and employs explicit normalization to further refine the approximation of the probability distribution of the embeddings. 

The InfoNCE loss \cite{oord2019cpc}, a variant of NCE, is widely adopted in SSL, particularly for contrastive learning models that leverage a large number of negative samples. It operates by maximizing the dot product similarity between positive pairs while minimizing similarity with negative samples.

\begin{equation}\label{eq:infonce}
    \text{InfoNCE}(z_1, z_2) = -\log \frac{\exp(z_1 \cdot z_2 / \tau)}{\sum_{k \neq 1} \exp(z_1 \cdot z_k / \tau)}
\end{equation}

where $z_1$ and $z_2$ are augmented views of the same image, $z_k$ represents other images in the batch, and $\tau$ is the temperature parameter. Subsequent developments have expanded upon the concepts introduced by InfoNCE \cite{balestriero2023cookbook}. 
\cite{chen2020simclr} introduced the NT-Xent (Normalized Temperature-scaled Cross Entropy) loss, which diverges from the traditional InfoNCE loss by utilizing cosine similarity instead of the dot product:

\begin{equation}
\text{NT-Xent}(z_1, z_2) = -\log \frac{\exp(\text{sim}(z_1, z_2)/\tau)}{\sum_{k \neq 1} \exp(\text{sim}(z_1, z_k)/\tau)}
\end{equation}

where \( \text{sim}(\cdot, \cdot) \) denotes cosine similarity and \( \tau \) is a temperature parameter. As in InfoNCE, $z_k$ represents the set of available negative samples, which can be drawn from a queue of previous samples \cite{he2020moco} or from other samples within the current minibatch \cite{chen2020simclr}. These negative samples are crucial as they serve as a reference point to enhance the model's ability to recognize and differentiate between similar and different features by contrasting them against the positive pair.

Building upon the foundation laid by InfoNCE, subsequent research has explored various refinements \cite{balestriero2023cookbook}. The modification introduced in \cite{yeh2022dcl} excludes the positive pair from the InfoNCE denominator, thereby intensifying the differentiation between distinct instances. This adjustment sharpens the model's discriminative power, helping it to better isolate and identify unique features. Additionally, \cite{dwibedi2021nnclr} enhanced the contrastive learning process by employing a queue to store embeddings of nearest neighbors. Moreover, \cite{mitrovic2020relic} incorporated a regularization term to enforce invariance in the learned representations, thus adding an additional layer of constraint to the optimization problem. Finally, \cite{li2021pcl} utilized prototypes, i.e., representative embeddings of clusters, to guide the learning process and capture the underlying data structure.

\subsection{Data Augmentation}
\label{sec:dataaugmentation}

Data augmentation plays a crucial role in SSL by generating diverse variations of input data, thereby enhancing model robustness and generalization. By applying various transformations to the original data while preserving semantic content, augmentation encourages the learning algorithm to identify consistent underlying patterns and instills representational consistency. This is essential for developing robust features applicable to downstream tasks. In SSL, data augmentation is instrumental in enriching the learning experience and improving the model's ability to generalize from its self-generated training signal.

Contrastive learning methods employing instance discrimination as the pretext task rely heavily on data augmentation to create distinct "views" of the same image. This approach is critical for training the model to recognize objects across different transformations. For example, by generating one augmented version of an image with color distortions and another with cropping or texture changes, the model learns to associate these disparate views as representing the same underlying object. The training objective is to bring the feature representations of these augmented views closer together in embedding space, while simultaneously distinguishing them from representations of different objects. This process significantly improves the model's ability to recognize objects under varying conditions, a key factor for strong performance on practical downstream tasks.

\begin{figure*}[!t]
\centering
\includegraphics[width=0.9\textwidth]{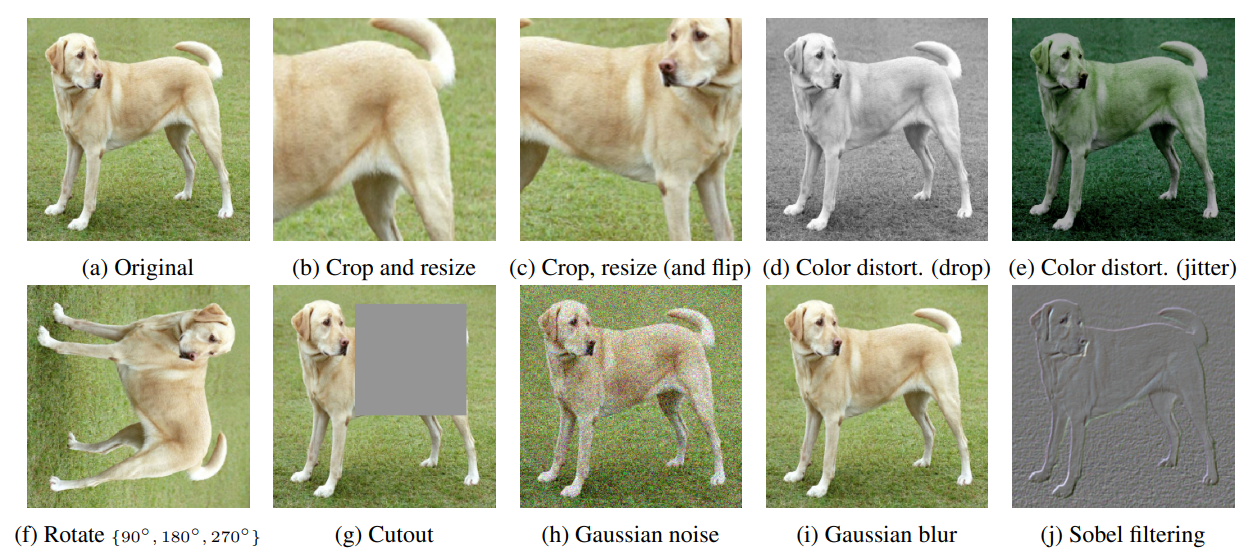}
\caption{Various data augmentation techniques applied to an original image of a yellow Labrador: (a) Original unmodified image; (b) Random crop and resize operation that focuses on a portion of the dog; (c) Similar crop and resize with possible horizontal flip; (d) Color distortion through grayscale conversion; (e) Color distortion through color jittering that alters the image's color properties; (f) Rotation applied at various angles (90°, 180°, 270°); (g) Cutout augmentation that masks a random square region of the image; (h) Gaussian noise injection that adds random pixel-level perturbations; (i) Gaussian blur that reduces image details while preserving structure; and (j) Sobel filtering that emphasizes edges and contours. These diverse transformations help self-supervised models learn invariant representations. Image from \cite{chen2020simclr}.}
\label{fig:fig2}
\end{figure*}

Key augmentation techniques include random resized cropping, random horizontal flipping, color jittering, random grayscale conversion, Gaussian blur, solarization, Sobel filtering, and multi-crop augmentation \cite{caron2020swav}. All these image manipulations aim to enhance the model's robustness to variations in orientation, color, and detail. The intensity and probability of each augmentation are controlled by specific parameters, offering a wide range of training variations. Figure~\ref{fig:fig2} illustrates examples of these techniques. To optimize augmentation strategies, frameworks like AutoAugment \cite{cubuk2019autoaugment} and RandAugment \cite{cubuk2019randaugment} have been developed. AutoAugment uses reinforcement learning to discover optimal augmentation policies for image classification, while RandAugment simplifies this process by controlling the magnitude of transformations with a single parameter, streamlining the identification of effective augmentations.

\section{Self-supervised Learning Methods}\label{sec:selfsupervisedmethods}

This review categorizes discriminative SSL methods into five distinct yet overlapping categories, following the framework established in~\cite{balestriero2023cookbook}. This classification provides a systematic basis for discussing how these techniques leverage unlabeled data to learn effective representations. While alternative classifications exist, we adopt this one to emphasize key attributes and mechanisms within each category.

{Contrastive methods} include techniques that learn representations by contrasting positive pairs against negative pairs. The key feature here is the use of a contrastive loss (e.g., InfoNCE or NT-Xent), which encourages similar samples (positives) to cluster closer in the embedding space while pushing dissimilar samples (negatives) further apart. These methods often rely on data augmentation to generate positive pairs and random sampling for negative pairs. The primary rationale for this categorization is the method’s reliance on explicit pairing and negative sampling to drive the learning process, focusing on maximizing discriminative power between different instances. Contrastive learning methods will be discussed in detail in Section~\ref{sec:contrastivemethods}.

\malakia|SimCLR|~\cite{chen2020simclr} and \malakia|MoCo|~\cite{he2020moco} are arguably two of the most seminal models in contrastive SSL for computer vision. \malakia|SimCLR|'s simplicity and strong performance, achieved through careful data augmentation and end-to-end training, established a powerful baseline. \malakia|MoCo|'s innovative use of a momentum-updated encoder and a dynamic queue of keys (negative samples) addressed the challenge of scaling contrastive learning, significantly improving performance and influencing subsequent work. Both models, which will be further analyzed later, have significantly advanced the field, often outperforming supervised learning approaches and bridging the gap between self-supervised and supervised learning in various computer vision tasks.

A key variant of contrastive learning is dense contrastive learning, which aims to learn dense feature representations essential for localization-based tasks like object detection and segmentation. These methods adapt the contrastive learning framework to address the unique challenges of these tasks. We dedicate a separate discussion to dense contrastive learning methods because their evaluation often differs from standard classification benchmarks, i.e., their objectives often lead to a trade-off that may reduce classification performance in favor of significant gains in object detection and segmentation. This distinct focus necessitates a dedicated analysis to highlight the unique challenges and solutions it presents.

{Clustering methods} utilise clustering algorithms to iteratively group and assign pseudo-labels to data points during training to supervise the learning process. This approach leverages the unsupervised clustering dynamics by promoting intra-cluster similarity and inter-cluster diversity to guide the representation learning without the need for negative samples. For example, \malakia|DeepCluster|~\cite{caron2019deepcluster} iteratively clusters features extracted from images and uses these cluster assignments to update the neural network, ensuring that the learned features are both diverse and informative. \malakia|DeepCluster|'s specifics will be examined in Section~\ref{sec:clusteringmethods}.

{Self-distillation methods} involve training a single network to learn from its own predictions or representations, often at different layers or using different augmentations. Unlike contrastive methods, which rely on positive and negative sample pairs, self-distillation focuses on consistency between different views of the same input. This can be seen as a form of regularization, encouraging the network to learn more robust and generalizable features by distilling knowledge from its own outputs. This approach is especially beneficial for smaller models and can capture higher-level semantic information without explicit contrastive objectives. For instance, the network architecture of \malakia|BYOL| (Bootstrap Your Own Latent) \cite{grill2020byol}, maximizes consistency between representations of two augmented views, focusing entirely on positive pair reinforcement and using techniques such as updating one encoder's weights with an EMA of the others to prevent feature collapse or using an asymmetrical architecture, introducing a predictor on the non-momentum branch. The detailed workings and effectiveness of \malakia|BYOL| will be explored in Section~\ref{sec:selfdistillationmethods}.

{Knowledge distillation methods} commonly utilize a teacher-student architecture, wherein a larger and more powerful pre-trained network (i.e., the teacher) distills its knowledge to a smaller network (i.e., the student). Knowledge distillation can be unsupervised, and when applied in an unsupervised manner, it can leverage SSL objectives to train the student network. This approach allows the student to learn from the richer representations captured by the teacher, often leading to improved performance, faster convergence, or the ability to train smaller, more efficient models. Essentially, the teacher guides the student, distilling its expertise and enhancing the student's ability to learn effective representations from unlabeled data. Knowledge distillation SSL methods will be discussed in detail in Section~\ref{sec:kdmethods}.

{Feature decorrelation methods} aim to learn representations where features are statistically independent or uncorrelated. This is often achieved by adding a decorrelation term to the loss function, encouraging the network to learn features that are diverse and non-redundant. By enforcing independence between features, they prevent dimensional collapse without relying on negative samples or asymmetric architectures. Unlike contrastive methods that focus on similarity and dissimilarity between different views, or self-distillation methods that emphasize internal consistency, feature decorrelation directly targets the statistical relationships. Feature decorrelation is exemplified by \malakia|Barlow Twins|~\cite{zbontar2021barlowtwins}, which enforces redundancy reduction across the features of its two neural networks by minimizing the deviation of the cross-correlation matrix from the identity matrix. This minimization enhances both discriminative ability and robustness. We analyse the implications and applications of \malakia|Barlow Twins| in Section~\ref{sec:featuredecorrelationmethods}.

\begin{figure*}[!t]
    \centering
    \includegraphics[width=1\textwidth]{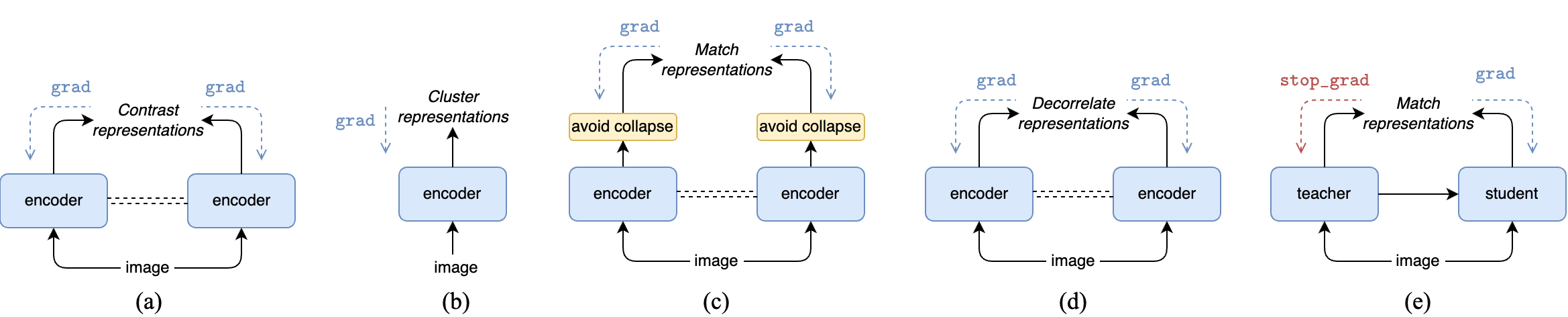}
    \caption{High level comparison of discriminative SSL methods. (a) Contrastive methods employ instance discrimination. (b) Clustering methods use clustering algorithms to provide supervision. (c) Self-distillation methods use various techniques to avoid collapse while matching representations of Siamese networks. (d) Feature decorrelation methods promote feature diversity. (e) Knowledge distillation methods transfer knowledge from a frozen teacher network to a student network.}
    \label{fig:fig4}
\end{figure*}

Figure~\ref{fig:fig4} provides a high-level overview, illustrating the key characteristics of the methods discussed above.

\subsection{Early Works and Enablers}
\label{sec:earlyworks}

This section discusses early representation learning methods which, while initially proposed for diverse applications, have since been adapted or categorized within SSL frameworks. These foundational works introduced various pretext tasks designed to enable representation learning by exploiting different aspects of visual data. Techniques like image colorization, spatial transformations, and context prediction leveraged the inherent structure and content of images to learn useful features without labeled data. These approaches established a versatile foundation for SSL, demonstrating how creatively formulated tasks can drive the learning of rich visual representations.

Among these early methods, \cite{zhang2016colorization} focused on grayscale image colorization by predicting color channels from the lightness channel, addressing the multimodal nature of colorization. Similarly, \cite{larsson2016learning} developed a system that predicts per-pixel color histograms, effectively combining low-level and semantic representations and demonstrating the utility of colorization in self-supervised visual representation learning.

\malakia|RotNet| \cite{gidaris2018rotnet} trained a model to predict the rotation applied to an image, leveraging spatial transformations to learn robust image representations without labeled data. This method has significantly influenced subsequent work, establishing a foundational technique in SSL that exploits geometric transformations as a supervisory signal \cite{caron2019deepercluster}. \malakia|SplitBrain| \cite{zhang2017splitbrain} utilized AEs to predict different color channels, focusing on cross-channel prediction within the network.

Context Encoders \cite{pathak2016context} employed a CNN trained to generate the contents of an arbitrary image region based on its surroundings, using a combination of reconstruction and adversarial loss. A related approach \cite{doersch2015unsupervised} trained a CNN to predict the relative position of image patches, again leveraging spatial context as a supervisory signal and requiring the recognition of objects and their parts.

Early work also explored training a CNN to solve jigsaw puzzles. \malakia|Jigsaw| \cite{noroozi2017unsupervised} used 
a Context-Free Network (CFN) to learn about object parts and their spatial relationships. Subsequent work, \malakia|Jigsaw++| \cite{noroozi2018boosting}, further improved performance by transferring knowledge, decoupling the self-supervised model's structure from the task-specific model, and leveraging pseudo-labels derived from clustered features of a pretext task.

Finally, adversarial learning saw significant progress with the introduction of Bidirectional Generative Adversarial Networks (BiGANs) \cite{donahue2016adversarial}, designed to learn the inverse mapping from data back to the latent space. While BiGANs are frequently employed in generative SSL frameworks, a discussion of these frameworks falls outside the scope of this paper due to their distinct generative objectives. Nevertheless, Limited Context Inpainting (LCI) \cite{jenni2020lci}, warrants mention. LCI aims to learn image representations by inpainting an image patch using a generator-discriminator architecture. Crucially, LCI preserves the patch boundary, maintaining local image statistics while altering global ones. The network is then trained to predict the type of transformation applied (rotation, LCI, or wrapping), as in \cite{gidaris2018rotnet}. These pioneering methods established a foundation for future SSL research and demonstrated the breadth of possible approaches and applications.


\subsection{Contrastive Methods}
\label{sec:contrastivemethods}

Contrastive learning methods within deep metric learning commonly employ instance discrimination as a pretext task \cite{balestriero2023cookbook}. These methods revolve around three core components: an "anchor" sample (the reference point), a "positive" sample (similar to the anchor), and one or more "negative" samples (dissimilar to the anchor) \cite{khosla2021supcon}. The objective is to minimize the distance between the anchor and positive sample embeddings while maximizing the distance between the anchor and negative sample embeddings. In the absence of labels or other annotations, positive samples are often generated through augmentations of the anchor, while negative samples are typically randomly selected images from the mini-batch. Essentially, positive samples are created by generating multiple views of each data point. This can be achieved through various techniques, such as decomposing an image into its luminance and chrominance components~\cite{tian2020cmc}, applying random augmentations to an image multiple times~\cite{chen2020simclr, he2020moco}, extracting patches from the same image~\cite{oord2019cpc}, or leveraging the outputs from student and teacher models~\cite{grill2020byol}. Negative pairs are typically formed from randomly chosen images. Theoretically, positive pairs can be considered as samples from a joint distribution over views, $p(x_{1}, x_{2})$, while negative pairs can be considered as samples from the product of marginals, $p(x_{1}) \cdot p(x_{2})$ \cite{tian2020infomin}. 

This section is organized into subsections based on the distinct characteristics and methodologies of each group of contrastive learning techniques. We begin by introducing foundational contrastive learning approaches (Section~\ref{sec:contrastive_foundational}). We then explore Siamese network methods, which employ two parallel (often weight-sharing) networks and focus on learning through direct comparisons of different augmentations of the same image (Section~\ref{sec:contrastive_siamese}). Next, we discuss momentum-based methods, which utilize a momentum encoder for gradual model updates, promoting stable learning on large, continually updated datasets (Section~\ref{sec:contrastive_momentum}). Following this, we examine neighbourhood-based methods, which leverage data space proximity, frequently using nearest neighbours to improve the quality of learned representations (Section~\ref{sec:contrastive_neighbor}). We then present several frameworks that discuss comprehensive systems and architectures for more systematic implementation of contrastive learning (Section~\ref{sec:contrastive_frameworks}). Subsequently, we cover adversarial methods, which incorporate adversarial training to refine the contrastive learning process, typically by generating challenging negative samples (Section~\ref{sec:contrastive_adversarial}). Finally, we address dense contrastive methods, which focus on acquiring dense features for localization-based tasks (Section~\ref{sec:contrastive_dense}).

\subsubsection{Foundational Approaches}
\label{sec:contrastive_foundational}

Early SSL methods, such as \malakia|InstDis|~\cite{wu2018instdis} and \malakia|PIRL|~\cite{misra2019pirl}, laid the groundwork for subsequent contrastive learning approaches. These initial methods demonstrated the potential of leveraging unlabeled data to learn robust representations by contrasting positive and negative sample pairs. In the following we summarize some of the most prominent foundational contrastive self-supervised methods.

Instance-level Discrimination (\malakia|InstDis|) \cite{wu2018instdis} pioneered instance discrimination in SSL by introducing the idea of treating each image as its own unique class. The method learns by drawing different augmented views of the same image closer together in the embedding space, while pushing them away from all other images. Effectively, this creates a classification problem with as many classes as there are images in the dataset. \malakia|InstDis| trained a deep neural network using a non-parametric softmax classifier to map each image to a unit vector on a hypersphere. To address the computational challenges of this large-scale classification task, \malakia|InstDis| implemented a memory bank to store feature representations of all images, facilitating efficient negative example sampling during training. This method, by emphasizing the contrast between augmented views of the same image and a large pool of negative samples, established instance discrimination as a crucial pretext task and significantly influenced the development of later contrastive learning methods.

Unsupervised Embedding Learning (\malakia|UEL|) \cite{ye2019uel} aims to learn effective representations from unlabeled images by focusing on two key properties, i.e., invariance to data augmentations and maximizing the spread between different instances. \malakia|UEL| employed a Siamese network with shared weights across two branches (each with a backbone and a projector) to process an original image and its augmented counterpart, along with a small batch of other images, generating $\ell_2$-normalized embeddings. Framing the problem as binary classification, \malakia|UEL| uses maximum likelihood estimation to identify the augmented sample as belonging to the original image, distinguishing it from other instances in the batch. The objective is to minimize the negative log-likelihoods across all instances. \malakia|UEL|'s key innovation is its instance-based softmax embedding approach, which directly optimizes instance features without relying on a memory bank. This emphasis on both invariance and spread-out influenced subsequent contrastive learning methods by demonstrating the importance of maintaining fine-grained visual consistency while effectively separating features from different instances. 

Pretext-Invariant Representation Learning (\malakia|PIRL|) \cite{misra2019pirl} learns image representations invariant to pretext transformations by encouraging similarity between representations of an original image and its transformed counterpart. It uses a single backbone network with two projection heads to learn representations invariant to image transformations, including the jigsaw pretext task. The method employs a memory bank to efficiently compare the current image with a large number of negative samples, allowing for effective contrastive learning without the need for large batch sizes \cite{wu2018instdis}. \malakia|PIRL|'s loss function incorporates two terms: the InfoNCE loss for positive pairs and its complement for differentiating positive from negative pairs. If the second term is disregarded, \malakia|PIRL| simplifies to an enhanced version of \malakia|InstDis|, referred to as \malakia|InstDis++|. By encouraging representations that are invariant under both data augmentations and pretext transformations, \malakia|PIRL| advanced contrastive learning by focusing on semantic consistency rather than task-specific features. This shift toward transformation-invariant representations directly influenced later contrastive frameworks, which also prioritize invariance properties. 

Deep InfoMax (\malakia|DIM|) \cite{hjelm2019dim} and Augmented Multiscale Deep InfoMax (\malakia|AMDIM|) \cite{bachman2019amdim} learn representations by maximizing mutual information (MI) between input data and encoder outputs. \malakia|DIM| uses a contrastive objective to align global features (high-level summaries) with local features (intermediate activations), employing adversarial learning to match encoder outputs to a prior distribution. \malakia|AMDIM| extends \malakia|DIM| by maximizing MI across independently augmented views and multiple scales of the encoder's hierarchy, capturing richer cross-view and cross-scale information. \malakia|DIM|'s key contribution was establishing MI maximization as a core SSL objective, formalizing contrastive learning as an information-theoretic problem and pioneering local-global hierarchical learning. \malakia|AMDIM| further advanced this by introducing multiview and multiscale contrast, enhancing robustness and capturing hierarchical semantic structures. These methods significantly influenced subsequent contrastive learning by emphasizing MI maximization and demonstrating the power of multiview and multiscale approaches, inspiring methods like \malakia|CPC| and \malakia|CMC| and providing a theoretical framework for understanding "good" representations.

Contrastive Predictive Coding (\malakia|CPC|) \cite{oord2019cpc} learns representations by predicting future encodings in latent space. It uses an autoregressive model to generate predictions of future representations based on a context vector, which is derived from past observations. The learning objective is to maximize the MI between the predicted future representations and the actual future observations, achieved through a contrastive loss (InfoNCE) that distinguishes true future samples from negative samples. \malakia|CPC-v2| \cite{henaff2020cpcv2} enhances the original \malakia|CPC| framework by using a larger architecture, replacing batch normalization with layer normalization, and introducing improved data augmentation techniques. It also predicts patches from multiple directions, leading to more robust representations and improved performance on downstream tasks.
By introducing the InfoNCE loss, \malakia|CPC| formalized contrastive learning as a probabilistic framework for maximizing MI. This loss function has proven to be highly effective in enhancing the learning of useful representations by maximizing the MI between related pairs of samples while minimizing it between unrelated ones. This approach influenced nearly all subsequent contrastive methods by highlighting the importance of temporal or sequential structure in representation learning.

Contrastive Multiview Coding (\malakia|CMC|) \cite{tian2020cmc} explicitly addresses the multiview nature of data by leveraging different sensory modality or spatial views (e.g., luminance (\textit{L}) and chrominance (\textit{ab} colour channels)) to learn view-invariant representations. \malakia|CMC|  trains a separate encoder for each view, maximizing mutual information between representations of different views of the same instance while minimizing it for different instances (as in \malakia|InstDis| \cite{wu2018instdis}), effectively learning view-invariant features that capture shared information across multiple perspectives. This approach emphasizes the complementary nature of information from different views, a concept central to how contrastive learning uses augmentations as distinct "views" of the same data.

These foundational approaches, by exploring various contrastive strategies, laid the groundwork for modern SSL. Their focus on instance discrimination, mutual information maximization, and invariance to transformations paved the way for more sophisticated contrastive learning frameworks.

\subsubsection{Siamese Networks Methods}
\label{sec:contrastive_siamese}

Siamese networks, a neural network architecture using two or more identical subnetworks with shared weights to process different inputs and compare their outputs, have been introduced in early 1990s to solve signature verification as an image matching problem \cite{bromley1998signature}.

However, it was the Simple framework for Contrastive Learning of visual Representations (\malakia|SimCLR|) \cite{chen2020simclr} that first adapted this architecture for self-supervised contrastive learning in computer vision, marking a significant milestone in the field. 
\malakia|SimCLR| introduced a remarkably simple yet highly effective framework that processes two augmented views of an image through identical networks, each consisting of a backbone and a projector, as shown in Figure \ref{fig:simclr}. The method generates these views through data augmentation and processes them through weight-sharing encoders to produce $\ell_2$-normalized embeddings. \malakia|SimCLR|'s learning objective focuses on maximizing the similarity between these embeddings using the NT-Xent loss, utilizing other batch samples as negative examples. \malakia|SimCLR| demonstrated that a simple contrastive approach, leveraging a Siamese architecture with shared weights to generate contrastive pairs through appropriate data augmentation and large batch sizes, could outperform more complex methods. This work paved the way for subsequent SSL techniques like \malakia|BYOL|~\cite{grill2020byol} and \malakia|SimSiam| \cite{chen2020simsiam} (formally introduced in Section \ref{sec:selfdistillationmethods}), which further built upon this Siamese network foundation.

\malakia|SimCLR-v2| \cite{chen2020simclrv2} built upon the original \malakia|SimCLR| by incorporating larger ResNet models, a deeper projection head, and a memory bank from \cite{he2020moco}, along with integrating self-supervised pre-training with supervised fine-tuning.

\malakia|G-SimCLR| \cite{chakraborty2020gsimclr} offers another modification to the original \malakia|SimCLR| framework. It proposed a novel batch preparation strategy, beginning with training a denoising autoencoder \cite{kumar2014static} to obtain robust image representations. These representations are then clustered to assign pseudo-labels, and training batches are subsequently constructed to ensure that no pseudo-label is repeated within a batch.

Inspired by \malakia|SimCLR|, Simple Contrastive Learning (\malakia|SimCL|) \cite{yang2022simcl} abandons data augmentation to create different views, and adopts adding uniform noise in the representation space. \malakia|C-SimCLR| \cite{lee2021cbyolcsimclr} modifies \malakia|SimCLR| by adding information compression using conditional entropy bottleneck. Supervised Contrastive (\malakia|SupCon|) \cite{khosla2021supcon} extends the \malakia|SimCLR| framework to a fully supervised setting, efficiently leveraging label information. It introduces the concept of multiple positives per anchor, along with a broad set of negatives.

Inspired by \malakia|SimCLR|, \malakia|DirectCLR| \cite{jing2022directclr} builds on the contrastive learning framework but distinguishes itself by eliminating the trainable projector typically employed in \malakia|SimCLR| to transform representations before applying the InfoNCE loss. Instead, \malakia|DirectCLR| directly optimizes a subvector of the representation space using the backbone encoder, retaining \malakia|SimCLR|'s use of data augmentation to create different views. This approach prevents dimensional collapse—where embedding vectors span a lower-dimensional subspace—without relying on an additional projection layer.

\subsubsection{Momentum-based Methods}
\label{sec:contrastive_momentum}

Many of the aforementioned SSL methods face several critical challenges in representation learning. These include training instability caused by rapidly changing representations in methods like \malakia|SimCLR|, scalability limitations due to memory-intensive negative sampling requirements, and the risk of representation collapse where models output identical features for all inputs. Momentum techniques have emerged as a powerful solution to these issues. 

\begin{figure*}[!t]
\centering
\begin{subfigure}[b]{0.195\textwidth}
    \centering
    \includegraphics[width=\linewidth,keepaspectratio]{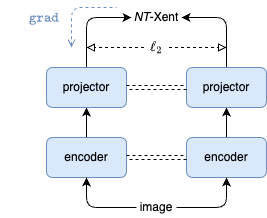}
    \caption{\xazomara{SimCLR}}
    \label{fig:simclr}
\end{subfigure}%
\begin{subfigure}[b]{0.25\textwidth}
    \centering
    \includegraphics[width=\linewidth,keepaspectratio]{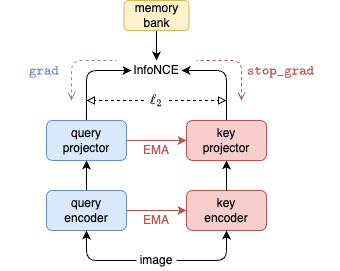}
    \caption{\xazomara{MoCo}}
    \label{fig:moco}
\end{subfigure}
\quad
\begin{subfigure}[b]{0.25\textwidth}
    \centering
    \includegraphics[width=\linewidth,keepaspectratio]{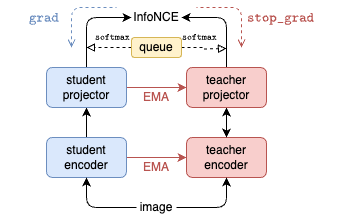}
    \caption{\xazomara{ReSSL}}
    \label{fig:ressl}
\end{subfigure}%
\begin{subfigure}[b]{0.25\textwidth}
    \centering
    \includegraphics[width=\linewidth,keepaspectratio]{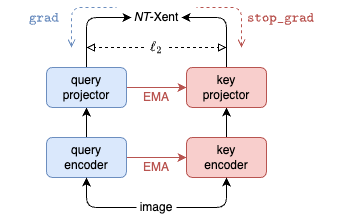}
    \caption{\xazomara{MoCLR}}
    \label{fig:moclr}
\end{subfigure}
\caption{Overview of contrastive learning frameworks. (a) \xazomara{SimCLR} uses simple data augmentations to generate positive pairs. (b) \xazomara{MoCo} leverages a dynamic memory bank for negative samples. (c) \xazomara{ReSSL} uses relation to look at relationships between instances. (d) \xazomara{MoCLR} improves upon the \xazomara{SimCLR}.}
\label{fig:contrastive-comparison}
\end{figure*}

Momentum contrast (\malakia|MoCo|) \cite{he2020moco} was the first contrastive method to address these key challenges by introducing EMA to stabilize encoder updates, facilitating gradual weight adjustments that prevent abrupt shifts in representations. \malakia|MoCo| employs two encoders (as shown in Figure \ref{fig:moco}): a query encoder (with parameters updated directly during training) consisting of a backbone and a projector, and a key encoder (with parameters updated using an EMA of the query network's parameters) with a similar architecture. This EMA update ensures that the keys in the dictionary are generated by a slowly evolving encoder, preventing the rapidly changing representations seen in other contrastive methods and maintaining representational consistency over time. Furthermore, \malakia|MoCo| utilizes a queue-based mechanism, rather than the memory bank used in \malakia|InstDis|, to maintain a large and consistent dictionary of negative samples. This approach decouples the dictionary size from the mini-batch size, enabling the model to contrast against a significantly larger set of negative examples without increasing memory requirements. During training, the current mini-batch is encoded by the query encoder, while the keys in the queue are encoded by the momentum-updated key encoder, yielding $\ell _2$-normalized embeddings. The contrastive loss (InfoNCE) is then computed between the query and the keys, where the positive key is the encoded augmented view of the same image, and the negative keys are drawn from the queue. Following each iteration, the oldest keys in the queue are replaced with the newly encoded keys from the current batch. This approach allows \malakia|MoCo| to efficiently leverage a large number of negative samples, enhancing the quality of learned representations while preserving computational efficiency.

\malakia|MoCo-v2| \cite{chen2020mocov2} builds upon the original \malakia|MoCo| framework, incorporating several design improvements inspired by \malakia|SimCLR| \cite{chen2020simclr}. While retaining \malakia|MoCo|'s momentum-updated key encoder for stable representation learning and its dynamic queue to decouple batch size from the number of negatives (addressing \malakia|SimCLR|'s reliance on large batch sizes), \malakia|MoCo-v2| introduces several key changes. It replaces the single fully connected layer in the original \malakia|MoCo| with a 2-layer MLP projection head for more complex feature transformations. \malakia|MoCo-v2| also adopts \malakia|SimCLR|'s emphasis on stronger data augmentations, including blur and color distortion, to generate more diverse views of the same image and learn more robust representations. Furthermore, like \malakia|SimCLR|, \malakia|MoCo-v2| uses $\ell _2$-normalized embeddings and temperature scaling ($\tau$) in its InfoNCE loss, focusing on directional similarity and stabilizing training for improved representation quality. These modifications have enabled \malakia|MoCo-v2| to establish stronger baselines, outperforming \malakia|SimCLR| while using smaller batch sizes and fewer training epochs.

\malakia|MoCo-v3| \cite{chen2021mocov3} builds upon its predecessors (\malakia|MoCo| and \malakia|MoCo-v2|) by introducing key architectural and methodological refinements tailored for ViTs and improving training stability. It maintains the Siamese network approach of the previous versions but simplifies the architecture by removing the memory queue, instead relying on in-batch negatives from larger batch sizes, and implements a symmetrized InfoNCE loss applied bidirectionally between query-key pairs. Moreover, it enhances stability, particularly for ViT training, by freezing the initial patch embedding layer or using fixed random projections to prevent gradient spikes, and adds an MLP prediction head to the query encoder for improved feature separation. Finally, it optimizes for ViTs by adopting the AdamW \cite{loshchilov2019adamw} optimizer, which is better suited for ViTs and enables stable large-batch training, along with a long warmup period of 40 epochs to mitigate early training instability. However, for ResNets it employs the LARS \cite{you2017lars} optimizer. By addressing ViT-specific optimization challenges and refining contrastive learning mechanics, \malakia|MoCo-v3| became a pivotal step toward modern SSL frameworks for vision. Table \ref{tab:moco_comparison} summarizes the main differences between \malakia|MoCo|, \malakia|MoCo-v2| and \malakia|MoCo-v3| methods. The rest of this subsection presents several methods designed to address various limitations and performance issues associated with the \malakia|MoCo| framework.

Mixing of Contrastive Hard NegatIves (\malakia|MoCHI|) \cite{kalantidis2020mochi} addresses performance issues of \malakia|MoCo| by focusing on the challenge of "hard" negatives in contrastive learning. \malakia|MoCHI| observes that as training progresses in \malakia|MoCo|, fewer negatives contribute significantly to the loss, indicating a need for harder negatives. To address this, it introduces a feature-level mixing strategy for hard negative samples, synthesizing new, more challenging negatives on-the-fly, leading to improved generalization and more efficient utilization of the embedding space. \malakia|MoCHI| demonstrates faster learning of transferable representations, showing higher performance gains over \malakia|MoCo-v2| for transfer learning after only 100 epochs of training. However, it's worth noting that \malakia|MoCHI|  does not show performance gains over \malakia|MoCo-v2| for linear classification, suggesting its benefits may be task-specific or more pronounced in certain scenarios.

Building upon the observations of \malakia|MoCHI| and within the \malakia|MoCo| framework, \malakia|SynCo| \cite{giakoumoglou2024synco} represents a significant advancement in contrastive learning through synthetic hard negative generation. While \malakia|MoCHI| introduced the concept of mixing hard negatives, \malakia|SynCo| substantially expands this approach by proposing four additional strategies beyond \malakia|MoCHI|'s initial two methods. These strategies—extrapolated negatives, noise-injected negatives, perturbed negatives, and adversarial negatives—collectively create a more diverse and challenging set of synthetic negatives that target different aspects of the representation space. Unlike \malakia|MoCHI|, which showed limited gains in linear classification tasks, \malakia|SynCo| demonstrates consistent improvements across both linear evaluation and transfer learning benchmarks. This comprehensive approach to synthetic hard negative generation allows \malakia|SynCo| to modulate proxy task difficulty more effectively, facilitating faster training convergence and producing representations with better uniformity properties. 

\malakia|SimMoCo| (Simple \malakia|MoCo|) and \malakia|SimCo| (\malakia|SimMoCo| with no MOmentum) \cite{zhang2022simco} simplify the \malakia|MoCo| framework by removing the dictionary and momentum encoder, respectively, and integrating an InfoNCE loss with a dual temperature (DT) mechanism. This DT mechanism modulates the scaling of positive and negative sample distances within the loss function, enabling a more refined contrastive learning process where the separation between similar and dissimilar pairs can be finely tuned for enhanced discriminative feature learning. Their results demonstrate that neither the dictionary nor the momentum encoder is essential for training, proving that a simplified contrastive learning framework can still achieve robust performance by effectively leveraging the InfoNCE loss with this novel DT approach.

Information Minimization (\malakia|InfoMin|) \cite{tian2020infomin} advances the \malakia|MoCo| framework by introducing a novel approach to data augmentation and contrastive learning. Building upon \malakia|MoCo|, \malakia|InfoMin| incorporates a specialized augmentation strategy called {InfoMin Aug}, which includes jigsaw augmentation alongside other techniques. The key innovation of \malakia|InfoMin| is its focus on minimizing MI between different views of the same image while maximizing the MI between representations and the underlying semantic content. In addition to jigsaw augmentation, {InfoMin Aug.} includes random resized crop, random horizontal flip, colour jitter, Gaussian Blur, RandAugment \cite{cubuk2019randaugment}, and random grayscale. This comprehensive augmentation approach aims to create more challenging and informative positive pairs, encouraging the model to learn more robust and transferable representations.

Leave-one-out Contrastive (\malakia|LooC|) \cite{xiao2021looc} builds upon the \malakia|MoCo|  framework, introducing a novel approach to multi-augmentation contrastive learning. While \malakia|MoCo| uses a queue of encoded samples as negative examples, \malakia|LooC| selectively leverages multiple augmentations of the same image. \malakia|LooC|'s key innovation is treating one augmentation as the anchor while using the others as positive samples, effectively leaving out one augmentation at a time to create the anchor. Furthermore, it optimizes multiple augmentation-sensitive contrastive objectives using a multi-head architecture with a shared backbone, experimenting with rotation, color jittering, and texture randomization as augmentations. Unlike \malakia|MoCo|, which relies solely on different images for negative samples, \malakia|LooC| learns from subtle differences within augmentations of the same image, potentially improving its ability to distinguish fine-grained features.

Transformation Invariance and Covariance Contrast (\malakia|TiCo|) \cite{zhu2022tico} advances \malakia|MoCo| by combining transformation invariance with covariance regularization. \malakia|TiCo| maximizes agreement among embeddings of different distorted versions of the same image, promoting transformation-invariant representations. To prevent trivial solutions, it regularizes the covariance matrix of embeddings from different images by penalizing low-rank solutions, effectively minimizing the distance between various augmentations of the same image and encouraging representations towards smaller eigenvalues of the covariance matrix, thus enhancing covariance contrast. Notably, \malakia|TiCo| can be viewed as a variant of \malakia|MoCo| with an implicit memory bank of unlimited size, but without additional memory cost, enabling it to outperform alternative methods with small batch sizes. By bridging contrastive and redundancy-reduction methods, \malakia|TiCo| provides new insights into joint embedding approaches and offers improved performance in computationally limited scenarios.

Relational Self-Supervised Learning (\malakia|ReSSL|) \cite{zheng2021ressl} advances \malakia|MoCo| by introducing a novel paradigm focused on maintaining relational consistency between instances under different augmentations, rather than explicitly pushing different instances apart as in \malakia|MoCo|. \malakia|ReSSL| models the relationship as a similarity distribution between a set of augmented images and uses this as a metric to align the same images with different augmentations. The method employs weak augmentations to represent more reliable relations, arguing that aggressive augmentations can hurt the reliability of target relations. \malakia|ReSSL| uses the KL divergence to measure and minimize the difference between the similarity distributions of two augmented views of the same image (see Figure \ref{fig:ressl}), effectively pushing for relational consistency. This approach, combined with a sharpened target distribution and a memory buffer with a momentum-updated network, allows \malakia|ReSSL| to achieve superior performance compared to \malakia|MoCo-v2|. Additionally, \malakia|ReSSL| demonstrates better performance on lightweight architectures, making it more suitable for smaller networks compared to traditional contrastive learning methods.

\malakia|MoBY| (\malakia|MoCo-v2| with \malakia|BYOL|) \cite{xie2021moby} combines key elements from \malakia|MoCo-v2| and \malakia|BYOL|~\cite{grill2020byol} for SSL with ViTs (like DeiT-S \cite{touvron2021deit} and Swin-T \cite{liu2021swin}). It adopts \malakia|MoCo-v2|’s momentum design, key queue, and contrastive loss, while incorporating \malakia|BYOL|'s asymmetric encoder architecture and data augmentations. This hybrid approach leverages a large negative sample dictionary and \malakia|BYOL|'s ability to learn without explicit negatives, resulting in high accuracy on linear evaluation with ViT backbones. Notably, \malakia|MoBY| uses a smaller batch size and a small memory of size 4096 to achieve competitive performance.

Momentum Contrastive Learning of Visual Representations (\malakia|MoCLR|) \cite{tian2021dnc} combines key elements from \malakia|MoCo|, \malakia|SimCLR| and \malakia|BYOL| for enhanced SSL. It incorporates \malakia|MoCo|'s momentum encoder and memory bank, \malakia|SimCLR|'s multiple augmentations and non-linear projection head, and \malakia|BYOL|'s prediction head and stop-gradient operations. \malakia|MoCLR| features two networks (shown in Figure \ref{fig:moclr}), each with a backbone and a projector, where the second network's parameters are updated via an EMA of the first network's parameters. Employing a symmetrized InfoNCE loss, \malakia|MoCLR| trains individual "expert" models on dataset subsets and distills their knowledge into a single model. Its unique dual contrastive learning mechanism applies loss between augmented views and  between online and momentum networks, leveraging strengths from each method to improve representation learning and downstream task performance.

\subsubsection{Neighborhood-based Methods}
\label{sec:contrastive_neighbor}

In contrastive learning, the terms "neighbor" or "neighborhood" typically refer to samples that are semantically or visually similar to a given anchor sample, often determined by proximity within the embedding space. This proximity is usually quantified using distance metrics like Euclidean distance or cosine similarity, which measure the closeness or separation of data points in the feature space. This subsection focuses on methods that leverage the concept of neighborhood to enhance contrastive learning by identifying and utilizing nearest neighbors.

\begin{figure*}[!t]
\centering
\begin{subfigure}[b]{0.25\textwidth}
    \centering
    \includegraphics[width=\linewidth,keepaspectratio]{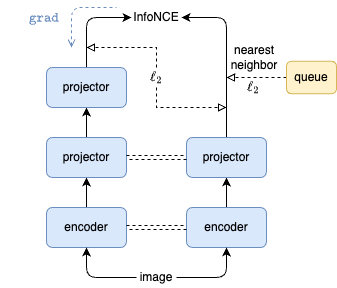}
    \caption{\xazomara{NNCLR}}
    \label{fig:nnclr}
\end{subfigure}%
\begin{subfigure}[b]{0.32\textwidth}
    \centering
    \includegraphics[width=0.95\linewidth,keepaspectratio]{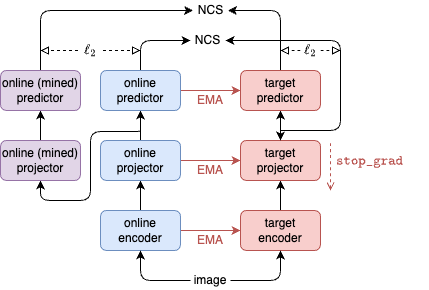}
    \caption{\xazomara{MYOW}}
    \label{fig:myow}
\end{subfigure}%
\begin{subfigure}[b]{0.23\textwidth}
    \centering
    \includegraphics[width=\linewidth,keepaspectratio]{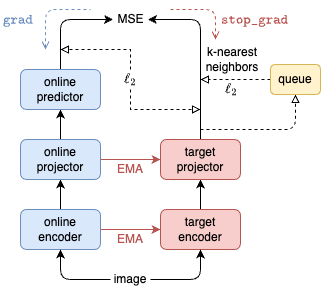}
    \caption{\xazomara{MSF}}
    \label{fig:msf}
\end{subfigure}
\caption{Comparison of contrastive SSL architectures employing nearest neighbors. (a) \xazomara{NNCLR} replaces the augmented view in \xazomara{SimCLR} with its neighbour from a memory bank. (b) \xazomara{MYOW} introduces a mined view to the \xazomara{BYOL} framework. (c) \xazomara{MSF} uses multiple nearest neighbours from a memory bank.}
\label{fig:nn-comparison}
\end{figure*}

Nearest-Neighbor Contrastive Learning of Visual Representations (\malakia|NNCLR|) \cite{dwibedi2021nnclr} samples nearest neighbors from the dataset in the latent space and treats them as positive examples, providing more semantic variations than pre-defined transformations (see Figure \ref{fig:nnclr}). It uses a support set, similar to \malakia|MoCo|'s queue, to store embeddings representing the full data distribution. However, unlike \malakia|MoCo| which uses the queue for negative samples, \malakia|NNCLR| uses it to find positive pairs through nearest neighbor search. The method employs a contrastive loss similar to \malakia|SimCLR|'s NT-Xent loss, but replaces one view of the positive pair with the nearest neighbor from the support set. This approach allows \malakia|NNCLR| to go beyond single-instance positives without resorting to clustering, leading to improved performance on various downstream tasks.

Mean Shift (\malakia|MSF|) \cite{koohpayegani2021msf} extends \malakia|NNCLR| by calculating the contrastive loss using the mean of the $K$ most similar sample representations from a memory bank (see Figure \ref{fig:msf}). This allows \malakia|MSF| to leverage multiple semantically similar samples, potentially providing a more robust and diverse set of positive examples.

\malakia|All4One| \cite{estepa2023all4one} advances \malakia|NNCLR| by introducing a novel contrastive SSL approach that combines three objectives: centroid contrast, neighbor contrast, and feature contrast. Drawing inspiration from other SSL methods, \malakia|All4One| incorporates elements from \malakia|NNCLR| for neighbor contrast, transformer-based models for the self-attention mechanism in centroid contrast, and redundancy reduction techniques like \malakia|Barlow Twins| (formally inroductd in \ref{sec:featuredecorrelationmethods}) for the feature contrast objective. The centroid contrast uses self-attention to create "centroids" from multiple neighbor representations, enabling the model to learn contextual information. The neighbor contrast, similar to \malakia|NNCLR|, facilitates learning directly from neighbors, while the feature contrast promotes learning unique feature representations.

Mine Your Own vieW (\malakia|MYOW|) \cite{azabou2021myow}, unlike \malakia|NNCLR|, employs a different mechanism to leverage a dataset's inherent diversity for richer representations. Building on \malakia|BYOL|'s dual network framework and representation learning without negative pairs, \malakia|MYOW| mines the dataset itself for similar yet distinct samples to serve as views for one another, rather than relying solely on data augmentations. \malakia|MYOW| actively mines these views by identifying neighboring samples in the network's representation space and then predicting one sample's latent representation from a nearby sample. Specifically, it uses mined views in addition to augmented ones, employing a projector-predictor setup for augmented views and a projector-projector-predictor setup for mined views, with the objective of minimizing the NCS between them (Figure \ref{fig:myow}). \malakia|MYOW| advances previous methods by overcoming limitations in domains where effective augmentations are unknown or limited, such as neuroscience, demonstrating the potential of harnessing data diversity for building rich representations where traditional augmentation techniques are insufficient.

MUlti-Granual Self Supervised learning (\malakia|Mugs|) \cite{zhou2022mugs}\footnote{\xazomara{Mugs} is also a framework and a dense method capturing detailed (fine-grained) to very abstract (coarse-grained) features for downstream tasks.} integrates elements from \malakia|MoCo|, \malakia|MSF|, \malakia|All4One|, and \malakia|PCL|, to boost performance in dense downstream tasks within a unified approach. Using a ViT architecture with momentum, \malakia|Mugs| combines instance discrimination with clustering via three distinct branches: instance-level, local-group, and group discrimination. The instance-level branch leverages a projector and predictor to implement the InfoNCE loss for direct instance comparison using a memory bank. The local-group branch combines patch tokens with their nearest neighbors through a bank using a transformer, aiming for localized context. Finally, the group discrimination branch abstracts to the cluster level using latent prototypes and a softmax-based CE loss.

\subsubsection{Frameworks}
\label{sec:contrastive_frameworks}

SSL frameworks and methods serve distinct roles in the SSL landscape. SSL methods are specialized algorithms or techniques designed to address specific self-supervised tasks, often tailoring their approaches to particular data types or applications. These methods implement unique combinations of components to achieve optimal performance in their target domains. In contrast, SSL frameworks provide comprehensive, flexible structures that encompass essential elements such as data augmentation strategies, pretext task definitions, and learning objectives. Frameworks offer a versatile foundation adaptable across diverse projects and domains, facilitating the development and comparison of multiple methods within a consistent environment. While methods focus on solving particular challenges with high efficiency, frameworks serve as the overarching architecture that enables the creation, implementation, and evaluation of various SSL approaches.

\begin{figure*}[!t]
\centering
\begin{subfigure}[b]{0.20\textwidth}
    \centering
    \includegraphics[width=\linewidth,keepaspectratio]{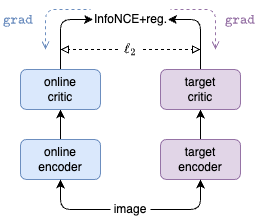}
    \caption{\xazomara{ReLIC}}
    \label{fig:relic}
\end{subfigure}%
\begin{subfigure}[b]{0.20\textwidth}
    \centering
    \includegraphics[width=\linewidth,keepaspectratio]{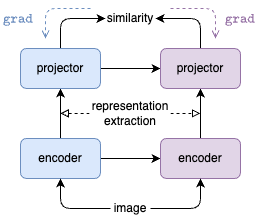}
    \caption{\xazomara{YADIM}}
    \label{fig:yadim}
\end{subfigure}
\begin{subfigure}[b]{0.232\textwidth}
    \centering
    \includegraphics[width=\linewidth,keepaspectratio]{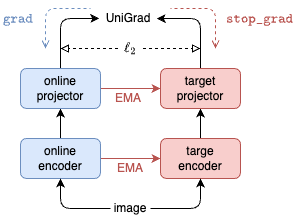}
    \caption{\xazomara{UniGrad}}
    \label{fig:unigrad}
\end{subfigure}
\caption{Comparison of different contrastive SSL frameworks providing unified theoretical perspectives. (a) \xazomara{ReLIC} incorporates causal mechanisms to learn invariant representations. (b) \xazomara{YADIM} offers a systematic characterization of contrastive methods through standardized data processing pipelines and representation extraction strategies. (c) \xazomara{UniGrad} presents a unified gradient analysis that connects seemingly disparate SSL approaches through their underlying gradient formulations.}
\label{fig:feature-comparison}
\end{figure*}

Yet Another \malakia|DIM| (\malakia|YADIM|) \cite{falcon2020yadim}, shown in Figure \ref{fig:yadim}, is a contrastive SSL approach that builds upon the conceptual framework characterizing contrastive methods in five aspects: data augmentation pipeline, encoder selection, representation extraction, similarity measure, and loss function. \malakia|YADIM| merges the data processing pipelines of \malakia|AMDIM| and \malakia|CPC| to create a new variant that demonstrates competitive performance on various downstream tasks. It shows improved robustness to the choice of encoder and representation extraction strategy compared to other methods. \malakia|YADIM| is part of a broader implementation that includes standardized versions of several prominent SSL methods, namely \malakia|AMDIM|, \malakia|CPC-v2|, \malakia|SimCLR|, \malakia|BYOL| and \malakia|MoCo-v2|.

On the other hand, Decoupled Contrastive Learning (\malakia|DCL|) \cite{yeh2022dcl} framework provides a comprehensive structure for implementing contrastive learning by addressing the negative-positive-coupling (NPC) effect, which can diminish gradient effectiveness, particularly with uninformative samples. In addition, \malakia|DCL| introduces a specific method within this framework, modifying the widely used InfoNCE loss by removing the positive term from the denominator, improving learning efficiency and reducing sensitivity to batch size. This decoupling allows \malakia|DCL| to achieve competitive performance without requiring large batch sizes, momentum encoding, or extensive training epochs, demonstrating improved training efficiency and robustness to suboptimal hyperparameters compared to methods like \malakia|SimCLR| and \malakia|MoCo-v2|. Building on \malakia|DCL|'s foundation, the variant \malakia|DCLW| introduces a weighting function for positive pairs, utilizing a negative von Mises-Fisher distribution to adjust their importance. Both \malakia|DCL| and \malakia|DCLW| can be combined with any InfoNCE loss-based framework.

Representation Learning via Invariant Causal mechanisms (\malakia|ReLIC|) is primarily considered an SSL method rather than a framework. \malakia|ReLIC| introduces a novel approach to SSL by leveraging causal mechanisms to learn invariant representations. \malakia|ReLIC| works by training an encoder to produce representations that are invariant to data augmentations while being predictive of the original input. It uses a contrastive loss function that encourages similarity between augmented views of the same image and dissimilarity between different images. Specifically, it minimizes the InfoNCE loss and a regularization term implemented as the KL divergence among different views under different augmentation (Figure \ref{fig:relic}). \malakia|ReLIC|'s key innovation is the incorporation of causal principles, aiming to capture invariant features that are causally related to the input data rather than spurious correlations. Although \malakia|ReLIC| is generically defined, which is why it can be considered a framework that implements various methods including \malakia|CPC|, \malakia|AMDIM|, \malakia|SimCLR|, and \malakia|BYOL|, it is primarily utilized as a standalone SSL method in practice. 
\malakia|ReLIC-v2| \cite{tomasev2022relicv2}\footnote{\xazomara{ReLIC-v2} is also a dense method.} is an extension of \malakia|ReLIC| that integrates an explicit invariance loss with a contrastive objective for dense downstream tasks. It incorporates a saliency mapping (images with background removed) within the augmentation and processes multiple data views of varying size. 

Contrastive Learning with Stronger Augmentations (\malakia|CLSA|) is also primarily an SSL method rather than a framework. \malakia|CLSA| builds upon existing contrastive learning approaches by introducing stronger data augmentations to improve the quality of learned representations. \malakia|CLSA| works by applying more aggressive data augmentations to create diverse views of input images. These augmentations likely include techniques such as color jittering, random cropping, and geometric transformations. The method then uses a contrastive loss function to encourage similarity between augmented views of the same image while pushing apart representations of different images. By using stronger augmentations, \malakia|CLSA| aims to create more challenging positive pairs, forcing the model to learn more robust and generalizable features. While \malakia|CLSA| itself is a method, it incorporates elements from other SSL approaches, particularly those in the contrastive learning family. It likely builds upon methods like \malakia|SimCLR| and \malakia|MoCo|, as evidenced by its use of contrastive learning principles and data augmentation strategies.

Finally, \malakia|UniGrad| \cite{tao2022unigrad} is an SSL framework that proposes a unified approach by analyzing and unifying different SSL methods through gradient analysis. It derives a unified gradient formula encompassing various approaches, including contrastive learning methods (e.g., \malakia|MoCo| and \malakia|SimCLR|), asymmetric network methods (e.g., \malakia|BYOL| and \malakia|SimSiam|) that use \texttt{stop-grad} and predictors, and feature decorrelation methods (e.g., \malakia|Barlow Twins| and \malakia|VICReg|), as shown in Figure \ref{fig:unigrad}. The framework demonstrates that these seemingly different methods can be unified into a similar gradient structure composed of positive and negative gradients. Based on this analysis, \malakia|UniGrad| proposes a concise gradient formula that explicitly maximizes the similarity between positive samples while expecting the similarity between negative samples to be zero. \malakia|UniGrad| does not explicitly include other SSL methods, but rather unifies and analyses them. It provides a framework for understanding and comparing different SSL approaches, showing that they work through similar mechanisms despite their apparent differences in loss functions. The \malakia|UniGrad| framework itself does not require a memory bank or a predictor network, making it a simple yet effective approach to SSL.

\subsubsection{Adversarial Methods}
\label{sec:contrastive_adversarial}

Adversarial methods and data augmentation both aim to improve model performance, but differ significantly. Augmentation uses predefined, computationally inexpensive transformations to increase data diversity and prevent overfitting, often resulting in visibly altered images. Adversarial methods, conversely, dynamically generate challenging, often imperceptible examples through computationally demanding optimization processes, leveraging gradient information to adapt to the model's current state and target specific weaknesses. While augmentations are generally model-agnostic and broadly applicable, adversarial methods can be tailored to specific model weaknesses, creating increasingly difficult examples as the model improves. In SSL, adversarial contrastive learning employs several techniques to enhance robustness and feature learning. These include generating instance-wise adversarial examples by adding small perturbations to maximize contrastive loss, producing harder negative pairs within a batch through adversarial optimization, and combining both approaches by creating challenging positive pairs through adversarial augmentations and harder negative pairs through adversarial training. Furthermore, techniques like multi-layer and cluster-driven perturbations generate adversarial examples based on features from different network layers or the representation space's clustering structure. Ultimately, while augmentation indirectly enhances performance through increased data diversity, adversarial methods in SSL directly challenge the model's understanding, fostering robustness and more discriminative feature learning by exposing the model to more challenging scenarios.

As an early adversarial SSL method, Contrastive Learning with Adversarial Examples (\malakia|CLAE|) \cite{ho2020clae} enhanced contrastive learning by incorporating adversarial examples. \malakia|CLAE| generates perturbations to augmented images, creating harder positive pairs for the contrastive loss. This forces the model to learn features invariant to these adversarial changes, resulting in more robust and discriminative representations. Influenced by contrastive learning methods like \malakia|SimCLR| and \malakia|MoCo|, as well as adversarial training techniques, \malakia|CLAE| contributed to the growing body of research combining adversarial learning with SSL, paving the way for more advanced techniques like adversarial data augmentation in SSL.

Different from \malakia|CLAE| which actively generates adversarial examples to "attack" the model during training, forcing it to learn features robust to those specific attacks, Robust Contrastive Learning \malakia|RoCL| \cite{kim2020rocl} takes a more defensive approach by using robust optimization. Instead of directly minimizing the contrastive loss on augmented views of an image, \malakia|RoCL| considers a "worst-case" scenario within a small perturbation set around each augmented view. It seeks to minimize the maximum contrastive loss that could be incurred by any perturbation within this set, effectively "defending" against a wider variety of possible variations, making it more robust. So, while both methods strive for robustness, \malakia|CLAE| is like training with specific attack simulations, while \malakia|RoCL| is like building a generally strong defence against any potential threat.

Self-Supervised Adversarial Training (\malakia|SAT|) \cite{chen2020sat} is a discriminative SSL method designed to enhance adversarial robustness. Unlike contrastive methods, \malakia|SAT| doesn't rely on contrastive loss or positive/negative pairs. Instead, it adapts traditional adversarial training to self-supervised settings, often using models like \malakia|AMDIM| as initial models. \malakia|SAT| generates adversarial examples via iterative gradient-based attacks, like Projected Gradient Descent (PGD), perturbing inputs within a defined bound. Specifically, it leverages pretext tasks (e.g., predicting image rotations or solving jigsaw puzzles) to create self-supervised pseudo-labels. Adversarial examples are then generated based on these self-supervised tasks, and the model is trained to maximize the mutual information between clean and perturbed samples, ensuring robustness against adversarial attacks. 

Adversarial Contrastive Learning (\malakia|AdCo|) \cite{hu2021adco} is another discriminative SSL method that directly learns a set of negative adversaries to play against the self-trained representation network. It works by alternately updating two players: the representation network and negative adversaries, to obtain challenging negative examples against which positive queries are trained to discriminate. It updates negative adversaries towards a weighted combination of positive queries by maximizing the adversarial contrastive loss, allowing them to closely track representation changes over time. \malakia|AdCo| was influenced by existing contrastive learning methods like \malakia|MoCo-v2|, but improved upon them by more efficiently updating negative examples.

Cooperative-adversarial Contrastive (\malakia|CaCo|) \cite{wang2022caco} extends \malakia|AdCo| by jointly learning positive and negative representations end-to-end. While \malakia|CaCo|, like\malakia|AdCo|, adversarially optimizes negative samples to maximize the contrastive loss, it cooperatively optimizes positive samples to minimize this loss, encouraging alignment with the query anchor and making them more challenging for the encoder. This dynamic interplay ensures that both positive and negative samples adapt continuously to the evolving representations of the query anchors across mini-batches.

Decoupled Adversarial Contrastive Learning (\malakia|DeACL|) \cite{zhang2022deacl} employs a two-stage framework to improve adversarial robustness in SSL without using labels. In the first stage, \malakia|DeACL| performs standard SSL to obtain a non-robust encoder. In the second stage, this pretrained encoder acts as a teacher model, generating pseudo-targets to guide supervised adversarial training on a student model. This approach decouples the complex problem of integrating adversarial training into SSL into two simpler sub-problems: non-robust SSL and pseudo-supervised adversarial training.

Finally, \malakia|DiRA| (Discriminative, Restorative, and Adversarial Learning)\cite{haghighi2022dira} is an SSL framework that unifies three complementary components (i.e., discriminative, restorative, and adversarial learning), to collaboratively enhance representation learning. The discriminative component focuses on learning high-level semantic representations by maximizing agreement between similar instances in the latent space. The restorative component enforces the model to reconstruct fine-grained details of input images, ensuring that localized information is preserved. The adversarial component strengthens feature learning by introducing adversarial perturbations during restoration, encouraging the model to capture robust and generalizable representations. \malakia|DiRA| combines these components into a unified training framework with a joint loss function, enabling it to assemble complementary visual information from unlabeled data. \malakia|DiRA| was influenced by earlier SSL methods like \malakia|MoCo-v2|, \malakia|SimSiam| and \malakia|Barlow Twins| for its discriminative learning principles, as well as generative approaches for its restorative component. By combining these ideas, it bridges the gap between discriminative and generative SSL methods. 

\subsubsection{Methods Inspired from Diverse Fields}
\label{sec:contrastive_inspired}

This subsection explores the cross-pollination of ideas from diverse fields, incorporating principles from areas outside traditional contrastive learning. By integrating concepts like max-margin classification or manifold learning, these methods aim to expand the capabilities of SSL.

Max-Margin Contrastive Learning (\malakia|MMCL|) \cite{shah2021mmcl} enhances contrastive learning by applying principles from Support Vector Machines (SVMs) to select high-quality negative samples and maximize the decision margin. \malakia|MMCL| computes a discriminative hyperplane for each positive example using an SVM objective, identifying the most informative negatives as support vectors. These "hard" negatives are then used to minimize their similarity with positive examples while maximizing positive pair similarity. To improve computational efficiency, \malakia|MMCL| simplifies the SVM objective using techniques like projected gradient descent or truncated least squares solvers, enabling end-to-end training. In contrast to methods like \malakia|MoCo|, \malakia|SimCLR| and \malakia|DCL|, \malakia|MMCL| requires neither large batch sizes nor momentum encoding, reduces training time, exhibits less sensitivity to hyperparameters, and improves generalization to downstream tasks like transfer learning and few-shot recognition.

Maximum Manifold Capacity Representations (\malakia|MMCR|) \cite{yerxa2023mmcr} takes a unique approach, drawing inspiration from "manifold capacity". Instead of contrasting image views, \malakia|MMCR| aims to maximize the number of linearly separable categories within the learned representation space. Unlike \malakia|MMCL|, which uses SVM-based max-margin principles, \malakia|MMCR| optimizes the geometry of the entire representation space by aligning and uniformly distributing embeddings. In simpler terms, \malakia|MMCR| tries to arrange objects in a space so that lines can be easily drawn to separate them into different groups; the more groups that can be separated, the better the arrangement. \malakia|MMCR| aims to learn representations that make this separation easy. It achieves this by introducing a simplified manifold capacity objective, optimizing it to learn representations that compress object manifolds while preserving class separability. Specifically, \malakia|MMCR| uses a backbone and projector, generates multiple augmented views and processes them through both encoders to produce aggregated $\ell_2$-normalized embeddings, calculates batch centroids, and optimizes the nuclear norm of the resulting centroid matrix. This approach enhances linear separability, allowing \malakia|MMCR| to match or surpass state-of-the-art SSL methods on benchmarks like linear evaluation.

Style Augmentations for Self-Supervised Learning (\malakia|SASSL|) \cite{rojasgomez2024sassl} addresses the limitations of conventional data augmentation methods in preserving semantic information during self-supervised pretraining. Unlike traditional augmentation techniques that may distort the inherent structure of natural images, \malakia|SASSL| leverages Neural Style Transfer to decouple semantic and stylistic attributes in images. By applying transformations exclusively to the style component while preserving content, \malakia|SASSL| generates diverse samples that better retain semantic information. The method introduces a controlled stylization process using feature blending and image interpolation factors to ensure semantic consistency. \malakia|SASSL| offers two strategies for style references: external stylization using pre-computed representations from alternate datasets, and in-batch stylization utilizing other images from the same minibatch. When integrated into established self-supervised methods like \malakia|MoCo|, \malakia|SimCLR|, and \malakia|BYOL|, \malakia|SASSL| demonstrates significant improvements in ImageNet classification accuracy and transfer learning performance without increasing pretraining computational costs.



Table \ref{tab:contrastive}, provides a summary of the various contrastive learning methods discussed above, highlighting essential attributes that define and differentiate each approach. The table is structured to showcase the critical aspects of these methods, including the type of loss objective employed, the network architecture (whether siamese or single), specifics on how positive examples are defined ("views"), and the strategy for sampling negatives. Since the introduction of \malakia|MoCo|, features such as momentum encoders \cite{he2020moco} and memory structures \cite{wu2018instdis} have become integral to the design of these methods.


\subsubsection{Dense Contrastive Methods}
\label{sec:contrastive_dense}

Dense contrastive methods in SSL address the limitations of traditional approaches by optimizing contrastive loss at the local feature or pixel level, making them ideal for dense prediction tasks like segmentation and object detection. Unlike methods that collapse spatial information into global image representations, dense methods employ dense projection heads to maintain spatial structure and compute the contrastive loss between corresponding local features across augmented views. Instance discrimination, for example, uses global image-level feature representations obtained by average pooling the final convolutional feature map. This approach lacks spatial consistency in the convolutional features, which is crucial for tasks like object detection using architectures such as Faster R-CNN \cite{ren2016fasterrcnn}. While image classification favors translation and scale invariance, reducing objects of various scales and locations to a single category, object detection requires translation and scale equivariance. The following section highlights the commonly employed SSL methods emphasizing dense contrastive learning. By retaining spatial details, dense contrastive methods enable better transferability to downstream applications requiring detailed spatial understanding, bridging the gap between self-supervised pre-training and tasks demanding pixel-level accuracy.

Dense Contrastive Learning (\malakia|DenseCL|) \cite{wang2021densecl} extends the \malakia|MoCo-v2| framework for dense prediction tasks by optimizing a pairwise contrastive (dis)similarity loss at the pixel level between two views of input images. \malakia|DenseCL| employs a specialized dense projection head to generate feature maps, enhancing the model's ability to process spatially detailed information. Adapting the InfoNCE loss to each dense local feature independently creates a more comprehensive learning objective, combining the traditional InfoNCE loss with a novel dense loss. Demonstrating the benefits of dense supervision, \malakia|DenseCL| is considered a key early contribution to dense contrastive learning.

View-Agnostic Dense Representation (\malakia|VADeR|) \cite{pinheiro2020vader}  is another early dense method that learns pixelwise representations by forcing local features to remain constant over different viewing conditions through pixel-level contrastive learning. It uses an encoder-decoder architecture and computes similarity on pixel-level representations, leveraging known pixel correspondences derived from the view generation process to match local features. Specifically, \malakia|VADeR| uses the InfoNCE loss, with temperature-calibrated cosine similarity at the pixel level.

Pixel-level consistency methods (\malakia|PixContrast| and \malakia|PixPro|) \cite{xie2021pixpro} employ a momentum encoder to distinguish each pixel within an image, treating each pixel as a distinct class. \malakia|PixContrast| identifies positive and negative pixel pairs based on their relative spatial locations, using a contrastive approach to emphasize spatial sensitivity in feature learning. Specifically, it considers pixels within a certain spatial proximity as positive pairs and pixels further apart as negative pairs. \malakia|PixPro| builds upon this by smoothing pixel features based on the similarity of neighboring pixels. A Pixel Propagation Module (PPM) aggregates information from surrounding pixels, refining each pixel's feature representation. Instead of explicit positive/negative pairs, \malakia|PixPro| emphasizes consistency between original and propagated (smoothed) pixel features, effectively capturing local spatial relationships without requiring explicit negative samples. This pixel-to-propagation consistency forms the basis of its learning objective.

Contrastive Detection (\malakia|DetCon|) \cite{henaff2021detcon} is built on top of \malakia|BYOL| and \malakia|SimCLR|, leading to \malakia|DetConB| and \malakia|DetConS| versions. \malakia|DetCon| introduces a novel self-supervised objective, tasking representations with identifying object-level features across augmentations. Using an unsupervised segmentation algorithm, \malakia|DetCon| generates masks for each view to identify approximate object regions in images. Mask-pooled representations, calculated via the dot product of masks and features, focus on these regions of interest identified through segmentation. Subsequent projection/prediction networks generate latent features upon which the InfoNCE loss is applied. \malakia|DetConS|'s key innovation is the incorporation of object-centric priors into SSL, making it well-suited for downstream detection tasks.

Region Similarity Representation Learning (\malakia|ReSim|) \cite{xiao2021resim} learns both regional representations for localization and semantic image-level representations through self-supervision. It operates by sliding a fixed-size window across overlapping areas between two image views, aligning these areas with corresponding convolutional feature map regions and maximizing feature similarity. Specifically, \malakia|ReSim| uses the Region Proposal Network (RPN) from Faster R-CNN \cite{ren2016fasterrcnn} to extract features from overlapping regions of the two views, sliding windows over the C3 or C4 feature maps (representing progressively higher-level abstractions, with C3 capturing localized details and C4 capturing global semantic information). \malakia|ReSim|'s objective is to make common overlapping regions similar and non-common regions dissimilar by optimizing a weighted sum of local and global (InfoNCE) losses. \malakia|ReSim|'s key contribution is learning spatially and semantically consistent feature representations throughout the convolutional feature maps, enabling downstream tasks to leverage these representations for localization.

Instance Localization (\malakia|InsLoc|) \cite{yang2021insloc} extends the \malakia|MoCo-v2| framework specifically for object detection pretraining. It creates composite images by pasting image instances at various locations and scales onto background images. Region of Interest Align (RoI Align) \cite{girshick2015fastrcnn} is utilized to extract foreground features from the backbone feature maps. These features are then used with the InfoNCE loss to learn representations by treating different augmentations of the same object as positive pairs and features from different objects as negative pairs. Its key innovation is integrating bounding box information into pretraining, promoting better task and architecture alignment for transfer learning to detection. A subsequent case study \cite{buettner2022insloc} further explored \malakia|InsLoc|, proposing strategies to augment views and enhance robustness in appearance-shifted and context-shifted scenarios.

\malakia|VICReg| Large (\malakia|VICRegL|) \cite{bardes2022vicregl} extends the original \malakia|VICReg| (Variance-Invariance-Covariance Regularization) SSL method, which aims to prevent representation collapse by applying three regularization terms: variance, invariance, and covariance (discussed further in Section \ref{sec:featuredecorrelationmethods}). \malakia|VICRegL| builds upon this by introducing a local loss computed on patches extracted from full-resolution feature maps, in addition to the global loss computed on pooled features. The pipeline of \malakia|VICRegL| involves feeding two augmented views of an image through an encoder, producing both pooled and unpooled representations. These representations are then passed through an expander to generate embeddings. The method applies the \malakia|VICReg| loss to the global embeddings and introduces a feature vector matching process for local embeddings. This approach allows \malakia|VICRegL| to learn both global and local features simultaneously, improving performance on dense prediction tasks like segmentation while maintaining strong performance on classification tasks.

Copy-Paste Contrastive Pretraining (\malakia|CPCP| or \malakia|CP2|) \cite{wang2022cp2}, builds upon \malakia|MoCo-v2| but extends it with a pixel-wise contrastive loss and a copy-paste mechanism. \malakia|CP2| creates composite images by copying random foreground crops onto different backgrounds, training a segmentation model with two objectives: distinguishing foreground from background pixels and identifying composed images sharing the same foreground. Using InfoNCE loss for both instance-level (applied to average foreground features) and pixel-level (applied to feature maps) contrastive learning, \malakia|CP2| introduces a Quick Tuning protocol—initializing with pretrained weights and fine-tuning the entire segmentation model—to quickly adapt classification models to segmentation. While sharing copy-paste techniques with \malakia|InsLoc|, \malakia|CP2| differs by focusing on dense prediction and using a segmentation head during pretraining, specifically targeting semantic segmentation (rather than object detection pretraining) with a dense contrastive loss for enhanced pixel-level representation learning.

Selective Object COntrastive learning (\malakia|SoCo|) \cite{wei2021soco} builds upon \malakia|MoCo-v2| by introducing object-level representations to SSL, specifically designed for object detection tasks. \malakia|SoCo| utilizes off-the-shelf selective search to generate object proposals, treating each proposal as an independent instance rather than considering the whole image. This allows for a scale-aware assignment strategy where object proposals are assigned to different pyramid levels according to their size, encouraging scale-invariant representations. Unlike image-level methods, \malakia|SoCo| performs pretraining over all network modules used in detectors (backbone, FPN, and R-CNN head), enabling well-initialized weights for all detector layers. Through multiple augmented views with different scales and locations of the same object, \malakia|SoCo| learns object-level translation and scale invariance, achieving state-of-the-art transfer performance on COCO detection with significant improvements over supervised pretraining baselines.

DetCo (\malakia|DetCo|) \cite{xie2021detco} builds upon contrastive learning frameworks like \malakia|MoCo-v2| to create a method that performs well on both instance-level detection tasks and image classification simultaneously. Unlike concurrent works that focused on detection but sacrificed classification performance, \malakia|DetCo| achieves a better trade-off through two key innovations: (1) multi-level supervision on features from different backbone stages, ensuring discriminative representations in each level of the feature pyramid, and (2) contrastive learning between global images and local patches using separate MLP heads. The global-local contrastive loss combines three components: global-to-global, local-to-local, and global-to-local, enabling consistent feature learning across scales. \malakia|DetCo| significantly outperforms methods like \malakia|DenseCL| and \malakia|InsLoc| on ImageNet classification while maintaining competitive performance on detection and segmentation tasks, demonstrating how detection and classification can mutually improve through proper architecture design.

Instance Consistency (\malakia|InsCon|) \cite{yang2022inscon} builds upon the framework of \malakia|MoCo-v2| to provide enhanced feature representations for downstream prediction tasks like object detection and instance segmentation. \malakia|InsCon| introduces three parallel learning heads to capture different levels of instance information: (1) the original single-instance head from \malakia|MoCo-v2| for basic instance discrimination, (2) a multi-instance head that processes a composite image created from multiple augmented views arranged in a grid pattern, enabling the model to handle complex multi-instance recognition scenarios, and (3) a cell-instance head that focuses on fine-grained features by establishing strict cell-level correspondence between views with overlapping areas. The multi-instance head uses an adaptive average pooling layer to generate multiple feature representations, while the cell-instance head employs Precise ROI Pooling to extract matched cell features from overlapping areas between views. By combining these three learning paradigms with a unified contrastive loss function, \malakia|InsCon| creates a complementary relationship between different granularities of feature representation, improving performance on object recognition, boundary localization, and content extraction in downstream tasks.

Unified Self-supervised Visual Pre-training (\malakia|UniVIP|) \cite{li2022univip} is designed to learn versatile visual representations from both single-centric-object (e.g., ImageNet) and non-iconic (e.g., COCO \cite{li2015coco}) datasets. \malakia|UniVIP| builds upon the \malakia|BYOL| framework, adopting its architecture and loss function as a foundation for scene-level consistency. However, \malakia|UniVIP| extends \malakia|BYOL| by incorporating instance-level learning and using an optimal transport algorithm to measure instance discrimination. The \malakia|UniVIP| pipeline involves three key steps: 1) Using the unsupervised Selective Search algorithm to generate candidate object instances, 2) Creating two scene views with overlapping regions containing instances to ensure global similarity, and 3) Applying representation learning at three levels: scene-scene similarity, scene-instance correlation, and instance-instance discrimination. 

Table \ref{tab:dense}, summarizes the key architectural and operational attributes of the aforementioned dense contrastive methods. 


\subsection{Clustering Methods}
\label{sec:clusteringmethods}

Clustering-based discriminative SSL methods aim to learn representations by grouping similar samples together in the feature space. The main idea behind these methods is to leverage the inherent structure of the data to create pseudo-labels or assignments, which are then used to guide the learning process. Unlike contrastive methods that focus on instance-level discrimination, clustering-based approaches operate at a higher level by forming clusters of semantically similar instances. However, they share similarities with contrastive methods in that both aim to learn discriminative features without relying on manual labels. Clustering-based SSL methods typically use techniques like k-means or more sophisticated algorithms to assign samples to clusters, and then optimize the model to predict these cluster assignments or to align representations with learned prototypes. These methods appeared after contrastive SSL methods, building upon the success of earlier approaches and addressing some of their limitations, such as the need for large batch sizes or negative samples. Clustering-based methods like \malakia|SwAV| \cite{caron2020swav} have shown competitive performance with contrastive methods while offering unique advantages in terms of scalability and the ability to capture hierarchical semantic structures in the data. This section provides a description of the popular self-supervised methods centered around clustering.

\malakia|Deep Cluster| \cite{caron2019deepcluster} is an early unsupervised learning method that jointly learns the parameters of a neural network and cluster assignments of the resulting features. It iteratively extracts features from images using a CNN, reduces their dimensionality, and clusters them using k-means. These cluster assignments then serve as pseudo-labels to update the network weights through backpropagation, alternating between feature clustering and network updates. While predating most contrastive SSL methods, \malakia|Deep Cluster| relies on the principle that good features should be clusterable, using cluster assignments as a form of self-supervision. \malakia|Deeper Cluster| \cite{caron2019deepercluster} extends \malakia|Deep Cluster| by incorporating a self-supervised pretext task—grouping features using pseudo-labels derived from angular rotations \cite{gidaris2018rotnet}—alongside the clustering objective. This combined approach leverages the benefits of both clustering-based and SSL, making \malakia|Deeper Cluster| \cite{caron2019deepercluster} more effective for learning from large-scale, uncurated datasets compared to \malakia|Deep Cluster|.

Local Aggregation (\malakia|LA|) \cite{zhuang2019la} aims to learn visual embeddings by optimizing a soft clustering structure in the feature space. Unlike \malakia|Deep Cluster|, which uses global clustering for pseudo-labels, \malakia|LA| identifies neighbors separately for each example, allowing for more flexible statistical structures. Specifically, \malakia|LA| iteratively extracts features using a CNN, identifies background and close neighbors for each sample, and optimizes a local aggregation metric that encourages similar samples to cluster (using k-means) together while separating dissimilar ones. \malakia|LA| employs a non-parametric softmax operation to compute sample recognition probabilities, which are then used in the objective function. Compared to \malakia|Deep Cluster|, \malakia|LA| does not require an additional readout layer or frequent recomputation of cluster labels, resulting in improved training efficiency.

Online Deep Clustering (\malakia|ODC|) \cite{zhan2020odc} seeks to overcome the instability of alternating clustering and network updates in \malakia|Deep Cluster| by performing these processes simultaneously within each network update iteration. The training process consists of forward and backward propagations, label re-assignment, and centroid updates in each iteration. \malakia|ODC| maintains two dynamic memory modules: a sample memory to store sample labels and features, and a centroid memory for centroid evolution. This approach allows for continuous label updates rather than periodic global clustering, enabling more stable representation learning. Unlike \malakia|Deep Cluster|, \malakia|ODC|, similar to \malakia|LA|, does not require an additional readout layer or frequent recomputation of cluster labels, improving training efficiency. \malakia|ODC| also introduces loss re-weighting to prevent training collapse and a strategy for handling small clusters to prevent drifting into a few large clusters.

Self Labeling (\malakia|SeLa|) \cite{asano2020sela} also seeks to overcome \malakia|Deep Cluster|'s limitation of alternating between clustering and network updates by optimizing a single objective function that combines both tasks. The process involves extracting image features using a CNN, applying a fast version \cite{cuturi2013sinkhorn} of the Sinkhorn-Knopp (SK) clustering algorithm to assign pseudo-labels, and then training the network to predict these labels using cross-entropy loss. \malakia|SeLa| introduces an equipartition constraint to prevent degenerate solutions, formulating the problem as an optimal transport task. The method can employ multiple projection heads for different clustering tasks, sharing the feature extractor parameters.

Constraint K-means (\malakia|CoKe|) \cite{qian2022coke} introduces online constrained k-means for clustering-based pretext tasks, effectively separating clustering from discrimination to optimize single-view instances. In the clustering phase, \malakia|CoKe| uses an online constrained k-means algorithm that only constrains the minimum size of each cluster (to avoid cluster collapse), enabling more flexible data structure capture compared to balanced clustering methods like \malakia|SeLa|. This online assignment method is theoretically guaranteed to approach the global optimum. In the discrimination phase, \malakia|CoKe| uses a standard normalized softmax loss with labels and centers recorded from the previous epoch to learn representations. Unlike \malakia|Deep Cluster|, which requires offline batch-mode clustering, \malakia|CoKe|'s online approach allows for more efficient training.

\begin{figure*}[!t]
\centering
\begin{subfigure}[b]{0.195\textwidth}
    \centering
    \includegraphics[width=\linewidth]{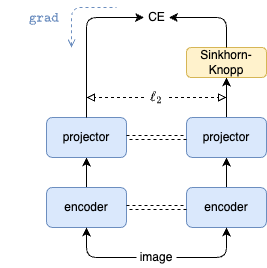}
    \caption{\xazomara{SwAV}}
    \label{fig:swav}
\end{subfigure}%
\begin{subfigure}[b]{0.24\textwidth}
    \centering
    \includegraphics[width=\linewidth]{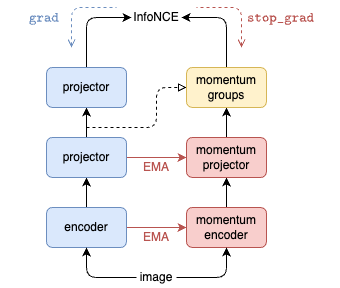}
    \caption{\xazomara{SMoG}}
    \label{fig:smog}
\end{subfigure}
\begin{subfigure}[b]{0.21\textwidth}
    \centering
    \includegraphics[width=\linewidth]{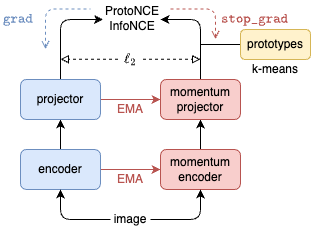}
    \caption{\xazomara{PCL}}
    \label{fig:pcl}
\end{subfigure}%
\begin{subfigure}[b]{0.165\textwidth}
    \centering
    \includegraphics[width=\linewidth]{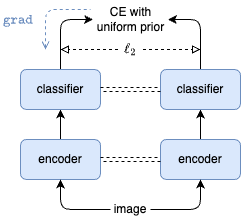}
    \caption{\xazomara{Self-Class}}
    \label{fig:selfclassifier}
\end{subfigure}
\caption{Comparison of different clustering SSL architectures. (a) \xazomara{SwAV}  utilizes siamese networks to process two views of an image, aligning embeddings with trainable prototypes via the SK algorithm . (b) In \xazomara{SMoG} , two networks learn synchronously with group assignments updated in real-time, using InfoNCE loss variants. (c) \xazomara{PCL} uses prototypes as latent variables to guide the network towards effective group discrimination. (d) \xazomara{Self-Classifier} replaces clustering in favor of direct label learning through a modified CE loss}
\label{fig:cluster-comparison}
\end{figure*}

Swapping Assignments between multiple Views of the same image (\malakia|SwAV|) \cite{caron2020swav} is a novel SSL method combining clustering and contrastive learning. It employs Siamese networks to generate multiple augmented views of an image (Figure \ref{fig:swav}), extracts features using a CNN, and computes cluster assignments (codes) for these features using learnable prototypes. The SK algorithm is utilized to efficiently compute these online cluster assignments, treating it as an optimal transport problem with an entropic constraint. \malakia|SwAV| introduces a "swapped" prediction mechanism, predicting the code of one view from the representation of another view of the same image, enforcing consistency between cluster assignments without explicit pairwise feature comparisons. This approach incorporates ideas from contrastive learning but replaces instance-level comparisons with cluster assignment comparisons. \malakia|SwAV| also introduces a multi-crop augmentation strategy for improved efficiency and performance and, unlike some other clustering-based methods, can be trained with both large and small batches and scales well to large datasets. \malakia|SwAV| further re-implemented and improved the \malakia|SeLa| and \malakia|Deep Cluster| frameworks, resulting in \malakia|SeLa-v2| and \malakia|Deep Cluster-v2|, respectively. These versions incorporate stronger data augmentation, an MLP projection head, a cosine learning rate schedule, a temperature parameter, a memory bank, and multi-clustering, as introduced in \malakia|SwAV|.

SElf-supERvised learning (\malakia|SEER|) \cite{goyal2021seer, goyal2022seer} is a large-scale SSL approach that trains RegNet-Y \cite{radosavovic2020regnet} architectures using the \malakia|SwAV| method on random, uncurated internet images. \malakia|SEER| extracts features, applies online clustering to group similar visual concepts, and optimizes a contrastive loss between different views of the same image. Leveraging billions of random internet images without supervision, \malakia|SEER| scales \malakia|SwAV|'s online clustering and contrastive learning principles to much larger models and datasets, efficiently handling extreme scale and diversity. Incorporating techniques like mixed precision training, gradient checkpointing, and sharded optimization, \malakia|SEER| distinguishes itself from other cluster-based methods through its ability to train at such scale.

Semantic Clustering by Adopting Nearest neighbors (\malakia|SCAN|) \cite{vangansbeke2020scan} introduces a novel two-step approach for unsupervised image classification. First, \malakia|SCAN| learns feature representations through a pretext task and mines nearest neighbors based on feature similarity, leveraging these neighbors as a semantic prior—unlike methods using k-means clustering after feature learning—and employing a self-labeling step to mitigate noise in nearest neighbor selection. Second, \malakia|SCAN| integrates this prior into a learnable approach, classifying each image and its mined neighbors together using a loss function that maximizes their dot product while encouraging consistent and discriminative predictions. \malakia|SCAN|'s reliance on more meaningful features, rather than network architecture alone, distinguishes it from end-to-end approaches.

Prototypical Contrastive Learning (\malakia|PCL|) \cite{li2021pcl} (see Figure \ref{fig:pcl}) introduces prototypes as latent variables in an Expectation-Maximization (EM) framework, where high-confidence samples become class prototypes. Its two-step process involves finding the prototype distribution via clustering (E-step) and optimizing the network via contrastive learning using a novel ProtoNCE loss (M-step). This ProtoNCE loss, a generalized InfoNCE loss, encourages representations closer to assigned prototypes while maintaining instance-level discrimination. Unlike \malakia|SCAN|, which uses nearest neighbors as a semantic prior, \malakia|PCL| directly learns prototypes through the EM framework. Building upon instance-wise contrastive learning, \malakia|PCL| incorporates semantic structures discovered through clustering, enabling it to learn both low-level features and higher-level semantics, improving performance, especially in low-resource transfer learning.

Synchronous Momentum Grouping (\malakia|SMoG|) \cite{pang2022smog} replaces instance-level contrastive learning with a group-level approach, effectively mimicking clustering-based methods within a contrastive learning framework. Using a Siamese network structure (with the first network comprising a backbone, projector, and predictor, and the second network comprising a backbone and projector, as shown in Figure \ref{fig:smog}) with momentum-updated encoders, \malakia|SMoG| extracts features from two augmented views of an image. Instead of directly contrasting instance features, \malakia|SMoG| dynamically assigns instances to groups and performs contrastive learning at the group level. A momentum grouping scheme synchronously conducts feature grouping with representation learning, enabling gradient propagation through group features. This approach addresses the supervisory signal hysteresis issue faced by clustering-based methods and reduces false negatives common in instance contrastive methods.

Finally, \malakia|Self-Classifier| \cite{amrani2022selfclassifier} is a novel, end-to-end self-supervised classification learning approach that simultaneously learns labels and representations in a single stage. It optimizes for same-class prediction of two augmented views of the same sample, using a shared network comprised of a backbone (e.g., CNN) and a classifier (e.g., projection MLP and linear classification head). To avoid degenerate solutions, \malakia|Self-Classifier| employs a mathematically motivated variant of the cross-entropy loss with a uniform prior asserted on the predicted labels. Unlike \malakia|SCAN| and other cluster-based SSL methods that often require pre-training, expectation-maximization, pseudo-labeling, or external clustering, \malakia|Self-Classifier| operates in a single stage without these additional steps. The pipeline is simpler and more scalable, processing two augmented views of an image through the shared network and minimizing their cross-entropy to promote same-class prediction (Figure \ref{fig:selfclassifier}).

Clustering-based SSL methods offer several advantages over contrastive approaches. By forming clusters of similar instances, they can operate at a higher semantic level, potentially capturing more meaningful data structures. Often requiring smaller batch sizes and not relying on negative samples, they offer improved scalability and efficiency. Furthermore, they can better capture hierarchical semantic structures. However, clustering methods also have drawbacks. They can be more sensitive to data augmentation choices, as overly aggressive transformations can distort image information and hinder effective clustering. Additionally, being a more recent development than contrastive learning, they may not have benefited from the same level of extensive research and optimization. Some clustering approaches may also require complex design strategies, such as careful management of learnable prototypes. Despite these challenges, clustering-based methods have demonstrated competitive performance while offering unique advantages in scalability and semantic structure capture. Table \ref{tab:cluster} summarizes the clustering methods discussed above, highlighting their structural differences and specific features, including the presence of a projector or predictor, the use of momentum or memory structures, the incorporation of prototypes for guiding clustering, and the employment of the SK algorithm for distribution alignment. The loss objectives used by each method are also detailed.

\subsection{Self-distillation Methods}
\label{sec:selfdistillationmethods}

Self-distillation methods in discriminative SSL improve representation learning by leveraging information from different parts of the same network. Unlike contrastive methods (which focus on instance-level discrimination) and clustering-based methods (which explicitly form clusters or prototypes), self-distillation methods prioritize internal consistency across different parts of the network or different augmentations of the same input. Learning without labeled data or negative samples, they often employ a teacher-student framework for knowledge transfer within the same model or between similar architectures. These methods typically use two encoders (or two views of the same encoder) processing differently augmented views of an image, with one encoder acting as a target for the other. To prevent encoder collapse, where a constant output is predicted for any input, techniques like updating one encoder's weights with a running average of the other's are commonly used. The following describes the most common self-distillation methods.

\begin{figure*}[!t]
\centering
\begin{subfigure}[b]{0.21\textwidth}
    \centering
    \includegraphics[width=\linewidth,keepaspectratio]{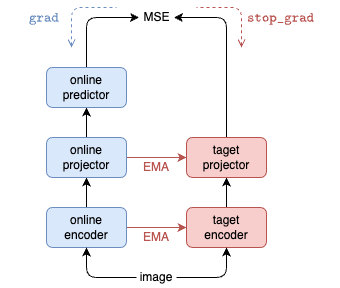}
    \caption{\xazomara{BYOL}}
    \label{fig:byol}
\end{subfigure}%
\begin{subfigure}[b]{0.21\textwidth}
    \centering
    \includegraphics[width=\linewidth,keepaspectratio]{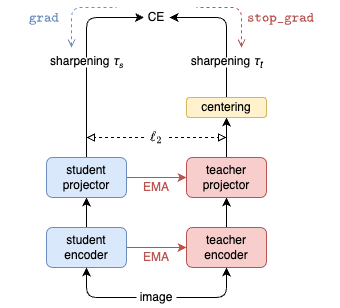}
    \caption{\xazomara{DINO}}
    \label{fig:dino}
\end{subfigure}
\begin{subfigure}[b]{0.19\textwidth}
    \centering
    \includegraphics[width=\linewidth,keepaspectratio]{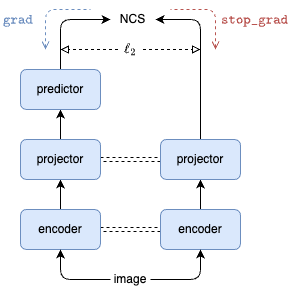}
    \caption{\xazomara{SimSiam}}
    \label{fig:simsiam}
\end{subfigure}
\caption{Comparison of different self-distillation SSL architectures. (a) \xazomara{BYOL} employs a dual network architecture where the target network's parameters are updated using an EMA of the online network's parameters. (b) \xazomara{DINO} uses a student-teacher setup where the student learns to replicate the teacher's output, with the teacher's parameters being a EMA of the student's. (c) \xazomara{SimSiam} operates two identical networks processing different views of the same image, using a stop-gradient operation to prevent output collapse.}
\label{fig:selfdistil-comparison}
\end{figure*}

Bootstrap Your Own Latent (\malakia|BYOL|) \cite{grill2020byol}, a seminal SSL method, uses two neural networks—an online network and a target network—to learn image representations without relying on negative pairs. The process involves passing two augmented views of an image through these networks to produce ($\ell_2$)-normalized embeddings. The online network (comprising an encoder, projector, and predictor) is trained to predict the target network's representation of the same image under a different augmentation (see Figure \ref{fig:byol}). The target network (sharing the same architecture as the online network, but without the predictor) is updated as an EMA of the online network. The objective is to minimize the MSE between the online network's prediction and the target network's embedding. \malakia|BYOL| avoids representation collapse through its asymmetric architecture, predictor network, stop-gradient operation on the target network, and the EMA update mechanism, which together prevent the networks from converging to trivial solutions. While influenced by contrastive SSL, \malakia|BYOL| eliminates the need for negative pairs, instead focusing on predicting representations across different augmented views. This approach has proven more robust to image augmentation choices and achieves state-of-the-art performance on various benchmarks.

\malakia|C-BYOL| \cite{lee2021cbyolcsimclr} advances \malakia|BYOL| by incorporating information compression using the Conditional Entropy Bottleneck (CEB), which allows for measuring and controlling the amount of information compression in the learned representation, leading to improved accuracy and robustness to domain shifts across various scenarios.

Self-DIstillation with NO labels (\malakia|DINO|) \cite{caron2021dino} is another seminal self-distillation method that uses a teacher-student architecture to learn visual representations without labels. The pipeline involves passing two augmented views of an image through student and teacher networks with identical architectures but different weights (see Figure \ref{fig:dino}). The student network is trained to predict the teacher network's output using a CE loss, with the teacher's output centered and sharpened to avoid collapse. The teacher network is updated as an EMA of the student network. Similar to \malakia|BYOL| in its use of a momentum encoder and self-distillation, \malakia|DINO| differs by directly predicting the teacher's output (without a prediction head) and using a CE loss instead of MSE. \malakia|DINO| has shown particularly strong performance with ViTs, demonstrating "emergent properties" such as unsupervised segmentation through attention maps.

\malakia|DINO-v2| \cite{oquab2023dinov2}, an improvement upon \malakia|DINO|, integrates features from \malakia|DINO|, \malakia|iBOT| \cite{zhou2022ibot}, and \malakia|SwAV|, along with a new regularizer \cite{sablayrolles2019koleo}, to enhance training stability and speed. It introduces several innovations, such as masked input patches and untied head weights, aimed at improving learning efficiency and feature diversity.

Simple Siamese (\malakia|SimSiam|) \cite{chen2020simsiam} method learns representations by predicting one view of an image from another view of the same image. It does this passing two augmented views through a shared encoder network followed by a projection MLP (see Figure \ref{fig:simsiam}). One branch then applies a prediction MLP, while the other uses a stop-gradient operation. The model is trained to maximize the cosine similarity between the predicted and stopped representations. \malakia|SimSiam| differs from \malakia|BYOL| by eliminating the need for a momentum encoder, and from \malakia|DINO| by not using a teacher-student framework or temperature-based softmax. The stop-gradient operation and predictor MLP, and the use of NCS loss, are crucial for preventing collapse and enabling effective learning without negative samples or momentum encoders. Fast Siamese (\malakia|FastSiam|) \cite{pototzky2022fastsiam} advances \malakia|SimSiam| by introducing a multi-crop strategy and a novel loss function that combines intra-image and inter-image consistency, allowing for faster convergence and improved performance. 

Finally, Online Bag-of-Visual-Words (\malakia|OBoW|) \cite{gidaris2021obow} uses a teacher-student architecture to learn visual representations without labels by processing two augmented views of an image through the teacher and student networks. The teacher network generates a Bag-of-Visual-Words (BoW) representation (a technique representing images as histograms of quantized local features, analogous to word histograms in text analysis) using an online-updated vocabulary of visual words. The student network is trained to predict this BoW representation from a different augmented view. \malakia|OBoW| performs online training of both networks and continuously updates the visual-word vocabulary, enabling fully online BoW-guided unsupervised learning. Unlike \malakia|BYOL|, \malakia|SimSiam|, and \malakia|DINO|, which focus on making global image embeddings invariant to augmentations, \malakia|OBoW| uses high-dimensional BoW vectors as targets, capturing multiple local visual concepts and aiming to promote contextual reasoning.

Table \ref{tab:selfdistil} summarizes the self-distillation methods discussed, illustrating their use of projectors, predictors, and momentum encoders. A key feature common to all these methods is the \texttt{stop-grad} operation, crucial for preventing network collapse into trivial solutions, and the lack of memory use. The "networks" column specifies the architectural setup, noting the different roles (e.g., "online/target" or "student/teacher") used in these Siamese network variations. The table also details the loss functions and symmetry of each method, providing insights into their specific optimization mechanisms.

In summary, self-distillation SSL methods offer several distinct advantages. While all SSL methods are sensitive to augmentations, self-distillation, by focusing on internal consistency, can be argued to be less reliant on carefully engineered augmentations than contrastive or clustering methods. This focus on internal consistency also simplifies training and reduces computational costs by avoiding the complexities of momentum encoders, negative samples, large memory queues, and complex clustering, making some self-distillation methods, like \malakia|SimSiam|, surprisingly simple to train. Furthermore, this focus can encourage more robust and meaningful representations, sometimes leading to remarkable emergent properties like unsupervised object segmentation, as seen in \malakia|DINO|. Finally, self-distillation methods are often more computationally efficient than traditional contrastive approaches, not requiring large batch sizes or memory banks, and have shown promise for improving performance on smaller datasets and models. While they technically do not require large batch sizes, they typically employ them in practice to achieve optimal performance. However, self-distillation methods also have disadvantages. They are susceptible to collapse, even with mitigation strategies, and can be sensitive to architectural and hyperparameter choices, often requiring careful tuning. Compared to contrastive learning, self-distillation is less understood theoretically, hindering the development of new methods and improvements to existing ones. Finally, there is a potential for redundancy in learned features, as different parts of the network might learn similar representations, which can limit the overall information captured. Like other SSL methods, self-distillation faces open challenges, including improving performance on smaller models, developing efficient training for large-scale datasets, and enhancing transferability to diverse downstream tasks.

\subsection{Knowledge Distillation Methods}
\label{sec:kdmethods}

Knowledge distillation methods in SSL leverage a pre-trained "teacher" model to guide the learning of a "student" model, even when labeled data is scarce or unavailable \cite{hinton2015distilling}. Unlike self-distillation, which uses different parts of the same network (or similar architectures for both teacher and student) to provide the learning signal, knowledge distillation employs a separate, typically larger or more powerful, teacher network. This teacher network, pre-trained using any available (even limited) labeled data or through a self-supervised task, provides "soft" targets (e.g., probability distributions or feature representations) to the student network. The student network, which can be smaller or have a different architecture, learns to mimic the teacher's behavior, effectively distilling the learned knowledge. This differs significantly from self-distillation, which relies on internal consistency within a single network. The main advantages of knowledge distillation in SSL include improving the performance of smaller models, enabling SSL in resource-constrained environments, and transferring rich representations from large models to more efficient architectures. These methods address the challenge of SSL's ineffectiveness with smaller models, which limits its applicability in edge devices and other computationally constrained scenarios. The following provides an overview of typical approaches used in self-supervised knowledge distillation.

Contrastive Representation Distillation (\malakia|CRD|) \cite{tian2022crd} offers a novel approach to knowledge distillation by leveraging contrastive learning objectives to transfer representational knowledge between networks. Unlike traditional knowledge distillation methods that minimize KL divergence between output probabilities, \malakia|CRD| focuses on maximizing MI between teacher and student representations. The method introduces a framework that treats knowledge transfer as a binary classification problem. \malakia|CRD| uses a contrastive loss that pushes representations of the same input closer together while pushing representations of different inputs apart.

Relational Representation Distillation (\malakia|RRD|) \cite{giakoumoglou2024rrd} approaches self-supervised knowledge distillation by addressing the limitations of existing contrastive learning methods. Unlike traditional approaches that minimize KL divergence or use rigid instance-discrimination techniques, \malakia|RRD| focuses on preserving the relative relationships between instances in the feature space. The method introduces separate temperature parameters for teacher and student distributions. By allowing sharper student outputs, \malakia|RRD| can precisely learn primary relationships while maintaining secondary similarities between instances. This approach goes beyond direct embedding transfer, capturing the broader structural relationships in the representation space, similar to \malakia|ReSSL|. This approach ensures that the self-supervised objective captures not just individual instance characteristics, but the broader structural relationships between instances. Theoretically connected to both InfoNCE loss and KL divergence, \malakia|RRD| aims to transfer knowledge more comprehensively.

Discriminative and Consistent Representation Distillation (\malakia|DCD|) \cite{giakoumoglou2024dcd} combines contrastive learning with consistency regularization to create a unique self-supervised knowledge distillation approach. The method introduces learnable temperature and bias parameters that adapt during training, allowing for more flexible self-supervised feature transfer. By balancing discriminative power with structural consistency, \malakia|DCD| uses self-supervised objectives to ensure that the student model not only mimics the teacher's individual instance representations but also captures the broader distributional characteristics of the teacher's learned features.

Multi-Mode Online Knowledge Distillation (\malakia|MOKD|) \cite{song2023mmkd} leverages SSL by enabling two models to learn collaboratively through both self-distillation and cross-distillation modes. Unlike traditional knowledge distillation, \malakia|MOKD| uses self-supervised contrastive learning to help heterogeneous models (like CNNs and ViTs) extract complementary features. The method's cross-distillation mechanism allows models to transfer knowledge through a self-supervised objective, focusing on enhancing semantic feature alignment across different model architectures.

Self-Supervised Knowledge Distillation (\malakia|SSKD|) \cite{xu2020sskd} directly incorporates self-supervision as an auxiliary task for knowledge transfer. By using contrastive prediction as a supplementary learning objective, \malakia|SSKD| extracts richer knowledge from the teacher network. The method explores various self-supervised tasks (contrastive learning \cite{chen2020simclr}, exemplar \cite{dosovitskiy2015exemplar}, jigsaw \cite{noroozi2017unsupervised}, and rotation prediction \cite{gidaris2018rotnet}) to create a more comprehensive knowledge transfer approach. Its selective transfer strategy ensures that only meaningful self-supervised signals are used, making it particularly effective in extracting structured knowledge from self-supervised representations.

Distill-on-the-Go (\malakia|DoGo|) \cite{bhat2021dogo} employs a unique approach to self-supervised knowledge distillation through mutual learning. It uses contrastive representation learning as its core self-supervised objective, where two models generate embeddings from augmented views of the same sample. The method's innovative contribution lies in aligning the softmax probabilities of similarity scores between peer models, effectively using self-supervised contrastive learning as a mechanism for knowledge transfer. Unlike other methods, \malakia|DoGo| focuses on collaborative learning where models simultaneously improve each other's representations.

\begin{figure*}[!t]
\centering
\begin{subfigure}[b]{0.17\textwidth}
    \centering
    \includegraphics[width=\linewidth,keepaspectratio]{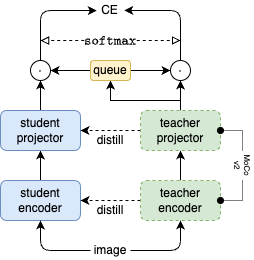}
    \caption{\xazomara{SEED}}
    \label{fig:seed}
\end{subfigure}%
\begin{subfigure}[b]{0.25\textwidth}
    \centering
    \includegraphics[width=\linewidth,keepaspectratio]{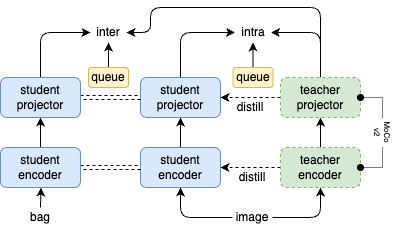}
    \caption{\xazomara{BINGO}}
    \label{fig:bingo}
\end{subfigure}
\begin{subfigure}[b]{0.25\textwidth}
    \centering
    \includegraphics[width=\linewidth,keepaspectratio]{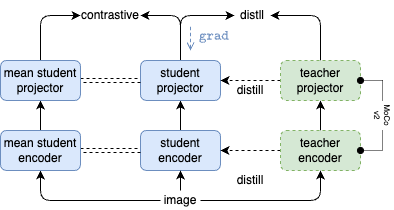}
    \caption{\xazomara{DisCo}}
    \label{fig:disco_1}
\end{subfigure}%
\begin{subfigure}[b]{0.27\textwidth}
    \centering
    \includegraphics[width=\linewidth,keepaspectratio]{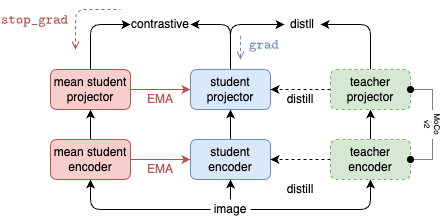}
    \caption{\xazomara{DisCo}}
    \label{fig:disco_2}
\end{subfigure}
\caption{Comparison of different knowledge-distillation SSL architectures. (a) \xazomara{SEED} utilizes a teacher-student architecture where the student network learns from the teacher by minimizing the cross-entropy loss between their softmax-transformed embeddings. (b) \xazomara{BINGO} aggregates features from an anchor image and similar instances into a bag of features, then processes them through both student and teacher networks, applying combined intra- and inter-contrastive losses. (c, d) \xazomara{DisCo} employs a triplet of networks — student, mean student, and teacher — where the student learns from both its average self and the pretrained teacher through a combination of contrastive and distillation losses. \xazomara{DisCo} can either use Siamese network that share parameters (c) or use momentum on the mean student (d).}
\label{fig:knowledgedistil-comparison}
\end{figure*}

SElf-SupErvised Distillation (\malakia|SEED|) \cite{fang2021seed} is an early method to introduce knowledge distillation to SSL, addressing the challenge of poor performance in smaller models trained directly with contrastive SSL methods. It is designed to improve the performance of smaller models by distilling knowledge from a larger pre-trained teacher model to a smaller student model without using labeled data (see Figure \ref{fig:seed}). \malakia|SEED| first trains a large teacher model using a self-supervised approach like \malakia|MoCo-v2| or \malakia|SimCLR|. Then, during distillation phase, \malakia|SEED| maintains a queue of data samples and computes softmax-transformed similarity scores between an input instance and the queued samples using both teacher and student networks. The student is trained to mimic the teacher's similarity score distribution, minimizing the cross-entropy between these distributions.

Distilled Contrastive Learning (\malakia|DisCo|) \cite{gao2022disco} also trains a large teacher model using a self-supervised approach and then distills knowledge to a smaller student model. Unlike \malakia|SEED|, which mimics similarity score distributions, it directly distills the teacher's final embeddings to the student. \malakia|DisCo| has two variants: a Siamese version where teacher and student process the same input (Figure \ref{fig:disco_1}), and a momentum version using a sample queue similar to \malakia|MoCo| (Figure \ref{fig:disco_2}). \malakia|DisCo| incorporates a "mean student" encoder, used alongside the student and teacher encoders in computing the contrastive and consistency losses, which helps stabilize training and improve distilled knowledge quality. Specifically, \malakia|DisCo| processes two augmented views of an image through all three networks, calculating a contrastive loss between student and mean student embeddings from different views, and a distillation loss as the sum of MSE between student and teacher embeddings for both views, encouraging the student to mimic the teacher's representations. \malakia|DisCo|'s main novelty is directly transferring the teacher's latent manifold, enabling more effective knowledge transfer, especially for lightweight models, and addressing the "distilling bottleneck" by enlarging the embedding dimension. Influenced by contrastive SSL, \malakia|DisCo| incorporates ideas from \malakia|MoCo-v2| and \malakia|SwAV|, adapting them for distillation (e.g., teacher pre-training).

Bag of InstaNces aGgregatiOn (\malakia|BINGO|) \cite{xu2022bingo} also uses student and teacher networks, each with a backbone and projector (see Figure \ref{fig:bingo}). However, unlike \malakia|SEED|, which distills similarity score distributions, and \malakia|DisCo|, which distills final embeddings, \malakia|BINGO| constructs bags of similar instances using teacher embeddings (pretrained with methods like \malakia|MoCo-v2| or \malakia|SwAV|). Each bag contains an anchor instance and other highly similar instances. It then applies a bag-aggregation distillation loss, comprising intra-sample distillation (pushing together embeddings of two augmentations of the same instance) and inter-sample distillation (pushing embeddings of all bag instances toward the anchor). \malakia|BINGO|'s main novelty is transferring relationships among similar samples, rather than simply mimicking teacher outputs or embeddings.

Table \ref{tab:knowledgedistill} compares the aforementioned methods, highlighting their diverse loss functions (e.g., softmax with CE, combinations of InfoNCE and MSE) for minimizing feature representation discrepancies between student and teacher. These methods are heavily influenced by {Hinton et al.} work \cite{hinton2015distilling}, which, while focused on supervised learning, introduced the core concepts of knowledge distillation and demonstrated its effectiveness in transferring knowledge from larger teacher models to smaller student models, laying the foundation for knowledge distillation in SSL. Knowledge distillation methods offer a number of advantages over self-distillation and contrastive methods. Compared to self-distillation, they can transfer knowledge from larger, more powerful teacher models, potentially improving performance on resource-constrained devices. Unlike contrastive methods, they do not require large batch sizes or negative samples, increasing efficiency and flexibility. Moreover, they often perform better on smaller datasets, especially with noisy or limited labels. However, knowledge-distillation methods typically require a two-stage process: pre-training the teacher and then distilling knowledge to the student, which can be more time-consuming than end-to-end approaches. While generally less computationally expensive during inference, they may require more resources during training due to this two-stage process. Common challenges include designing effective distillation tasks, handling domain shifts, and balancing model compression and performance. Furthermore, the choice of teacher model significantly impacts student performance, and achieving consistent performance across different architectures remains a challenge.

\subsection{Feature Decorrelation Methods}
\label{sec:featuredecorrelationmethods}

The self-supervised feature decorrelation family proposes novel losses to address the representation learning paradigm with self-supervision \cite{balestriero2023cookbook}. As a subset of it, the canonical correlation analysis family originates with the Canonical Correlation Analysis (CCA) \cite{hotelling1992relations}. The high-level goal of CCA is to infer the relationship between two variables by analyzing their cross-covariance matrices \cite{zbontar2021barlowtwins}. Feature decorrelation methods focus on creating representations that are diverse and independent, aiming to enhance the robustness and generalizability of learned features. By reducing redundancy among features, these approaches aim to improve the robustness and generalization of models. The following is a description of the most popular feature decorrelation methods.

\begin{figure*}[!t]
\centering
\begin{subfigure}[b]{0.195\textwidth}
    \centering
    \includegraphics[width=\linewidth]{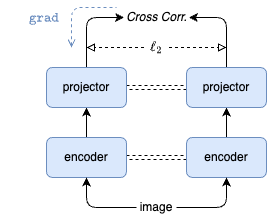}
    \caption{\xazomara{Barlow Twins}}
    \label{fig:bt}
\end{subfigure}%
\begin{subfigure}[b]{0.185\textwidth}
    \centering
    \includegraphics[width=\linewidth]{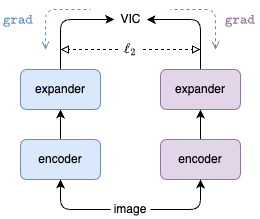}
    \caption{\xazomara{VICReg}}
    \label{fig:vicreg}
\end{subfigure}
\begin{subfigure}[b]{0.195\textwidth}
    \centering
    \includegraphics[width=\linewidth]{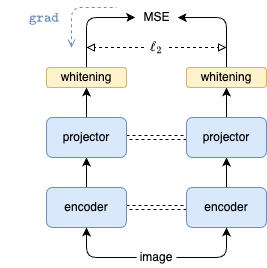}
    \caption{\xazomara{W-MSE}}
    \label{fig:wmse}
\end{subfigure}%
\begin{subfigure}[b]{0.195\textwidth}
    \centering
    \includegraphics[width=\linewidth]{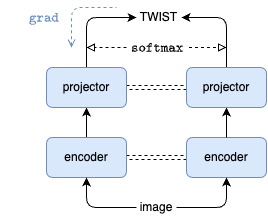}
    \caption{\xazomara{TWIST}}
    \label{fig:twist}
\end{subfigure}
\caption{Comparison of different feature decorrelation SSL architectures. (a) \xazomara{Barlow Twins} utilizes two siamese networks aiming to minimize redundancy across the embeddings by aligning the cross-correlation matrix to the identity matrix. (b) \xazomara{VICReg} employs a variance-invariance-covariance regularization strategy on the embeddings from two distinct views processed through separate networks. (c) \xazomara{W-MSE} implements a whitening MSE approach to normalize embeddings into a spherical distribution. (d) \xazomara{TWIST} utilizes twin siamese networks to generate class distribution probabilities for two augmented views of the same image, aiming for consistency across these distributions.}
\label{fig:cca-comparison}
\end{figure*}

\malakia|Barlow Twins| \cite{zbontar2021barlowtwins} (named after the neuroscientist {Horace Barlow}) employs a novel approach to SSL by leveraging the redundancy reduction principle, originally proposed in neuroscience. This method utilizes two identical Siamese networks, each consisting of a backbone and a projector, to process two different views of the same image generated through stochastic augmentation (see Figure \ref{fig:bt}. In this setup, each image view is transformed into embeddings by the encoder network, which are then mean-centered along the batch dimension to ensure an average output of zero over the batch, effectively normalizing the embeddings (i.e., $\ell _1$-normalized). \malakia|Barlow Twins| to minimize a unique loss function based on the cross-correlation matrix between the embeddings of the two views. This loss function aims to ensure embeddings remain consistent across augmentations while minimizing redundancy among their components, effectively promoting invariance and feature decorrelation. \malakia|Barlow Twins| stand out among other methods as they do not demand large batch sizes \malakia|SimCLR|, nor rely on asymmetric components like prediction networks \malakia|BYOL|, \malakia|SimSiam|, momentum encoders \malakia|MoCo|, \malakia|MoCo-v2|, \malakia|BYOL|, or non-differentiable operations or stop-gradients \malakia|SimSiam|. It benefits from employing high-dimensional embeddings. 

Variance-Invariance-Covariance Regularization (\malakia|VICReg|) \cite{bardes2022vicreg} uses two regularization terms applied to embeddings separately. It employs two networks for learning that are not required to share weights or the same architecture, as shown in Figure \ref{fig:vicreg}. Each network comprises a backbone and an expander serving as the projection head. Given an image, \malakia|VICReg| generates two views using stochastic augmentations. These views are processed by the backbone and expander of each network to produce embeddings. To ensure robust and informative feature learning, \malakia|VICReg| incorporates two regularization terms: one that maintains the variance of each embedding dimension above a specified threshold and another that decorrelates each pair of embedding variables using the off-diagonal elements of the covariance matrix, similar to \malakia|Barlow Twins|. Additionally, \malakia|VICReg| incorporates the MSE between the embeddings of the different views to prevent the model from collapsing into trivial solutions where the embeddings become constant or non-informative across different inputs. This dual regularization approach ensures that the variance of each embedding dimension is maintained while decorrelating variables, effectively preventing encoder collapse and ensuring that the learned representations remain meaningful and distinct.

Whitening Mean Squared Error (\malakia|W-MSE|) \cite{ermolov2021wmse} is a loss function that is based on the whitening of the latent space features. The framework utilizes a network that comprises of a backbone and a projector, as depicted in Figure \ref{fig:wmse}. It derives multiple positive pairs from a single image, deeming negatives unnecessary. These views are processed by the network to produce the embeddings. The embeddings are mapped onto a spherical distribution using a whitening transform and then undergo $\ell _2$-normalization. The normalized whitened representations are used to compute the MSE implemented as NCS over all positive pairs. \malakia|W-MSE| conceptually aligns with \malakia|Barlow Twins|. It diverges from approaches using momentum networks \malakia|BYOL| or stop-gradient networks \malakia|SimSiam|, promotes non-degenerate distributions through a whitening operation that normalizes representations to a spherical distribution with zero mean and identity covariance matrix. This method avoids the use of negative pairs by bringing positive pairs closer together, using batch slicing to manage the variability of the whitening matrix.

Subsequent developments like \malakia|Mixed Barlow Twins|, \malakia|TLDR|, and \malakia|ARB| build upon \malakia|Barlow Twins|' framework, incorporating distinctive elements to refine and stabilize the learning. \malakia|Mixed Barlow Twins| tackles the potential decrease in feature quality with higher embedding dimensions by enhancing sample interactions and adding a regularization term to prevent overfitting. \malakia|TLDR| and \malakia|ARB| expand on the primary concept by incorporating nearest neighbors and the Nearest Orthonormal Basis (NOB) into their methodologies, respectively. More precisely, \malakia|Mixed Barlow Twins| \cite{bandara2023mixedbarlowtwins} enhances the original \malakia|Mixed Barlow| framework by incorporating linearly interpolated samples to mitigate feature overfitting. Utilizing two identical Siamese networks, this approach processes two distinct views of the same image, generated via stochastic augmentation. A third, mixed view is also created by linearly blending these two augmented views. \malakia|Mixed Barlow Twins| retains the objective of \malakia|Barlow Twins|, supplementing it with a regularization component. This is achieved by aligning the cross-correlation matrix of the mixed view with that of each original view against the respective ground truth cross-correlation matrices. The overall loss for \malakia|Mixed Barlow Twins| combines the original \malakia|Barlow Twins| objective with this additional regularization term. Align Representations with Base (\malakia|ARB|) \cite{zhang2022arb} builds on the \malakia|Barlow Twins| framework but uses the NOB \cite{strang2006linear} as an objective to maximize the MI between the immediate variable generated from one view and representations of the other view. Specifically, it minimizes the output matrix of one view and the nearest orthonormal basis matrix of the other view. Twin Learning for Dimensionality Reduction (\malakia|TLDR|) \cite{kalantidis2022tldr} builds on the \malakia|Barlow Twins| framework but employs the nearest neighbor along a view as a pair whose proximity is preserved for the purpose of dimensionality reduction.

Different from the above, Twin Class Distribution Estimation (\malakia|TWIST|) \cite{wang2021twist} utilizes Siamese networks, concluding with a softmax operation to yield twin class distribution probabilities for two augmented versions of the same image (see Figure \ref{fig:twist}). It aims for consistency across these class distributions by maximizing MI between the inputs and class predictions. \malakia|TWIST| further sharpens the class distribution for each image to promote decisive class assignments and fosters diversity among predictions to avoid biased classifications towards a dominant class, achieving a balanced and accurate classification regime. Finally, \malakia|Truncated Triplet| \cite{wang2021truncatedtriplet} builds upon the \malakia|BYOL| framework, introducing a truncated triplet loss which extends the conventional triplet loss \cite{sohn2016improved}. This new loss function aims to maximize the relative distance between positive pairs and negative pairs, addressing issues of under-clustering and over-clustering common in self-supervised settings.

Orthogonality Regularization (\malakia|OR|) \cite{he2024or} applies regularization across the encoder, targeting both convolutional and linear layers during pretraining. By enforcing orthogonality within weight matrices, \malakia|OR| prevents dimensional collapse—a phenomenon where a few large eigenvalues dominate the eigenspace—ensuring that filters remain uncorrelated and maintain a norm of $1$. This regularization indirectly stabilizes hidden features and representations, promoting uniform eigenvalue distributions throughout the network. Leveraging two key regularizers, Soft Orthogonality (SO) and Spectral Restricted Isometry Property (SRIP), \malakia|OR| integrates seamlessly with existing SSL frameworks, enhancing performance.

In Table \ref{tab:cca}, we outline the key components of various feature decorrelation methods, highlighting their similarities and differences. Most methods listed employ a projector and utilize a Siamese architecture, reflecting a common framework for this category of methods. The loss functions employed by these methods are notably distinct and innovative, as they work directly on the features themselves to enhance decorrelation. For instance, methods like \xazomara{Barlow Twins} and \xazomara{TLDR} use a cross-correlation matrix to minimize redundancy between features, while \xazomara{VICReg} employs a combination of variance, invariance, and covariance regularization to achieve feature decorrelation. This table provides a clear view of how each method approaches the challenge of reducing feature redundancy to enhance model generalizability and robustness.

\section{Benchmarking of Self-supervised Learning Methods}
\label{sec:benchmarking}

Self-supervised evaluation is crucial for assessing the quality and transferability of learned representations in computer vision tasks. It provides a standardized framework for comparing different SSL methods and their effectiveness in capturing meaningful features from unlabeled data. The main evaluation frameworks include linear evaluation, where a linear classifier is trained on frozen features to measure representation quality; semi-supervised fine-tuning, which assesses performance with limited labeled data; and transfer learning to downstream tasks such as object detection and segmentation. The importance of SSL evaluation lies in its ability to validate the generalization capabilities of self-supervised models, their data efficiency in low-label regimes, and their potential to reduce reliance on large-scale labeled datasets. As SSL continues to advance, robust evaluation protocols ensure that progress in the field translates to real-world improvements across various computer vision applications. This section details the standard protocols used to assess SSL-trained model performance, focusing on how these representations perform under various evaluation setups. We also compiled and summarized performance results from relevant papers. This section is organized as follows. First, we examine evaluation protocols on the source dataset in Section~\ref{sec:evaluationofsourcedata}, covering both linear evaluation and semi-supervised approaches. Next, we analyze transfer learning capabilities to other classification tasks in Section~\ref{sec:tlcapabilities}, followed by an assessment of transferability to more complex vision tasks such as detection and segmentation in Section~\ref{sec:tlcapabilities2}. Throughout each subsection, we present both the methodological frameworks and comparative results across various self-supervised learning approaches.

\subsection{Evaluation on the Source Dataset}
\label{sec:evaluationofsourcedata}

This section describes the evaluation protocols employed to measure the effectiveness of the SSL method on the source dataset. This process involves training a model on unlabeled data using SSL techniques, then evaluating its performance on the same dataset using labeled samples. Common evaluation methods include linear probing, where a linear classifier is trained on frozen features, and fine-tuning, where the entire model is updated using limited labeled data. These evaluations provide insights into the model's ability to capture meaningful features from unlabeled data, its generalization capabilities, and its potential for downstream tasks. 

\subsubsection{Evaluation Protocols}

Linear evaluation in SSL assesses representation quality by freezing a pretrained encoder (i.e., the backbone), training only a linear classifier on labeled data, and measuring classification accuracy on held-out test sets to determine feature separability and transferability. It was popularized by SSL frameworks like \malakia|SimCLR|, \malakia|MoCo|, and \malakia|BYOL|, and previous studies \cite{kolesnikov2019revisiting}, although the concept of linear evaluation existed in earlier forms. 

More specifically, linear evaluation includes the following steps: (1) a model (e.g., ResNet, ViT) is pretrained on unlabeled data using SSL objectives; (2) the pretrained encoder's weights are frozen to isolate representation quality; (3) a single linear layer (or MLP) is appended to the frozen encoder and trained exclusively on labeled data (e.g., ImageNet) using CE loss and Stochastic Gradient Descent (SGD) \cite{ruder2017sgd} with Nesterov momentum, typically for up to 100 epochs, without regularization - during this phase, only basic augmentations are applied for preprocessing: training images undergo resizing to $256 \times 256$, random resized cropping to $224 \times 224$, horizontal flipping, and normalization, while validation images undergo resizing to $256 \times 256$, center cropping to $224 \times 224$, and normalization, excluding heavy pretraining augmentations like color distortion; (4) the model's quality is evaluated by measuring classification accuracy on a held-out test set (e.g., ImageNet validation set), where higher accuracy indicates better feature separability and generalizability.

Linear evaluation assesses two critical properties of SSL representations: linear separability (whether features naturally cluster by semantic class) and transferability (whether representations generalize to downstream tasks without task-specific tuning), all while isolating feature quality by freezing the encoder. The protocol has gained widespread adoption due to its simplicity (requiring minimal computational resources), reproducibility (enabling direct comparison across SSL methods), and theoretical alignment with the principle that quality representations should permit simple linear decision boundaries. By focusing solely on the linear classifier's performance atop frozen features, this method provides a controlled benchmark that distinguishes representation quality from fine-tuning capabilities, though it may not fully capture performance on complex non-linear tasks.

Semi-supervised evaluation assesses the quality of learned representations by combining SSL pretraining with limited labeled data during downstream task adaptation, following protocols from \cite{kornblith2019do, zhai2020large}. The method emerged alongside SSL breakthroughs, driven by frameworks like \malakia|SimCLR| and \malakia|BYOL| and has become a cornerstone for measuring SSL's label efficiency and generalization capabilities. While linear evaluation (training a classifier on frozen features) was standardized earlier, semi-supervised evaluation gained traction as a more practical benchmark for real-world scenarios where labels are scarce. 

Semi-supervised evaluation assesses the label efficiency and transferability of SSL representations through a three-stage pipeline: (1) a backbone model (e.g., ResNet, ViT) is pretrained on unlabeled data using SSL objectives; (2) semi-supervised fine-tuning integrates a small labeled subset (e.g., 1-10$\%$ of ImageNet) with the remaining unlabeled data, and the encoder weights are updated using these limited labels; (3) the fine-tuned model's performance is evaluated using classification accuracy on held-out test sets, such as the ImageNet validation set. 

Fine-tuning and linear evaluation differ fundamentally in their methodological approach and evaluation focus.  
Fine-tuning updates all or most pre-trained encoder weights using smaller learning rates (e.g., 0.01-0.001) with layer-wise learning rate decay to prevent catastrophic forgetting of pretrained features, adapting the feature extraction for the specific downstream task, and evaluating the adaptability of representations by measuring how well the pre-trained features serve as initialization for task-specific learning compared to random initialization. Similar spatial augmentations are applied during both training and testing phases. Optimization of the CE loss continues under the same conditions as in the linear setup. In addition to exploring various learning rates, the duration of training epochs is also varied to determine the optimal settings that yield the best results on a local validation set (usually the epochs vary between 30 and 50). Conversely, linear evaluation freezes the entire pre-trained encoder backbone, training only a linear classifier layer with high learning rates to assess the linear separability of learned representations, essentially testing if the SSL objective has created a feature space where classes naturally cluster. Thus, linear evaluation focuses on the inherent quality of the learned representations, while fine-tuning assesses their adaptability to specific tasks.

\subsubsection{Datasets}

The most commonly used dataset utilized as the primary source for training and evaluating SSL models is ImageNet ILSVRC-2012 (also known as ImageNet-1K), a subset of the larger ImageNet-22K \cite{deng2009imagenet} database that has become a standard benchmark in computer vision. It contains 1,000 object categories organized according to the WordNet hierarchy, with a total of 1,281,167 training images, 50,000 validation images, and 100,000 test images. ImageNet-1K features RGB images of varying resolutions that were collected by scraping online image search engines using synonyms in multiple languages. The images are quality-controlled and human-annotated, with each image labeled with exactly one class identifier. To facilitate a broader evaluation, datasets such as CIFAR-10, CIFAR-100 \cite{krizhevsky2009cifar}, Places205 \cite{zhou2018places365}, Instagram-1B \cite{yalniz2019instagram} (private dataset), ImageNet-22K (a superset of ImageNet-1K) \cite{deng2009imagenet}, ImageNet-100 (a subset of ImageNet-1K as proposed in \cite{tian2020cmc}), and iNaturalist18 \cite{iNaturalist} are also employed. Table \ref{tab:datasets_pretext} details these datasets. The effectiveness of the representations learned through SSL is typically evaluated using linear evaluation and fine-tuning techniques, following the same protocol as \cite{kornblith2019do, zhai2020large}.

\begin{table}[!t]
\centering
\caption{Source datasets commonly used for self-supervised pretext task training and evaluation.}
\label{tab:datasets_pretext}
\begin{tabular}{lcc}
\hline
Dataset & Classes & Images (train/validation) \\ 
\hline
ImageNet-1K \cite{deng2009imagenet} & 1,000 & 1,281,167 / 50,000 \\ 
ImageNet-100 \cite{tian2020cmc} & 100 & 130,000 / 5,000 \\ 
ImageNet-22K \cite{deng2009imagenet} & 21,841 & 14,197,122 / 50,000 \\ 
Instagram-1B \cite{yalniz2019instagram} & - & 1,000,000,000 / - \\
iNaturalist18 \cite{iNaturalist} & 8,142 & 437,513 / 24,426 \\ 
CIFAR-10 \cite{krizhevsky2009cifar} & 10 & 50,000 / 10,000 \\ 
CIFAR-100 \cite{krizhevsky2009cifar} & 100 & 50,000 / 10,000 \\ 
Places205 \cite{zhou2018places365} & 205 & 2,500,000 / 20,000 \\ 
\hline
\end{tabular}
\end{table}

\subsubsection{Comparative Results: Linear Evaluation on ImageNet-1K}

We collected and compiled performance results of SSL methods in linear evaluation tasks using a ResNet-50 \cite{he2015resnet} encoder for fair comparison. Table \ref{tab:lineval} shows the top-1 and top-5 accuracies for various SSL methods under linear evaluation on ImageNet-1K, while Table \ref{tab:lineval_multicrop} shows methods that employ multi-crop augmentation. We present methods employing multi-crop augmentation separately, as it is fair to do so due to the performance boost it offers, albeit at the cost of much longer training. Notably, \xazomara{ReLIC-v2} achieves the highest top-1 accuracy of 77.1\% and a top-5 accuracy of 93.3\%, demonstrating its strong performance in extracting useful representations. This result is attributed to the innovative approach of combining an explicit invariance loss with a contrastive objective over a varied set of appropriately constructed data views, as proposed by \xazomara{ReLIC}. This method effectively avoids learning spurious correlations and obtains more informative representations, outperforming previous self-supervised approaches by significant margins. \xazomara{CLSA} follows closely with a top-1 accuracy of 76.2\%, benefiting from the strong augmentations it employs. \xazomara{AdCo} and \xazomara{C-BYOL} also perform well with top-1 accuracies of 75.7\% and 75.6\%, respectively. These results indicate that methods such as \xazomara{ReLIC-v2} and \xazomara{CLSA} are particularly effective in learning high-quality representations that generalize well in a linear evaluation setup.

\begin{table}[!thb]
\centering
\footnotesize
\caption{Performance comparison of various SSL methods under linear evaluation protocol on ImageNet-1K using ResNet-50 backbone.}
\label{tab:lineval}
\begin{tabular}{lccccc}
\hline
Method & Projector & Batch size & Epoch & Top-1 & Top-5 \\
\hline
\rowcolor{gray!20} \textit{Supervised} & None & 256 & 90 & 76.5 & 93.0 \\
\xazomara{AdCo} \cite{hu2021adco} & MLP & 256 & 800 & 72.8 & - \\
\xazomara{All4One} \cite{estepa2023all4one} & MLP & 1024 & 1000 & 66.6 & 87.5 \\
\xazomara{ARB} \cite{zhang2022arb} & MLP & 2048 & 100 & 68.2 & 88.9 \\
\xazomara{Barlow Twins} \cite{zbontar2021barlowtwins} & MLP & 2048 & 1000 & 73.2 & 91.0 \\
\xazomara{BYOL} \cite{grill2020byol} & MLP & 4096 & 1000 & 74.3 & 91.6 \\
\xazomara{CaCo} \cite{wang2022caco} & MLP & 4096 & 800 & 75.3 & - \\
\xazomara{C-BYOL} \cite{lee2021cbyolcsimclr} & MLP & 4096 & 1000 & 75.6 & 92.7 \\
\xazomara{CLSA} \cite{wang2022clsa} & MLP & 256 & 800 & 72.2 & - \\
\xazomara{CMC} \cite{tian2020cmc} & Linear & 128 & 200 & 60.0 & 82.3 \\
\xazomara{CoKe} \cite{qian2022coke} & MLP & 256 & 800 & 71.4 & - \\
\xazomara{CPC-v2} \cite{henaff2020cpcv2} & MLP & 512 & 200 & 63.8 & 85.3 \\
\xazomara{C-SimCLR} \cite{lee2021cbyolcsimclr} & MLP & 4096 & 1000 & 71.6 & 90.5 \\
\xazomara{DeepCluster} \cite{caron2019deepcluster} & MLP & 256 & 400 & 52.2 & - \\
\xazomara{InfoMin} \cite{tian2020infomin} & MLP & 256 & 800 & 73.0 & 91.1 \\
\xazomara{InstDis} \cite{wu2018instdis} & Linear & 256 & 200 & 54.0 & - \\
\xazomara{InstDis++} \cite{misra2019pirl} & Linear & 256 & 800 & 59.0 & - \\
\xazomara{Jigsaw} \cite{noroozi2017unsupervised} & None & 256 & 90 & 45.7 & - \\
\xazomara{LA} \cite{zhuang2019la} & Linear & 256 & 200 & 60.2 & - \\
\xazomara{Mixed Barlow Twins} \cite{bandara2023mixedbarlowtwins} & MLP & 256 & 300 & 72.2 & - \\
\xazomara{MMCL} \cite{shah2021mmcl} & MLP & 256 & 200 & 63.8 & - \\
\xazomara{MoCHI} \cite{kalantidis2020mochi} & MLP & 256 & 800 & 68.7 & - \\
\xazomara{MoCLR} \cite{tian2021dnc} & MLP & 4096 & 1000 & 74.3 & 92.2 \\
\xazomara{MoCo} \cite{he2020moco} & Linear & 256 & 200 & 60.6 & - \\
\xazomara{MoCo-v2} \cite{chen2020mocov2} & MLP & 512 & 800 & 71.1 & - \\
\xazomara{MoCo-v2 + DCL} \cite{yeh2022dcl} & MLP & 256 & 200 & 67.6 & - \\
\xazomara{MoCo-v3} \cite{chen2021mocov3} & MLP & 4096 & 800 & 73.8 & - \\
\xazomara{MSF} \cite{koohpayegani2021msf} & MLP & 256 & 200 & 72.4 & - \\
\xazomara{NNSiam} \cite{dwibedi2021nnclr} & MLP & 256 & 4096 & 72.6 & 90.4 \\
\xazomara{OBoW} \cite{gidaris2021obow} & MLP & 256 & 200 & 73.8 & - \\
\xazomara{ODC} \cite{zhan2020odc} & MLP & 256 & 400 & 55.7 & - \\
\xazomara{PCL} \cite{li2021pcl} & None & 256 & 200 & 61.5 & - \\
\xazomara{PCL-v2} \cite{li2021pcl} & Linear & 256 & 200 & 67.6 & - \\
\xazomara{PIRL} \cite{misra2019pirl} & Linear & 1024 & 800 & 63.6 & - \\
\xazomara{ReLIC} \cite{mitrovic2020relic} & MLP & 4096 & 1000 & 70.3 & 89.5 \\
\xazomara{ReLIC-v2} \cite{tomasev2022relicv2} & MLP & 256 & 1000 & \textbf{77.1} & 93.3 \\
\xazomara{ReSSL} \cite{zheng2021ressl} & MLP & 4096 & 200 & 69.9 & - \\
\xazomara{RotNet} \cite{gidaris2018rotnet} & None & 256 & 256 & 48.9 & - \\
\xazomara{SeLa} \cite{asano2020sela} & Linear & 256 & 400 & 61.5 & 84.0 \\
\xazomara{Self-Classifier} \cite{amrani2022selfclassifier} & MLP & 4096 & 800 & 74.1 & - \\
\xazomara{SimCLR} \cite{chen2020simclr} & MLP & 4096 & 1000 & 69.3 & 89.0 \\
\xazomara{SimCLR + DCL} \cite{yeh2022dcl} & MLP & 256 & 200 & 65.8 & - \\
\xazomara{SimCLR + DCLW} \cite{yeh2022dcl} & MLP & 256 & 200 & 66.9 & - \\
\xazomara{SimCLR-v2} \cite{chen2020simclrv2} & MLP & 4096 & 800 & 71.7 & 90.4 \\
\xazomara{SimSiam} \cite{chen2020simsiam} & MLP & 256 & 800 & 71.3 & - \\
\xazomara{SMoG} \cite{pang2022smog} & MLP & 4096 & 800 & \underline{74.5} & 91.9 \\
\xazomara{SynCo} \cite{giakoumoglou2024synco} & MLP & 256 & 800 & 70.7 & - \\
\xazomara{TiCo} \cite{zhu2022tico} & MLP & 4096 & 1000 & 73.4 & 91.6 \\
\xazomara{TWIST} \cite{wang2021twist} & MLP & 4096 & 800 & 72.6 & 91.0 \\
\xazomara{UniGrad} \cite{tao2022unigrad} & MLP & 4096 & 100 & 70.3 & - \\
\xazomara{UniVIP} \cite{li2022univip} & MLP & 4096 & 300 & 74.2 & - \\
\xazomara{VICReg} \cite{bardes2022vicreg} & MLP & 4096 & 1000 & 73.2 & 91.1 \\
\xazomara{VICRegL} \cite{bardes2022vicregl} & MLP & 256 & 300 & 71.2 & - \\
\xazomara{W-MSE} \cite{ermolov2021wmse} & MLP & 4096 & 400 & 72.5 & - \\
\hline
\end{tabular}
\end{table}

\begin{table}[!t]
\centering
\caption{Performance comparison of SSL methods employing multi-crop augmentation under linear evaluation protocol on ImageNet-1K using ResNet-50 backbone.}
\label{tab:lineval_multicrop}
\begin{tabular}{lccccc}
\hline
Method & Projector & Batch size & Epoch & Top-1 & Top-5 \\
\hline
\rowcolor{gray!20} \textit{Supervised} & None & 256 & 90 & 76.5 & 93.0 \\
\xazomara{AdCo} \cite{hu2021adco} & MLP & 256 & 800 & 72.8 & - \\
\xazomara{AdCo} \cite{hu2021adco} & MLP & 1024 & 800 & 75.7 & - \\
\xazomara{CaCo} \cite{wang2022caco} & MLP & 4096 & 800 & 75.7 & - \\
\xazomara{CLSA} \cite{wang2022clsa} & MLP & 256 & 800 & \underline{76.2} & - \\
\xazomara{DeepCluster-v2} \cite{caron2020swav} & MLP & 256 & 400 & 74.3 & - \\
\xazomara{DINO} \cite{caron2021dino} & MLP & 4096 & 300 & 75.3 & - \\
\xazomara{MMCR} \cite{yerxa2023mmcr} & MLP & 256 & 100 & 72.1 & - \\
\xazomara{NNCLR} \cite{dwibedi2021nnclr} & MLP & 4096 & 1000 & 75.6 & 92.4 \\
\xazomara{ReSSL} \cite{zheng2021ressl} & MLP & 4096 & 200 & 74.7 & - \\
\xazomara{SeLa-v2} \cite{caron2020swav} & MLP & 256 & 400 & 71.8 & - \\
\xazomara{SMoG} \cite{pang2022smog} & MLP & 4096 & 400 & \textbf{76.4} & 93.1 \\
\xazomara{SwAV} \cite{caron2020swav} & MLP & 4096 & 800 & 75.3 & - \\
\xazomara{UniGrad} \cite{tao2022unigrad} & MLP & 4096 & 800 & 75.5 & - \\
\hline
\end{tabular}
\end{table}
\newpage

\subsubsection{Comparative Results: Semi-supervised Evaluation on ImageNet-1K}

Top-1 and top-5 accuracy are key evaluation metrics in machine learning classification tasks, and are used for performance comparison of SSL methods herein. Top-1 accuracy measures the percentage of test samples where the model's highest probability prediction matches the correct class, representing the conventional definition of accuracy. For example, classifying a cat image as "tiger" would be incorrect under top-1 accuracy. In contrast, top-5 accuracy measures the percentage of samples where the correct class appears among the model's five highest probability predictions, allowing for some flexibility in classification. This metric is particularly useful for problems with many classes, scenarios with multiple valid interpretations, recommendation systems, and complex datasets like ImageNet. Top-5 accuracy is always equal to or higher than top-1 accuracy, and the gap between these metrics can provide insights into model behaviour, potentially indicating that the model is learning relevant features but struggling with final discrimination between similar classes.

In the following, we gathered performance results of SSL methods in semi-supervised tasks using a ResNet-50 encoder for fair comparison. Table \ref{tab:semisupervised} presents the performance achieved when fine-tuning representations on classification tasks with small subsets (1\% and 10\%) of ImageNet-1K's training set. The results demonstrate that \xazomara{SimCLR-v2} (+KD) achieves the highest top-1 accuracy—73.9\% with just 1\% of labels and 91.5\% with 10\% of labels—significantly outperforming other methods. This superior performance can be attributed to the knowledge distillation technique employed in \xazomara{SimCLR-v2}, though it should be noted that this approach is technically classified as semi-supervised rather than purely self-supervised. \xazomara{UniGrad} follows closely behind, excelling with a top-5 accuracy of 83.8\% using 1\% of the labeled set and a top-1 accuracy of 71.5\% with 10\% of labels, respectively. \xazomara{NNCLR} also demonstrates competitive performance, particularly when using 10\% of labels, where it achieves a top-1 accuracy of 69.8\%.

\begin{table}[!t]
\centering
\caption{Performance comparison of SSL methods under semi-supervised fine-tuning protocol using limited labeled data from ImageNet-1K (1\% and 10\% of labels) with ResNet-50 backbone.}
\label{tab:semisupervised}
\begin{tabular}{lccccc}
\hline
\multirow{2}{*}{Method} & \multicolumn{2}{c}{{1\%}} & \multicolumn{2}{c}{{10\%}} \\
 & {Top-1} & {Top-5} & {Top-1} & {Top-5} \\ \hline
\rowcolor{gray!20} \textit{Supervised} & 25.4 & 48.4 & 56.4 & 80.4 \\
\xazomara{All4One} \cite{estepa2023all4one} & 39.0 & - & 60.1 & - \\
\xazomara{Barlow Twins} \cite{zbontar2021barlowtwins} & 55.0 & 79.2 & 69.7 & 89.3 \\
\xazomara{BYOL} \cite{grill2020byol} & 53.2 & 78.4 & 68.8 & 89.0 \\
\xazomara{C-BYOL} \cite{lee2021cbyolcsimclr} & 60.6 & 70.5 & \textbf{83.4} & 90.0 \\
\xazomara{C-SimCLR} \cite{lee2021cbyolcsimclr} & 51.1 & 77.2 & 64.5 & 86.5 \\
\xazomara{InstDis} \cite{wu2018instdis} & - & 39.2 & - & 77.4 \\
\xazomara{MMCR} (2 views) \cite{yerxa2023mmcr} & 46.6 & - & 63.9 & - \\
\xazomara{MMCR} (8 views) \cite{yerxa2023mmcr} & 51.0 & - & 67.7 & - \\
\xazomara{MoCHI} \cite{giakoumoglou2024synco} & 50.4 & 76.2 & 65.7 & 87.2 \\
\xazomara{MSF} \cite{koohpayegani2021msf} & 55.5 & 79.9 & 66.5 & 87.6 \\
\xazomara{NNCLR} \cite{dwibedi2021nnclr} & 56.4 & 80.7 & 69.8 & 89.3 \\
\xazomara{PCL} \cite{li2021pcl} & - & 75.3 & - & 85.6 \\
\xazomara{PCL-v2} \cite{li2021pcl} & - & 73.9 & - & 85.0 \\
\xazomara{PIRL} \cite{misra2019pirl} & 30.7 & 57.2 & 60.4 & 83.8 \\
\xazomara{ReLIC-v2} \cite{tomasev2022relicv2} & 58.1 & 72.4 & \underline{81.3} & \underline{91.2} \\
\xazomara{SimCLR} \cite{chen2020simclr} & 48.3 & 75.5 & 65.6 & 87.8 \\
\xazomara{SimCLR-v2} (+KD) \cite{chen2020simclrv2} & \textbf{73.9} & \textbf{91.5} & 77.5 & \textbf{93.4} \\
\xazomara{SMoG} \cite{pang2022smog} & 58.0 & 81.6 & 71.2 & 90.5 \\
\xazomara{SwAV} \cite{caron2020swav} & 53.9 & 78.5 & 70.2 & 89.9 \\
\xazomara{SynCo} \cite{giakoumoglou2024synco} & 50.8 & 77.5 & 66.6 & 88.0 \\
\xazomara{TiCo} \cite{buettner2022insloc} & 53.0 & 79.2 & 66.8 & 88.0 \\
\xazomara{UniGrad} \cite{tao2022unigrad} & \underline{60.8} & \underline{83.8} & 71.5 & 90.6 \\
\xazomara{UniVIP} \cite{li2022univip} & 53.0 & 78.8 & 67.1 & 88.5 \\
\xazomara{VICReg} \cite{bardes2022vicreg} & 54.8 & 79.4 & 69.5 & 89.5 \\
\hline
\end{tabular}
\end{table}

The comparison of SSL techniques across linear and semi-supervised evaluation frameworks reveals a consistent pattern in performance results. Methods that excel in linear evaluation generally demonstrate robust performance in semi-supervised settings as well, validating the effectiveness of their learned representations in label-constrained environments. For instance, \xazomara{ReLIC-v2}, which leads in linear evaluation with a top-1 accuracy of 77.1\%, also exhibits strong semi-supervised performance, achieving a notable top-1 accuracy of 81.3\% with 10\% of the labels. Similarly, \xazomara{SimCLR-v2}, though technically classified as semi-supervised due to its knowledge distillation approach, outperforms other methods in its category, suggesting that strategies enhancing performance in linear evaluation settings transfer effectively to semi-supervised contexts. This correlation underscores the potential of SSL techniques not only in leveraging unlabeled data but also in efficiently utilizing small amounts of labeled data, presenting a compelling case for their broader applicative value in realistic machine learning scenarios. Such consistency is crucial for validating the robustness of SSL methods and confirming their suitability across diverse real-world applications.

\subsection{Transfer Learning Capabilities to Other Classification Tasks}
\label{sec:tlcapabilities}

Transfer learning evaluation differs from linear evaluation and semi-supervised fine-tuning on the source dataset by specifically measuring how well SSL-pretrained models generalize to new domains with semantically distinct data distributions and visual characteristics. Unlike linear evaluation on the source dataset, which freezes the encoder and trains only a classifier on the same dataset distribution, transfer learning requires adaptation to domain shifts through more extensive fine-tuning strategies, including domain-specific augmentations and architectural modifications. While evaluations on the source dataset have become relatively standardized, transfer learning protocols remain more fragmented, though efforts like \cite{wang2022usb} aim to standardize cross-domain evaluation. Furthermore, transfer learning evaluation more directly assesses real-world utility in specialized domains with scarce labeled data, revealing whether SSL methods learn truly generalizable features or merely dataset-specific patterns. Specifically, the features learned on ImageNet-1K are examined to determine their generality and usefulness across different image domains, and to assess whether they are inherently ImageNet-specific.

\subsubsection{Evaluation Protocols}

In transfer learning evaluation for computer vision SSL, the linear evaluation protocol assesses representation quality by freezing the pretrained encoder and training only a linear classifier on labeled target data. This process typically employs L-BFGS optimization with $\ell_2$-regularization, as the convex nature of linear classification makes this approach efficient and stable. Training images undergo resizing to $256 \times 256$, random resized cropping to $224 \times 224$, horizontal flipping, and normalization, while validation images undergo resizing to $256 \times 256$, center cropping to $224 \times 224$, and normalization. The pipeline includes training with minimal hyperparameter tuning. This standardized approach enables direct comparison of representation quality across SSL methods while requiring minimal computational resources.

Semi-supervised fine-tuning evaluation, in contrast, updates some or all encoder weights alongside the classifier using limited labeled data and unlabeled target samples. This protocol employs SGD with Nesterov momentum (0.9) or AdamW (for ViTs or low-memory scenarios) with layer-wise learning rates, complemented by weight decay ($\ell_2$-regularization) and consistency losses for pseudo-labeled data. The pipeline incorporates both weak augmentations (random crops/flips) and strong augmentations (e.g., RandAugment) with domain-specific adjustments, while the momentum parameter for batch normalization statistics is set to $\max(1-10/s, 0.9)$ where $s$ denotes the number of steps per epoch. This approach achieves higher absolute performance than linear evaluation at the cost of increased computational requirements and hyperparameter sensitivity, making it more reflective of real-world deployment scenarios.

\subsubsection{Datasets}

Transfer learning for classification is performed on the following datasets: Food-101 \cite{bossard2014food101}, CIFAR-10 \cite{krizhevsky2009cifar} and CIFAR-100 \cite{krizhevsky2009cifar}, Birdsnap \cite{berg2014birdsnap}, the SUN397 scene dataset \cite{xiao2010sun397}, Stanford Cars \cite{krause2013cars}, FGVC Aircraft \cite{maji2013aircraft}, the PASCAL VOC 2007 classification task \cite{everingham2009voc}, the Describable Textures Dataset (DTD) \cite{cimpoi2014dtd}, Oxford-IIIT Pets \cite{parkhi2012pets}, Caltech-101 \cite{feifei2007caltech101}, and Oxford 102 Flowers \cite{nilsback2008flowers102}. Table \ref{tab:datasets_down1} presents details of these datasets. After linear evaluation or fine-tuning, performance is reported using standard metrics for each benchmark, and results are provided on a held-out test set after hyperparameter selection on a validation set. In the following subsection, we adopt the methodologies outlined in the respective dataset publications to report top-1 accuracy for Food-101, CIFAR-10, CIFAR-100, Birdsnap, SUN397, Stanford Cars, and DTD, and mean per-class accuracy for FGVC Aircraft, Oxford-IIIT Pets, Caltech-101, and Oxford 102 Flowers.

\begin{table}[!t]
\centering
\caption{Standard benchmark datasets used for evaluating transfer learning capabilities of self-supervised models on diverse classification tasks (following Kornblith et al. \cite{kornblith2019do} evaluation protocol).}
\begin{tabular}{lcc}
\hline
Dataset & Classes & Images (train/test) \\ \hline
Food-101 \cite{bossard2014food101} & 101 & 101,000 (75,750/25,250) \\
CIFAR-10 \cite{krizhevsky2009cifar} & 10 & 60,000 (50,000/10,000) \\
CIFAR-100 \cite{krizhevsky2009cifar} & 100 & 60,000 (50,000/10,000)  \\
Birdsnap \cite{berg2014birdsnap} & 500 & 49,829 (47,386/2,443) \\
SUN397 \cite{xiao2010sun397} & 397 & 39,700 (19,850/19,850)  \\
StanfordCars \cite{krause2013cars} & 196 & 16,185 (8,144/8,041)  \\
FGVC Aircraft \cite{maji2013aircraft} & 100 & 10,000 (6,667/3,333) \\
VOC 07 \cite{everingham2009voc} & 20 & 9,963 (5,011/4,952) \\
DTD \cite{cimpoi2014dtd} & 47 & 5,640 (3,760/1,880)  \\
Oxford-IIIT Pets \cite{parkhi2012pets} & 37 & 7,049 (3,680/3,369)  \\
Caltech-101 \cite{feifei2007caltech101} & 101 & 9,144 (3,060/6,084)  \\
Oxford 102 Flowers \cite{nilsback2008flowers102} & 102 & 8,189 (2,040/6,149)  \\ \hline
\end{tabular}
\label{tab:datasets_down1}
\end{table}

\subsubsection{Comparative Results}

In the following, we compiled transfer learning evaluation results for several SSL methods, all using a ResNet-50 encoder. Table \ref{tab:tl_1} presents the results for various target datasets with the encoder frozen (linear evaluation), and Table \ref{tab:tl_2} shows the results with the encoder fine-tuned.

Table \ref{tab:tl_1} shows the performance of SSL methods under linear evaluation across several classification datasets. Supervised refers to models pre-trained on ImageNet using labeled data with direct class supervision. \xazomara{ReLIC-v2} stands out, achieving the highest accuracies on datasets such as Food-101 (80.6\%), SUN397 (66.2\%), and Aircraft (64.8\%), indicating that \xazomara{ReLIC-v2} learns representations that generalize well to diverse image domains. \xazomara{NNCLR} also performs exceptionally well, leading on CIFAR-10 (93.7\%), CIFAR-100 (79.0\%), and Birdsnap (61.4\%). \xazomara{BYOL} demonstrates strong results on various datasets, with notable scores on Pets (90.4\%) and Flowers (96.1\%). These results highlight the effectiveness of \xazomara{ReLIC-v2}, \xazomara{NNCLR}, and \xazomara{BYOL} in learning high-quality representations that perform well in linear evaluation.

\begin{sidewaystable}[!thbp]
\centering
\caption{Transfer learning performance of various SSL methods on diverse classification datasets using features from ImageNet-1K pretrained ResNet-50 under linear evaluation protocol (frozen backbone).}
\label{tab:tl_1}
\begin{tabular}{lcccccccccccc}
\hline
Method & \rotatebox{0}{Food-101} & \rotatebox{0}{CIFAR-10} & \rotatebox{0}{CIFAR-100} & \rotatebox{0}{Birdsnap} & \rotatebox{0}{SUN397} & \rotatebox{0}{Cars} & \rotatebox{0}{Aircraft} & \rotatebox{0}{VOC2007} & \rotatebox{0}{DTD} & \rotatebox{0}{Pets} & \rotatebox{0}{Caltech-101} & \rotatebox{0}{Flowers} \\
\hline
\rowcolor{gray!20} \textit{Supervised} & 72.3 & 93.6 & 78.3 & 53.7 & 61.9 & 66.7 & 61.0 & {82.8} & 74.9 & 91.5 & 94.5 & 94.7 \\
\xazomara{BYOL} \cite{grill2020byol} & 75.3 & 93.1 & 78.4 & 57.2 & 62.2 & 67.8 & 60.6 & 82.5 & 75.5 & 90.4 & 94.2 & \textbf{96.1} \\
\xazomara{SimCLR} \cite{chen2020simclr} & 68.4 & 90.6 & 71.6 & 37.4 & \textbf{65.8} & 50.3 & 50.3 & 80.5 & 74.5 & 83.6 & 90.3 & 91.2 \\
\xazomara{NNCLR} \cite{dwibedi2021nnclr} & \textbf{76.7} & \textbf{93.7} & \textbf{79.0} & \textbf{61.4} & 62.5 & 67.1 & \textbf{64.1} & 83.0 & 75.5 & 91.8 & 91.3 & 95.1 \\
\xazomara{C-SimCLR} \cite{lee2021cbyolcsimclr} & 73.0 & 91.6 & 75.2 & 38.2 & 62.3 & 52.7 & 53.5 & - & 73.0 & 84.0 & 91.2 & 89.0 \\
\xazomara{ReLIC-v2} \cite{tomasev2022relicv2} & 80.6 & 92.8 & 78.2 & 65.4 & 66.2 & \textbf{75.1} & 64.8 & - & \textbf{77.4} & \textbf{92.4} & 92.8 & 95.6 \\
\xazomara{MSF} \cite{koohpayegani2021msf} & 71.2 & 92.6 & 76.3 & - & 59.2 & 55.6 & 53.7 & - & 73.2 & 88.7 & 92.7 & 92.0 \\
\xazomara{TWIST} \cite{wang2021twist} & 91.2 & 74.4 & 66.8 & - & 66.8 & 55.2 & 53.6 & \textbf{85.7} & 76.6 & 91.6 & \textbf{94.5} & 93.4 \\
\hline
\end{tabular}
\end{sidewaystable}

Table \ref{tab:tl_2} presents the results when the encoder is fine-tuned. Supervised learning refers to models pre-trained on ImageNet using labeled data with direct class supervision, serving as a strong baseline. Random init (initialization) refers to models with randomly initialized weights and no pre-training. \xazomara{TWIST} achieves the highest accuracies on several datasets, including Food-101 (89.3\%), CIFAR-10 (97.9\%), and Birdsnap (76.7\%). \xazomara{ReLIC-v2} also performs well, showing leading results on Food-101 (88.7\%), Cars (92.3\%), and Aircraft (88.7\%). \xazomara{BYOL} continues to demonstrate strong performance, particularly on Pets (91.7\%) and Aircraft (88.1\%). These observations indicate that \xazomara{TWIST}, \xazomara{ReLIC-v2}, and \xazomara{BYOL} are highly effective in transfer learning tasks, achieving high accuracy across various datasets when fine-tuned.

\begin{sidewaystable}[!thbp]
\centering
\caption{Transfer learning performance of various SSL methods on diverse classification datasets using fine-tuning of ImageNet-1K pretrained ResNet-50 models.}
\label{tab:tl_2}
\begin{tabular}{lcccccccccccc}
\hline
Method & \rotatebox{0}{Food-101} & \rotatebox{0}{CIFAR-10} & \rotatebox{0}{CIFAR-100} & \rotatebox{0}{Birdsnap} & \rotatebox{0}{SUN397} & \rotatebox{0}{Cars} & \rotatebox{0}{Aircraft} & \rotatebox{0}{VOC2007} & \rotatebox{0}{DTD} & \rotatebox{0}{Pets} & \rotatebox{0}{Caltech-101} & \rotatebox{0}{Flowers} \\
\hline
\rowcolor{gray!20} \textit{Supervised} & 88.3 & 97.5 & 86.4 & 75.8 & {64.3} & 92.1 & 86.0 & 85.0 & 74.6 & 92.1 & 93.3 & 97.6 \\
\rowcolor{gray!20} \textit{Random init.} & 86.9 & 95.9 & 80.2 & \textbf{76.1} & 53.6 & 91.4 & 85.9 & 67.3 & 64.8 & 81.5 & 72.6 & 92.0 \\
\xazomara{BYOL} \cite{grill2020byol} & 88.5 & 97.8 & 86.1 & 76.3 & 63.7 & 91.6 & \textbf{88.1} & 85.4 & 76.2 & 91.7 & 83.8 & 97.0 \\
\xazomara{SimCLR} \cite{chen2020simclr} & 88.2 & 97.7 & 85.9 & 75.9 & 63.5 & 91.3 & 88.1 & 84.1 & 73.2 & 89.2 & 92.1 & 97.0 \\
\xazomara{MMCL} \cite{shah2021mmcl} & 82.4 & 96.2 & 82.1 & - & - & 89.2 & 85.4 & - & 73.5 & - & 87.8 & 95.2 \\
\xazomara{ReLIC-v2} \cite{tomasev2022relicv2} & 88.7 & 97.7 & 85.3 & 76.7 & 64.7 & \textbf{92.3} & 88.7 & - & \textbf{76.9} & 92.2 & 93.2 & \textbf{97.9} \\
\xazomara{TWIST} \cite{wang2021twist} & \textbf{89.3} & \textbf{97.9} & \textbf{86.5} & - & \textbf{67.4} & 91.9 & 85.7 & \textbf{86.5} & 76.4 & \textbf{94.5} & \textbf{93.5} & 97.1 \\
\hline
\end{tabular}
\end{sidewaystable}

The varying performance of SSL methods across vision tasks can be attributed to how different techniques capture and prioritize features during training. Methods like \xazomara{ReLIC-v2} and \xazomara{BYOL}, which excel at capturing high-level semantic features, tend to perform well on diverse datasets requiring a broad understanding of object categories and visual concepts, such as Food-101 and Stanford Cars. On the other hand, methods like \xazomara{NNCLR}, which may emphasize finer textural or shape details, excel on datasets like CIFAR-10 and Birdsnap, where such attributes are crucial for classification.

The performance variation also suggests that no single SSL approach is universally superior across all image domains, reflecting the specialized capabilities of different learning mechanisms. Datasets with fine-grained categories, like FGVC Aircraft and Oxford 102 Flowers, may benefit more from methods that capture intricate details and subtle within-category variations. Conversely, tasks requiring robustness to changes in background, lighting, or pose might favour methods good at generalizing from comprehensive scene understandings, such as those trained on SUN397 or VOC 2007.

\subsection{Transfer Learning Capabilities to Other Vision Tasks}
\label{sec:tlcapabilities2}

This evaluation assesses the generalization of representations beyond classification tasks, using various datasets and tasks, including but not limited to object detection, semantic segmentation, instance segmentation, and depth estimation. Object detection identifies and localizes objects in an image by drawing bounding boxes and assigning class labels. Semantic segmentation assigns a class label to every pixel in an image, creating a mask where all pixels of the same class share the same label. Instance segmentation, a subset of image segmentation, goes further by distinguishing individual instances of the same class, producing separate masks for each object. Depth estimation predicts the distance between the camera and points in a scene, generating a depth map that represents the 3D structure. 
The evaluation pipeline typically involves pretraining an encoder on unlabeled data (e.g., ImageNet-1K) using SSL objectives, integrating it into task-specific architectures (e.g., Faster R-CNN \cite{ren2016fasterrcnn} for detection, DeepLabv3+ \cite{chen2017deeplabv3} for semantic segmentation), fine-tuning on labeled target datasets, and measuring performance using task-specific metrics like Average Precision (AP) for detection and mean Intersection over Union (mIoU) for semantic segmentation, or Root Mean Square Error (RMSE) for depth estimation. Detailed evaluation protocols for each task, including target datasets and comparative results, are provided in the following subsections.

\subsubsection{Evaluation Protocols and Datasets}

In SSL transfer learning evaluation, ResNet-50 (hereby referred as R50) serves as the standard backbone architecture, offering a consistent framework for comparing various SSL methods across downstream tasks. This 50-layer deep convolutional neural network, featuring residual blocks and skip connections to address the vanishing gradient problem, provides a reliable basis for assessing representation quality. Performance metrics such as linear probing accuracy, fine-tuning results on object detection and segmentation, and transfer learning efficiency to different domains are commonly reported using R50 as the base encoder, enabling fair comparisons of SSL techniques. The widespread adoption of R50 in computer vision research, due to its robust design and ability to train very deep networks effectively, has solidified its position as a crucial benchmark in the field.

For object detection evaluation, the Faster R-CNN \cite{ren2016fasterrcnn} detector (a two-stage object detection algorithm that uses a Region Proposal Network (RPN) and convolutional neural networks to identify and locate objects in images, providing bounding box predictions and class labels) with either frozen or fine-tuned representations, is employed. Backbone choices include R50-C4, R50-DC5, or R50-FPN, representing, respectively, the fourth stage output of ResNet-50, dilated convolutions at the fifth stage, and a feature pyramid network built on ResNet-50. This setup is implemented in Detectron2 \cite{wu2019detectron2}, an open-source computer vision library providing a flexible framework for state-of-the-art object detection and segmentation, including pre-trained models and efficient training pipelines. Object detection performance is evaluated on the VOC 2012 \cite{everingham2009voc} and COCO \cite{li2015coco} datasets using Faster R-CNN for bounding box predictions and Mask R-CNN \cite{he2018maskrcnn} for both object detection and instance segmentation, with Mask R-CNN extending Faster R-CNN by adding a branch for predicting pixel-level segmentation masks for each detected object.

\begin{table}[!t]
\centering
\caption{Datasets used for evaluating transfer learning capabilities of self-supervised models on complex vision tasks beyond classification, including object detection, instance segmentation, semantic segmentation, and depth estimation.}
\begin{tabular}{lccc}
\hline
Dataset & Task & Classes & Images (train/test) \\ \hline
VOC 07 \cite{everingham2009voc} & Object detection & 21 & 5,011 / 4,952 \\ 
VOC 07+12 \cite{everingham2009voc} & Object detection & 21 & 16,551 / 4,952 \\ 
COCO \cite{li2015coco} & Object detection/Segmentation & 80 & 118,287 / 5,000 \\ 
LVIS v0.5 \cite{gupta2019lvis} & Instance segmentation & 1,203 & 100,170 / - \\
ADE20K \cite{zhou2018ade20k} & Semantic segmentation & 150 & 20,210 / 2,000 \\ 
VOC 2012 \cite{everingham2009voc} & Semantic segmentation & 21 & 11,530 / 1,449 \\
Cityscapes \cite{cordts2016cityscapes} & Semantic segmentation & 30 & 2,975 / 500 \\ 
NYU v2 \cite{silberman2012nyu} & Depth estimation & - & 24,000 / 654 \\
\hline
\end{tabular}
\label{tab:datasets_down2}
\end{table}

For instance segmentation, the Mask R-CNN \cite{he2018maskrcnn} detector with frozen or fine-tuned representations is used, with the same backbone options: R50-C4, R50-DC5, or R50-FPN. This setup is also implemented in Detectron2 \cite{wu2019detectron2}. Instance segmentation is assessed on the COCO \cite{li2015coco}, LVIS v0.5 \cite{gupta2019lvis}, and Cityscapes \cite{cordts2016cityscapes} datasets, mirroring the approach in MoCo \cite{he2020moco}. For semantic segmentation, a fully-convolutional network (FCN)-based \cite{long2015fcn} architecture as in MoCo \cite{he2020moco} is used. The backbone consists of a pretrained ResNet-50. The FCN is trained using a standard per-pixel softmax CE loss. Semantic segmentation is evaluated on the VOC 2012 \cite{everingham2009voc}, Cityscapes \cite{cordts2016cityscapes}, and ADE20K \cite{zhou2018ade20k} datasets, using methodologies from prior research such as \cite{long2015fcn}. Finally, for depth estimation, the protocol from \cite{laina2016deep} is followed, utilizing a standard ResNet-50 backbone. The \xazomara{conv5} features are processed through four fast up-projection blocks with filter sizes of 512, 256, 128, and 64. Training is performed using a reverse Huber loss function, as specified in \cite{laina2016deep} and depth estimation is evaluated on the NYU v2 dataset \cite{silberman2012nyu}. Table \ref{tab:datasets_down2} summarizes details of the aforementioned datasets.

\subsubsection{Comparative Results}

We collected performance evaluation results of several SSL methods on transfer learning tasks for object detection and instance segmentation using a ResNet-50 backbone for fair comparison. The results are summarized in Table \ref{tab:tl_det_seg}. For detection tasks Faster R-CNN is employed, while Mask R-CNN is used for segmentation. AP (Average Precision), AP$_{50}$, and AP$_{75}$ are key metrics for evaluating SSL in object detection/instance segmentation. AP is the mean average precision calculated over multiple Intersection over Union (IoU) thresholds, typically from 0.5 to 0.95 in 0.05 increments, providing a comprehensive assessment of detection/segmentation quality. AP$_{50}$ specifically measures the average precision at an IoU threshold of 0.5 (50\% overlap), offering a more lenient evaluation that's particularly useful for assessing performance on smaller or harder-to-detect objects. AP$_{75}$, with its higher IoU threshold of 0.75, demands more precise localization and is thus a stricter measure of accuracy. These metrics collectively offer a nuanced view of an SSL model's ability to transfer learned representations to detection and instance segmentation tasks, balancing between coarse object localization (AP$_{50}$) and more precise boundary delineation (AP$_{75}$), with AP providing an overall performance summary.

Among the evaluated methods, \xazomara{SoCo} and \xazomara{InsCon} achieve state-of-the-art performance on both VOC and COCO benchmarks. On VOC detection, \xazomara{InsCon} leads with the highest scores across all metrics (AP = 59.1, AP$_{50}$ = 83.6, AP$_{75}$ = 66.6), demonstrating its effectiveness in object recognition and boundary localization. For COCO detection, \xazomara{SoCo} outperforms all methods with AP$^{bb}$ = 40.4, AP$^{bb}_{50}$ = 60.4, and AP$^{bb}_{75}$ = 43.7, while \xazomara{InsCon} follows closely in second place. On COCO segmentation, \xazomara{InsCon} achieves the top performance with AP$^{msk}$ = 35.1 and AP$^{msk}_{75}$ = 37.6, while \xazomara{SoCo} leads in AP$^{msk}_{50}$ = 56.8. These results highlight how methods focusing on object-level representations (\xazomara{SoCo}) and multi-instance consistency (\xazomara{InsCon}) significantly outperform both supervised baselines and earlier self-supervised approaches.

\begin{table*}[!t]
\centering
\setlength{\tabcolsep}{1mm}
\caption{Transfer learning performance of self-supervised methods on object detection and instance segmentation tasks using ResNet-50-C4 backbone. Results are reported on PASCAL VOC for detection and COCO for both detection and segmentation using standard Average Precision metrics (AP, AP\(_{50}\), AP\(_{75}\)). Detection metrics are denoted by \(bb\) (bounding box) and segmentation by \(msk\) (mask). All methods used 200-epoch pretraining except where noted in parentheses.}
\label{tab:tl_det_seg}
\begin{tabular}{lccccccccc}
\hline
\multirow{2}{*}{Method}  & \multicolumn{3}{c}{VOC detection} & \multicolumn{3}{c}{COCO detection} & \multicolumn{3}{c}{COCO segmentation} \\ 
 \cmidrule(lr){2-4}  \cmidrule(lr){5-10}
 & AP & AP$_{50}$ & AP$_{75}$ & $AP^{bb}$ & $AP^{bb}_{50}$ & $AP^{bb}_{75}$ & $AP^{msk}$ & $AP^{msk}_{50}$ & $AP^{msk}_{75}$ \\
\hline
\rowcolor{gray!20} \textit{Supervised} & 53.5 & 81.3 & 58.8 & 38.2 & 58.2 & 41.2 & 33.3 & 54.7 & 35.2 \\
\rowcolor{gray!20} \textit{Random init} & 33.8 & 60.2 & 33.1 & 26.4 & 44.0 & 27.8 & 29.3 & 46.9 & 30.8 \\
\xazomara{Barlow Twins} \cite{zbontar2021barlowtwins} & 56.8 & 82.6 & 63.4 & 39.2 & 59.0 & 42.5 & 34.3 & 56.0 & 36.5 \\
\xazomara{BYOL} \cite{grill2020byol} & 51.9 & 81.0 & 56.5 & - & - & - & - & - & - \\
\xazomara{DenseCL} \cite{wang2021densecl} & \underline{58.7} & 82.8 & 65.2 & - & - & - & - & - & - \\
\xazomara{DetCo} \cite{xie2021detco} & 57.8 & 82.6 & 64.2 & 39.8 & 59.7 & 43.0 & 34.7 & 56.3 & 36.7 \\
\xazomara{InfoMin} \cite{tian2020infomin} & 57.6 & 82.7 & 64.6 & 39.0 & 58.5 & 42.0 & 34.1 & 55.2 & 36.3 \\
\xazomara{InsCon} \cite{yang2022inscon} & \textbf{59.1} & \textbf{83.6} & \textbf{66.6} & \underline{40.3} & \underline{60.0} & \underline{43.5} & \textbf{35.1} & \underline{56.7} & \textbf{37.6} \\
\xazomara{InsLoc} \cite{yang2021insloc} & 57.9 & 82.9 & 64.9 & 39.5 & 59.1 & 42.7 & 34.5 & 56.0 & 36.8 \\
\xazomara{InstDis} \cite{wu2018instdis} & 55.2 & 80.9 & 61.2 & 37.7 & 57.0 & 40.9 & 33.0 & 54.1 & 35.2 \\
\xazomara{MoCHI} \cite{kalantidis2020mochi} & 57.5 & 82.7 & 64.4 & 39.2 & 58.9 & 42.4 & 34.3 & 55.5 & 36.6 \\
\xazomara{MoCo} \cite{he2020moco} & 55.9 & 81.5 & 62.6 & 38.5 & 58.3 & 41.6 & 33.6 & 54.8 & 35.6 \\
\xazomara{MoCo-v2} \cite{chen2020mocov2} & 57.0 & 82.4 & 63.6 & 38.9 & 58.6 & 41.9 & 34.1 & 55.5 & 36.0 \\
\xazomara{PIRL} \cite{misra2019pirl} & 55.5 & 81.0 & 61.3 & 37.4 & 56.5 & 40.2 & 32.7 & 53.4 & 34.7 \\
\xazomara{ReSim} \cite{xiao2021resim} & \underline{58.7} & \underline{83.1} & \underline{66.3} & 39.7 & 59.0 & 43.0 & 34.6 & 55.9 & \underline{37.1} \\
\xazomara{SimCLR} \cite{chen2020simclr} & 56.3 & 81.9 & 62.5 & - & - & - & - & - & - \\
\xazomara{SimSiam} \cite{chen2020simsiam} & 57.0 & 82.4 & 63.7 & 39.2 & 59.3 & 42.1 & 34.4 & 56.0 & 36.7 \\
\xazomara{SoCo} \cite{wei2021soco} & \textbf{59.1} & 83.4 & 65.6 & \textbf{40.4} & \textbf{60.4} & \textbf{43.7} & \underline{34.9} & \textbf{56.8} & 37.0 \\
\xazomara{SwAV} \cite{caron2020swav} & 56.1 & 82.6 & 62.7 & 38.4 & 58.6 & 41.3 & 33.8 & 55.2 & 35.9 \\
\xazomara{SynCo} \cite{giakoumoglou2024synco} & 57.2 & 82.6 & 63.9 & 39.9 & 59.6 & 43.3 & \underline{34.9} & 56.5 & 36.9 \\
\hline
\end{tabular}
\end{table*}

\section{Discussion}
\label{sec:discussion}

This section delves into a comprehensive discussion of fundamental challenges inherent in discriminative SSL and their impact on the efficacy of learned representations. We begin with an analysis of the architectural evolution of SSL frameworks (Section~\ref{sec:architecture}), exploring how base networks, pretext tasks, and training objectives have transformed over time. Next, we examine the fundamental challenges facing discriminative SSL methods (Section~\ref{sec:challenges}), including representation collapse, dimensionality reduction, and hyperparameter sensitivity. We then address the intrinsic trade-offs encountered within SSL (Section~\ref{sec:tradeoffs}), exploring the tensions between seemingly desirable properties such as invariance-equivariance and practical limitations of scalability and efficiency. The discussion continues with critical considerations for benchmarking and evaluation protocols (Section~\ref{sec:benchmarking_eval}), including limitations of current benchmarks and assessment of robustness. Finally, we examine the theoretical foundations underpinning SSL methods (Section~\ref{sec:theory}), highlighting both formal guarantees and gaps in our understanding. The primary objective of these in-depth analyses is to foster a deeper understanding for the reader, identify prevailing trends, and propose promising avenues for future research.

\subsection{Evolution of Architectural Frameworks}
\label{sec:architecture}

\begin{figure*}[!t]
\centering
\includegraphics[width=0.9\linewidth]{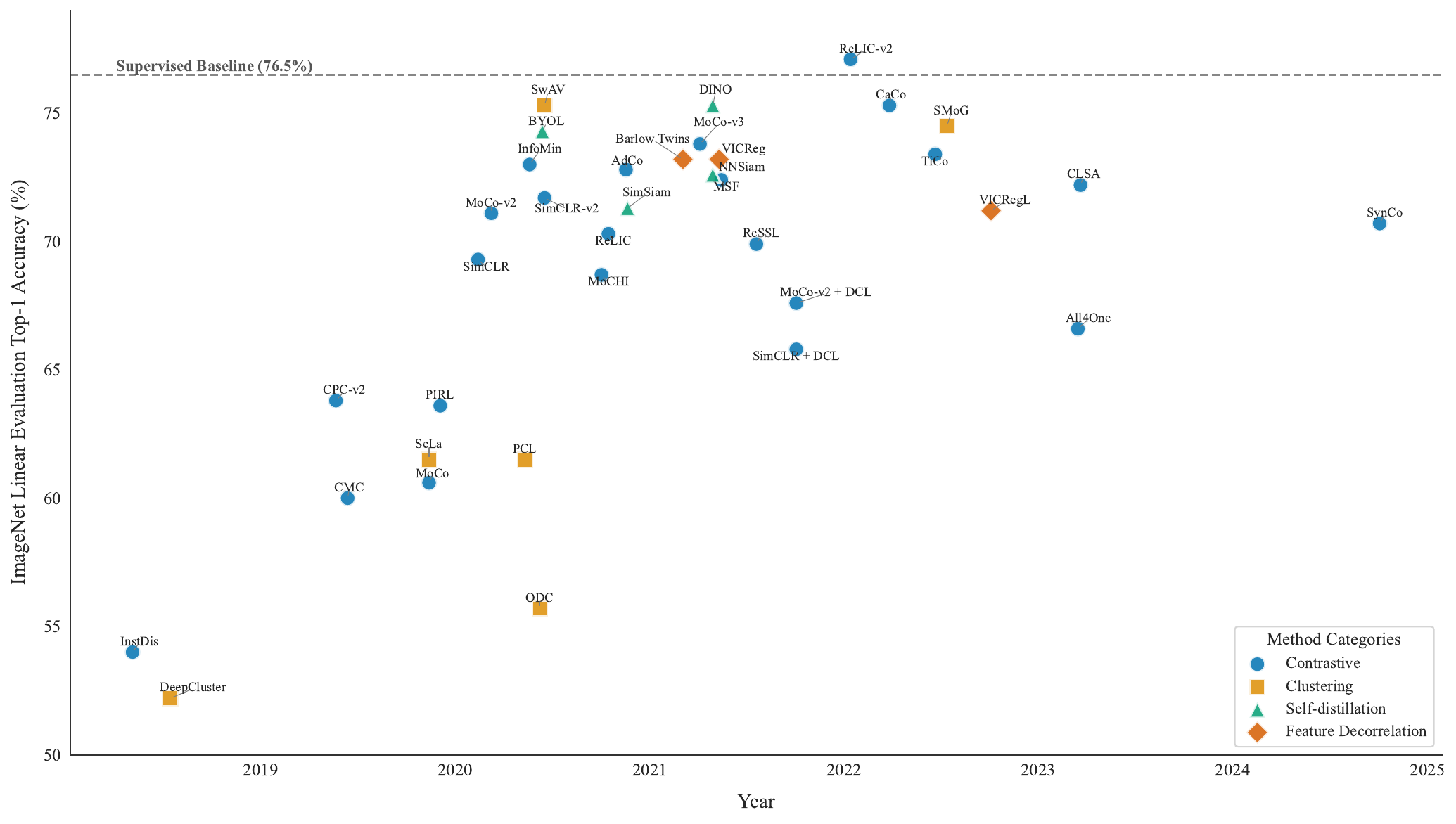}
\caption{Evolution of SSL performance on ImageNet-1K linear evaluation using ResNet-50 encoder across different methods and years (2017-2025).}
\label{fig:ssl_methods_comparison}
\end{figure*}

The evolution of SSL methods in computer vision has shown significant progress in recent years. As shown in Figure \ref{fig:ssl_methods_comparison}, there is a clear upward trend in top-1 accuracy on ImageNet-1K, showing the steady improvement of SSL techniques. This progress reflects the field's development, with recent methods like \malakia|ReLIC-v2| even surpassing the supervised learning baseline of 76.5\%, an important milestone that highlights the potential of learning without labels. Examining contrastive learning frameworks reveals a clear pattern of development. Early methods like \malakia|MoCo| used simple projection heads, while later approaches such as \malakia|SimCLR| and \malakia|MoCo-v3| used more complex MLPs with more hidden layers. The use of asymmetric architectures, introduced by \malakia|BYOL|, has also been very effective. This trend suggests that architectural complexity correlates with performance improvements, following a \textit{"bigger is better"} principle. 

This upward trend shifts after 2023, where the top-1 accuracy begins to decrease slightly. This change occurs because newer methods focus on different objectives rather than just improving ImageNet accuracy. Recent approaches address specific limitations in previous methods (e.g., \malakia|SynCo|), combine SSL techniques with transformer architectures (e.g., \malakia|All4One|), or optimize for specific downstream tasks beyond classification (e.g., \malakia|VICRegL| for object detection). These developments show how the field is expanding beyond benchmark performance to solve more practical problems.

However, current frameworks appear to be approaching a performance bottleneck, indicating the need to revisit their fundamental limitations rather than simply scaling them further. For example, a significant challenge in contrastive learning is the diminishing contribution of negative samples as training progresses \cite{kalantidis2020mochi,giakoumoglou2024synco}. Several approaches have attempted to address this limitation, including methods like \malakia|MoCHI| and \malakia|SynCo| that generate synthetic hard negatives, and \malakia|DCL| that modifies the InfoNCE loss to decouple negative and positive pairs. This highlights the importance of identifying and addressing specific weaknesses in existing frameworks rather than solely developing entirely new paradigms. In this section, we summarize the architectural framework evolution of SSL methods in terms of the base network, the shift in pretext tasks, and the evolution of training objectives and loss functions, and discuss how each of these aspects contributes to the advancement of the field.

\subsubsection{Base Network Evolution}

While ResNet-50 serves as a consistent baseline for fair comparisons, the pursuit of better performance has led to the adoption of more advanced encoder architectures like larger ResNet variants and Vision Transformers, actively exploring and adopting advanced architectures to enhance SSL capabilities. More specifically, the evolution of base networks in discriminative SSL progressed from ResNet-dominated frameworks (2018–2020)—where methods like \malakia|MoCo| (2019), \malakia|SimCLR| (2020), and \malakia|SwAV| (2020) leveraged ResNet-50’s hierarchical features for contrastive and clustering tasks—to intermediate architectures like Wide ResNets and deeper hybrids (e.g., ResNet-50 $\times2$ in \malakia|MoCo-v2|), which enhanced feature capacity and stability. \malakia|ReLIC-v2| (2022) further optimized ResNet backbones (up to ResNet-200) with multi-view invariance and saliency masking, achieving 80.6$\%$ ImageNet-1K accuracy and surpassing supervised baselines, demonstrating that advanced SSL objectives could revitalize CNNs even as ViTs gained prominence. 

Following the introduction of ViTs, hierarchical architectures like EsViT (2022) integrated CNN-like multi-stage processing with self-attention, achieving 81.3$\%$ accuracy. While lightweight CNNs prioritized efficiency, ViTs, exemplified by \malakia|DINO| (2021) with its ViT-Small and self-distillation (78.2$\%$ accuracy), demonstrated superior performance post-2021. \malakia|Barlow Twins| (2021) also refined ResNet training through projection heads, further illustrating the evolving landscape. However, the direct adaptation of ViT architectures for SSL proved challenging, necessitating specific modifications for stable training. For instance, \malakia|MoCo-v3| freezes the initial patch embedding layer to mitigate gradient spikes, while \malakia|MoBY| employs asymmetric drop path rates for its encoders. The distinct global dependency capture of ViTs' self-attention, compared to CNNs' hierarchical feature extraction, suggests the potential for novel self-supervision forms. These challenges underscore the need for SSL techniques tailored to the unique characteristics of transformer architectures, rather than mere adaptations of CNN-based methods. Conversely, the success of \malakia|ReLIC-v2| highlighted that architectural refinements and training protocols, such as saliency-aware augmentation, can bridge the ResNet-ViT performance gap, providing a robust, CNN-compatible alternative to ViT-driven methods like \malakia|DINO|. This trajectory reflects a dual focus: scaling ViTs for global context capture while optimizing CNNs through SSL innovations, ensuring the continued viability of both architectures in discriminative SSL.

\subsubsection{Shift in Pretext Tasks}

The evolution of pretext tasks in SSL for computer vision transitioned from diverse early approaches (including colorization \cite{zhang2016colorization}, inpainting \cite{pathak2016context}, jigsaw puzzles \cite{noroozi2017unsupervised}, and rotation prediction \cite{gidaris2018rotnet}) to data augmentation-dominated frameworks, reflecting the field's prioritization of invariance-driven objectives. Early pretext tasks required models to solve spatial-contextual challenges (e.g., predicting relative patch positions or restoring color to grayscale images), which fostered rich but task-specific representations. However, the rise of contrastive learning (e.g., \malakia|SimCLR|, \malakia|MoCo|) shifted focus to augmentation-based pretext tasks, where models learn invariance to transformations like cropping, flipping, or color jittering \cite{chen2020simclr}. While this shift improved scalability and alignment with downstream tasks, it risked narrowing the diversity of learned features by over-relying on augmentation-induced invariance. Despite this trend, methods like \malakia|CPC| \cite{oord2019cpc}, which leverage temporal or spatial predictive coding, remain relevant for capturing complementary structural patterns, underscoring the importance of maintaining methodological diversity in SSL. This progression highlights a tension between standardization for efficiency and the preservation of task-agnostic representation breadth.

Recent advancements (2023–present) introduced automated pretext task design, exemplified by evolutionary augmentation policies (e.g., \malakia|EAPO| \cite{barrett2023evolutionary}) optimizing data transformations via genetic algorithms, and multi-task hybrids combining contrastive learning with spatial reasoning (e.g., rotation and contrastive objectives). These innovations improved performance across domains: \malakia|PIRL| boosted PASCAL VOC detection by 9$\%$, \cite{barrett2023evolutionary} elevated CIFAR-10 accuracy by 12$\%$, and hybrid tasks enhanced ImageNet linear probing by 7$\%$. Applications expanded to medical imaging (e.g., patch reconstruction for tumor detection) and video analysis (frame-order prediction for action recognition), reflecting a shift from manual heuristic design to systematic, optimization-driven approaches that rival supervised learning in label-scarce settings.

\subsubsection{Evolution of Training Objectives and Loss Functions}

The evolution of training objectives in discriminative SSL reflects a deliberate shift from heuristic designs to mathematically grounded principles. Early contrastive methods like \malakia|SimCLR| relied on NCE, which explicitly compared positive pairs against negatives to maximize mutual information. While effective, this approach faced scalability challenges due to its dependence on large negative sample batches and sensitivity to augmentation strategies. This limitation spurred the development of non-contrastive paradigms like \malakia|BYOL|, which eliminated negatives entirely by enforcing consistency between a student network and an EMA teacher. The success of such methods revealed that feature collapse could be prevented through architectural mechanisms (e.g., predictor networks) rather than explicit contrastive terms.

Concurrently, feature decorrelation objectives emerged as a unifying framework across families. Methods like \malakia|VICReg| and \malakia|Barlow Twins| replaced instance discrimination with covariance matrix constraints, explicitly minimizing redundancy between feature dimensions. This shift aligned with theoretical insights showing that dimensional collapse – not just complete collapse – hindered representation quality. Hybrid approaches soon followed: \malakia|DINO| combined self-distillation with centering and sharpening operations to stabilize training, while \malakia|MSN| integrated MIM with clustering objectives. These innovations demonstrated that multi-objective loss functions could balance invariance (to augmentations) with variance (across semantic content), addressing earlier trade-offs between representation robustness and discriminability.

Future advancements in training objectives will likely focus on integrating causal reasoning and task-adaptive dynamics to bridge the gap between pretraining and downstream applications. Inspired by insights from dimensional collapse analysis, next-generation loss functions may explicitly model causal feature disentanglement, penalizing spurious correlations induced by augmentation strategies while preserving invariant semantic factors. For instance, objectives could incorporate counterfactual terms that simulate interventions on nuisance variables (e.g., lighting, occlusions), fostering more robust feature spaces. Simultaneously, meta-learned loss functions might emerge, where the objective itself is optimized during training to balance competing goals like invariance, decorrelation, and task-specific discriminability. This could build on frameworks like hypernetworks or gradient-based meta-learning, dynamically adapting loss landscapes to dataset statistics. Additionally, scalable multi-objective hybrids may unify contrastive, distillation, and decorrelation mechanisms through learnable weightings, avoiding the manual tuning seen in current hybrid SSL pipelines. Such systems could automatically prioritize feature diversity in early training before shifting to semantic alignment, mimicking curriculum learning principles. These directions aim to move beyond static, handcrafted losses toward adaptive, theoretically coherent objectives that generalize across domains.

\subsubsection{Beyond Contrastive and Clustering Methods}

While contrastive learning and clustering approaches have largely defined the SSL landscape, other methods deserve more attention. Self-distillation techniques, such as \malakia|BYOL| and \malakia|SimSiam|, and feature decorrelation methods, including \malakia|Barlow Twins|, \malakia|VICReg|, and \malakia|ReLIC|, offer useful alternatives that have not been explored as much. In particular, regularization-based methods like \malakia|VICReg| and \malakia|ReLIC| show promise for future work. This aligns with ideas from {Yann LeCun}, who has suggested focusing more on regularization techniques in representation learning. However, contrastive methods often suffer from a {"semantic repulsion"} problem \cite{giakoumoglou2024rrd}, where semantically similar instances are forced apart even when they should maintain similarity. These regularization approaches could address this issue, along with other limitations of contrastive methods, such as the need for many negative samples or the risk of losing valuable information. By setting direct constraints on the statistical properties of representations, they may improve robustness and generalizability. Future research should examine these regularization methods closely, aiming to create new objectives that ensure invariance to irrelevant transformations while keeping information relevant to downstream tasks. 

\subsection{Fundamental Challenges in Discriminative Self-supervised Learning}
\label{sec:challenges}

The preceding sections highlighted that SSL methods, while demonstrably successful, are susceptible to fundamental challenges that can impede their efficacy and the fidelity of learned representations. These challenges include representation collapse, dimensionality collapse, semantic repulsion, early training degeneration, and acute hyperparameter sensitivity. A thorough understanding of the interrelationships among these issues and their manifestation across diverse SSL methods is essential for advancing the field and developing robust, reliable techniques. This subsection provides a comprehensive overview of these interrelationships, identifying the specific vulnerabilities of each SSL method family and exploring potential inherent immunities or effective mitigation strategies.

\subsubsection{Representation Collapse}

Representation collapse, also known as system collapse, represents a critical failure mode in SSL where the learned representations become trivial and fail to capture any meaningful information from the input data \cite{jing2022directclr,he2024or}. This often manifests as the model mapping all input instances to the same constant vector or to a very limited subspace of the representation space. Such a scenario renders the learned representations unusable for downstream tasks because they lack the discriminative power to distinguish between different inputs. One of the primary causes of representation collapse is a poorly designed loss function that does not incorporate sufficient constraints to prevent the model from converging to these trivial solutions. For instance, if the objective is solely to maximize the similarity between different augmented views of the same input, without any mechanism to ensure diversity in the representations of different inputs, the model might simply learn to output the same constant vector for all inputs, thereby achieving maximal similarity within positive pairs but losing all discriminative information. Representation collapse can be considered an extreme form of dimensionality collapse, where the representation space effectively shrinks to a single point or a very low-dimensional space. Most SSL methods incorporate specific design elements to avoid this fundamental failure, but the strategies employed can sometimes lead to other challenges, such as dimensionality collapse in a less extreme form or an over-reliance on certain aspects of the training process. 

The field has developed a diverse array of techniques to mitigate this degenerate solution. For example, methods like \malakia|DINO| employ sharpening and centering mechanisms, \malakia|BYOL| and \malakia|MoCo| utilize momentum encoders to maintain a slowly evolving representation of past states, while \malakia|SwAV| introduces swapping assignments between views. Feature decorrelation methods like \malakia|VICReg| and \malakia|ReLIC| implement explicit regularization terms to ensure diverse representations.

\subsubsection{Dimensionality Collapse}

Dimensionality collapse describes a scenario where learned representations, though not entirely constant, reside within a lower-dimensional subspace of the model's embedding space, occurring when only a few dimensions capture significant data variance, leaving many others unused or redundant, often evidenced by a covariance matrix with dominant eigenvalues indicating variance concentration along few principal components. This collapse restricts the model's ability to encode comprehensive data information, potentially hindering performance on tasks requiring detailed distinctions, and can manifest not only in final representations but also in hidden features and encoder weight matrices, where it leads to self-related, redundant filters that limit network expressiveness. However, a degree of dimensionality reduction can be beneficial for separability, as models naturally focus on discriminative features, and recent findings suggest it might be an emergent property for informative representation learning; thus, the objective is not to eliminate reduction entirely, but to ensure the resulting subspace maintains sufficient information for downstream tasks.

Various techniques have been proposed to mitigate dimensionality collapse. \malakia|DirectCLR| addresses dimensional collapse by eliminating the trainable projector and directly optimizing a subvector of the representation space, preserving the full dimensionality of the embeddings. \malakia|OR| applies orthogonality regularization, enforcing orthogonal weight matrices to prevent dimensional collapse across weights, hidden features, and representations, ensuring filter diversity and uniform eigenvalue distributions, targeting both convolutional and linear layers during pretraining. By ensuring that the weight matrices are orthogonal, \malakia|OR| helps to stabilize the distribution of their outputs, thereby preventing the dimensional collapse of hidden features and final representations. Feature whitening (used in \malakia|Barlow Twins| and \malakia|VICReg|) is another technique that aims to decorrelate the features. However, some evidence suggests that feature whitening might only address the issue at the feature level and might not prevent or could even exacerbate collapse in the weight matrices. This diversity of approaches reflects the challenge’s significance and the field’s ingenuity in addressing it. The necessity of these "{tricks}" underscores that representation collapse remains a fundamental theoretical issue in SSL that warrants continued attention.

\subsubsection{Semantic Repulsion}

Semantic repulsion is a counterintuitive outcome where learned representations, instead of grouping semantically similar instances, push them apart, a phenomenon arising from imbalances or flaws in the learning objective or training process, particularly in contrastive and clustering-based methods intended to do the opposite. Potential causes include overly aggressive negative sampling that treats semantically related instances as negative pairs, or poor initial cluster assignments and refinement in clustering methods that misclassify similar instances, leading to distant representations. This repulsion detrimentally affects representation quality, hindering downstream tasks reliant on semantic similarity, such as image retrieval and semantic classification, by distorting the representation space's reflection of underlying semantic relationships; mitigating this requires careful design of SSL objectives to balance attraction and repulsion of instances, and robust mechanisms for identifying and grouping semantically related data points.

\subsubsection{Early Degeneration}

Early degeneration describes the premature plateauing or decline of model performance, measured by downstream task representation quality, hindering the model's ability to learn optimal, generalizable representations. This issue arises from factors like convergence to suboptimal local minima, overfitting to pretext tasks, and problematic training dynamics, such as imbalances in positive and negative signals or optimization flaws, including the uniformity-tolerance dilemma in contrastive learning or teacher model stagnation in self-distillation. Consequently, learned representations become less informative and robust, leading to diminished downstream task performance compared to potential outcomes with extended or stable training; mitigating early degeneration necessitates meticulous hyperparameter tuning, exploration of alternative optimization strategies, and adjustments to SSL objectives to promote sustained learning and improved generalization.

\subsubsection{Hyperparameter Sensitivity}

Data augmentations are fundamental to discriminative SSL, enabling models to learn invariant features and generalize by creating multiple transformed views of input data, thereby focusing on essential semantics; however, SSL performance is highly sensitive to the choice, type, and strength of these augmentations, where appropriate augmentations significantly improve performance while poorly chosen or overly aggressive ones can degrade learned representations by losing critical information or learning spurious invariances, and conversely, weak augmentations may fail to provide sufficient diversity, leading to trivial feature learning. This sensitivity varies across SSL families, with contrastive learning being particularly susceptible due to its reliance on positive and negative pair creation, and recent research challenging the label-preserving assumption by exploring label-destroying augmentations as feature dropout. Notably, large datasets may reduce the necessity for complex augmentations, revealing a complex interaction between data scale and augmentation strategies; consequently, there is growing interest in adaptive selection or learning of optimal augmentations tailored to specific tasks and datasets to mitigate this sensitivity.

\subsubsection{Addressing the Challenges}

Contrastive methods excel in self-supervised representation learning by discerning between similar and dissimilar data pairs, mitigating complete representation collapse through loss functions that maximize inter-instance distance, often reliant on numerous negative samples. Despite this, dimensionality collapse remains a concern, necessitating techniques like projector networks, feature decorrelation, or orthogonality constraints. Furthermore, the definition of negative pairs can induce semantic repulsion, pushing semantically related instances apart. Early degeneration may stem from the uniformity-tolerance dilemma or suboptimal negative sample selection. Finally, performance is highly sensitive to augmentation choices and hyperparameters, such as temperature and learning rate.

Clustering-based SSL groups similar data points, avoiding collapse via overclustering or multi-view prediction tasks. Although the clustering objective can encourage dimensional usage, dimensionality collapse remains a possibility. Semantic repulsion is less likely when clustering accurately reflects semantic similarity, but poor initialization or training instability can separate related instances. Early degeneration is a risk if clustering converges prematurely. The method's effectiveness hinges on hyperparameters related to the clustering algorithm and the overall training process.

Self-distillation, exemplified by \malakia|BYOL| and \malakia|SimSiam|, trains a student network to predict a teacher's output, eschewing negative samples and relying on asymmetric architectures and momentum encoders to prevent collapse. While dimensionality collapse remains a potential issue, the prediction task can promote dimensional utilization. Semantic coherence in the teacher's representations can mitigate semantic repulsion, though early degeneration can occur if the teacher stagnates or the student struggles to learn. Performance is contingent upon hyperparameters, particularly learning rate and momentum.

Knowledge distillation in SSL transfers knowledge from a teacher to a student, leveraging the teacher's collapse-avoidance capabilities. The student's susceptibility to dimensionality collapse is influenced by the teacher's representation space. Semantic repulsion can be inherited from the teacher or arise during student training. Early degeneration in the student can result from ineffective knowledge transfer or limited capacity. Both teacher and student training, as well as the distillation process, are sensitive to hyperparameter tuning.

Feature decorrelation methods explicitly combat collapse and dimensionality reduction by promoting statistical independence among representation dimensions, often via loss function penalties. While effective against collapse, care must be taken to avoid inadvertently inducing semantic repulsion. Early degeneration remains a concern, requiring robust training strategies. Performance is influenced by hyperparameters related to decorrelation strength and broader training parameters.

\subsection{Trade-offs}
\label{sec:tradeoffs}

In the following we delve into the inherent trade-offs encountered in SSL, exploring the fundamental tensions between seemingly desirable properties and practical limitations. We begin by examining the invariance-equivariance spectrum in representation learning, followed by a discussion on the interplay between scalability and efficiency in SSL methodologies. Finally, we address the trade-offs that arise when optimizing SSL methods for specific downstream tasks.

\subsubsection{Invariance-Equivariance Trade-off}

In the realm of representation learning, particularly within SSL, there exists a fundamental trade-off between learning representations that are invariant to certain transformations and those that are equivariant to them. Invariance refers to the property where a representation remains unchanged despite the input undergoing a specific transformation, such as rotation or cropping. This is often desirable for high-level tasks like image classification, where the category of an object should be recognized regardless of its orientation or position. Equivariance, on the other hand, means that the representation transforms in a predictable way that corresponds to the transformation applied to the input. This property is often more beneficial for tasks that require spatial awareness, such as object detection or semantic segmentation, where the location and spatial relationships of objects are crucial.  

Different families of SSL methods handle this trade-off in various ways. Contrastive learning, for example, often aims to learn representations that are invariant to the data augmentations used during training. By encouraging different augmented views of the same image to have similar representations, contrastive methods implicitly promote invariance to these transformations. However, this pursuit of invariance can sometimes lead to the loss of information that might be valuable for certain downstream tasks. Recognizing this limitation, recent research has explored ways to incorporate equivariance into SSL frameworks, often resulting in methods termed "contrastive-equivariant self-supervised learning". These approaches aim to learn representations that are both robust to certain variations (invariant) while also preserving information about specific transformations (equivariant).  

Clustering-based SSL methods typically aim for invariance within the learned clusters, meaning that instances belonging to the same semantic cluster should have similar representations regardless of certain transformations. However, the definition of these clusters and the features used for clustering can influence the degree of invariance or equivariance captured. Feature decorrelation methods often focus on learning representations where the different dimensions are statistically independent, which can implicitly affect the invariance or equivariance properties of the learned features. Recent research continues to explore methods that can learn both invariant and equivariant features simultaneously or adaptively, based on the requirements of the specific downstream task. This highlights the ongoing effort to find the right balance in the invariance-equivariance spectrum for different applications of SSL.

\subsubsection{Scalability and Efficiency}

Scalability and efficiency in discriminative SSL methods exhibit both complementary dynamics and inherent trade-offs, shaped by architectural choices and optimization strategies. Scalability—the ability to maintain performance gains as model size or data volume increases—often conflicts with efficiency, which prioritizes reduced computational costs (e.g., memory, training time). Early contrastive methods like \malakia|SimCLR| achieved scalability through large batch sizes but suffered inefficiency due to quadratic memory costs from negative samples. This spurred innovations like \malakia|MoCo|’s memory-efficient queue-based negatives and \malakia|BYOL|’s elimination of negatives via self-distillation, decoupling scalability from batch size constraints. ViTs further altered this balance: while their self-attention mechanisms scale poorly with input resolution, techniques like patch-based contrastive learning and gradient checkpointing enabled efficient training on high-resolution images. Recent trends show a shift toward non-contrastive paradigms (e.g., \malakia|VICReg|, \malakia|DINO|) that avoid explicit negative comparisons, achieving scalability through feature decorrelation or self-distillation objectives while maintaining sublinear computational growth.

Families differ markedly in their scalability-efficiency profiles. Contrastive methods remain computationally heavy for large datasets but benefit from parallelizable architectures. Self-distillation (e.g., \malakia|DINO|) and feature decorrelation families (e.g., \malakia|Barlow Twins|) excel in efficiency by eliminating negative samples and simplifying loss calculations, though they may require longer training cycles to stabilize. Clustering-based methods like \malakia|SwAV| balance both dimensions through online cluster assignments but face diminishing returns on extremely heterogeneous data. The future lies in adaptive hybrid frameworks: methods like \malakia|MSN| combine masked image modeling with clustering to dynamically allocate compute resources. As shown in \malakia|DINO-v2|’s scaling to 7B parameters, ViT-based SSL models will likely dominate, leveraging sparse attention mechanisms and mixed-precision training to reconcile scalability with manageable compute budgets

\subsubsection{Trade-offs in Task-Specific Optimization}

Recent years have witnessed an increasing focus on task-specific SSL methods, particularly for dense prediction tasks like object detection and segmentation. Methods such as \malakia|DenseCL|, \malakia|InsLoc|, and \malakia|SoCo| have achieved strong results on these tasks, often outperforming supervised pretraining. However, this specialization comes with trade-offs. As evidenced by methods like \malakia|ReLIC-v2| and \malakia|DINO-v2|, optimizing for dense representation learning can marginally reduce classification accuracy while significantly boosting performance on localization tasks. This highlights a fundamental tension in SSL: {learning local features often comes at the expense of global features}, creating a situation where improving performance on one downstream task may degrade performance on another. Future research should seek to develop methods that better balance these competing objectives.

\subsection{Benchmarking and Evaluation}
\label{sec:benchmarking_eval}

Evaluation protocols for discriminative SSL in computer vision have shifted from simplistic in-domain metrics (e.g., linear probing on ImageNet) to comprehensive frameworks assessing generalization, robustness, and architectural adaptability. Early protocols prioritized linear separability but overlooked hyperparameter sensitivity and domain shifts. Modern benchmarks now evaluate out-of-distribution performance across diverse datasets, quantify invariance-variance trade-offs, and measure few-shot adaptability. Standardization efforts introduced input normalization (e.g., batch norm before linear evaluation) and stratified hyperparameter testing to reduce inconsistencies between SSL families. Evaluations increasingly prioritize real-world viability, testing on niche industrial tasks (e.g., medical imaging) and isolating backbone contributions (e.g., ViTs vs. CNNs). Future directions emphasize multi-modal zero-shot transfer (e.g., \malakia|CLIP|-style alignment) and causal robustness to inference-time perturbations, reflecting a maturation toward practical applicability over narrow benchmark optimization.

The subsequent discussion will focus on critical considerations pertaining to the evaluation and practical application of SSL methods. We delve into the limitations of current benchmarking practices and the crucial aspect of data diversity, followed by an analysis of influential SSL frameworks. We then address the often-overlooked dimensions of robustness and out-of-distribution generalization, provide a brief comparison with the burgeoning field of masked image modeling, and conclude by discussing the significant practical constraints related to computational accessibility.

\subsubsection{Benchmark Limitations and Data Diversity}

The reliance on ImageNet as the primary benchmark for SSL raises important questions about the field's direction. While ImageNet provides a standardized evaluation platform, it is a meticulously curated, balanced dataset that may not reflect the challenges of learning from truly uncurated unblanaced data. The aspiration of SSL is to leverage vast amounts of unfiltered, imbalanced, and potentially noisy data, yet most methods are demonstrated on carefully selected datasets. Future research should explore benchmarks that better reflect real-world data distributions, including significant class imbalance, noisy labels, and greater domain diversity. This would provide a more realistic assessment of SSL methods' ability to extract useful information from imperfect data sources. Additionally, transparency in pretraining datasets is essential for fair comparison, as some commercial entities utilize private datasets that cannot be replicated by the broader research community.

\subsubsection{Evaluating Self-supervised Learning Frameworks}

When evaluating \textit{"what is the best self-supervised learning framework?"}, no single answer emerges. Different methods perform well in different contexts: some achieve superior classification performance, while others demonstrate strengths in detection or segmentation tasks. However, in terms of practical impact and adoption, \malakia|MoCo-v2|, \malakia|SwAV|, and \malakia|DINO| stand out as particularly influential frameworks for representation learning. \malakia|MoCo| introduced key innovations in contrastive learning, \malakia|SwAV| introduced effective clustering-based approaches, and \malakia|DINO| demonstrated strong performance with emerging properties like unsupervised segmentation. These frameworks have been widely adapted for domain-specific applications, including medical imaging and related fields.

\subsubsection{Robustness and Out-of-Distribution Evaluation}

A notable gap in current SSL evaluation protocols is the limited assessment of robustness and out-of-distribution (OOD) generalization. The primary objective of SSL is to create representations from large, uncurated datasets that generalize well to diverse scenarios. However, most evaluations focus on performance on well-curated, balanced datasets like ImageNet. This disconnect between the stated goal and evaluation practice represents a significant limitation in current research. Only a few methods, such as \malakia|SynCo|, explicitly address robustness to distribution shifts and adversarial attacks. A more systematic evaluation framework that includes diverse OOD datasets and robustness metrics would provide a more comprehensive assessment of SSL methods' real-world utility. This would also help determine whether learned representations are truly generic or merely domain-specific.

\subsubsection{Comparative Analysis with Masked Image Modeling}

Although beyond the scope of this review, it is worth briefly discussing the relationship between discriminative SSL methods and generative approaches like MIM. Methods like \malakia|BEiT| \cite{bao2022beit} and Masked Autoencoders (MAEs) \cite{he2021mae} have gained prominence, particularly with ViT architectures. While generative approaches typically require end-to-end fine-tuning for downstream tasks, discriminative SSL methods often perform well with linear probing. Generative methods like MAE can be more computationally efficient during pretraining (requiring only one forward pass instead of two), but their linear probing performance generally lags behind contrastive methods. Combining contrastive objectives with generative approaches (as in \malakia|SimMIM| \cite{xie2022simmim}) can improve performance but at significant computational cost. This suggests complementary strengths between these paradigms that future research might productively integrate.

\subsubsection{Practical Constraints and Accessibility}

A significant practical challenge in advancing SSL research is the computational burden imposed by many state-of-the-art methods. Methods like \malakia|SimCLR|, \malakia|MoCo-v3|, \malakia|Barlow Twins|, and \malakia|VICReg| typically require large batch sizes (often 4096 or more) for optimal performance. Methods employing multi-crop augmentation, such as \malakia|SwAV| and \malakia|SMoG|, further increase computational demands. These requirements make SSL research inaccessible to many academic labs and smaller organizations lacking high-end computing infrastructure. Addressing this accessibility gap represents an important research direction. Methods like \malakia|MoBY| that demonstrate competitive performance with smaller batch sizes offer promising examples. Developing more computationally efficient SSL approaches would democratize research in this field and potentially lead to more diverse and innovative contributions from a broader community of researchers.

\subsection{Theoretical Foundations}
\label{sec:theory}

Most discriminative SSL methods are underpinned by various theoretical interpretations, although formal guarantees are still evolving. Contrastive learning and feature decorrelation methods stand out for their robust theoretical foundations, rooted in spectral graph theory, mutual information maximization, and information bottleneck principles, providing guarantees for linear separability and redundancy reduction through techniques like covariance matrix standardization; clustering-based approaches, while less formally grounded, leverage manifold learning to create transferable clusters aligned with data structures, using pseudo-labels for training; self-distillation methods, such as \malakia|BYOL| and \malakia|DINO|, focus on preventing representation collapse through momentum encoders and specific loss components, with emerging theories suggesting implicit decorrelation and mutual information maximization between teacher-student models, despite lacking formal performance guarantees; knowledge distillation, though sharing student-teacher concepts, is primarily a supervised technique for model compression or transfer, not an SSL pre-training method; overall, SSL methods vary from rigorous, theory-backed approaches like contrastive and decorrelation, to heuristic-driven but empirically effective methods like clustering and distillation, reflecting a spectrum from spectral and information-theoretic principles to practical, semantic manifold or label consistency-focused strategies.

Regarding performance guarantees, theoretical work primarily provides assurances for linear probe performance under specific assumptions about data distributions and augmentations, suggesting the learned features can be linearly separable for downstream tasks. However, providing rigorous guarantees for the more common and complex scenario of fine-tuning deep networks on diverse downstream tasks remains a significant challenge due to the non-linearities involved. Much of the theoretical analysis for methods like self-distillation focuses on proving the avoidance of representation collapse rather than direct downstream performance. Thus, while theoretical foundations offer valuable insights into why these methods work, their widespread adoption is largely driven by strong empirical results, with ongoing research seeking to bridge the gap between theory and practice, especially concerning robust performance guarantees.

\section{Conclusion}
\label{sec:conclusion}

This review has provided a comprehensive analysis of discriminative SSL methods in computer vision, categorizing them into contrastive methods, clustering methods, self-distillation methods, knowledge distillation methods, and feature decorrelation methods. Through extensive evaluation and comparison, we have traced the evolution of these approaches and identified key trends, challenges, and opportunities in the field.

The progress in SSL has been substantial, with recent methods even surpassing supervised learning baselines on standard benchmarks. This success demonstrates the potential of learning from unlabeled data, a capability with significant implications for domains where labeled data is scarce or expensive to obtain. However, several important challenges remain, including preventing representation collapse, balancing task-specific optimization, improving computational efficiency, and ensuring robustness to distribution shifts.

Future research in SSL should focus on addressing these challenges, particularly through: (1) developing more efficient methods that perform well with smaller batch sizes; (2) exploring novel regularization approaches that promote desired representational properties; (3) designing SSL techniques specifically tailored to transformer architectures; (4) establishing more comprehensive evaluation protocols that assess robustness and OOD generalization; and (5) creating benchmarks that better reflect the diversity and imperfections of real-world data.

As the field continues to develop, SSL has the potential to transform how we approach computer vision tasks, reducing reliance on labeled data while improving generalization to diverse real-world scenarios. By learning from vast amounts of unlabeled data, these methods may ultimately capture more nuanced and comprehensive visual representations than is possible through supervised learning alone.

\begin{sidewaystable}[!t]
\centering
\caption{Comparison of architectural and training characteristics across the \xazomara{MoCo} family of self-supervised learning methods.}
\label{tab:moco_comparison}
\begin{tabular}{lccc}
\hline
\multicolumn{1}{l}{} & \malakia|MoCo| & \malakia|MoCo-v2| & \malakia|MoCo-v3| \\
\hline
Backbone & ResNet & ResNet & ViT and ResNet \\
Negative sampling & Queue/Memory & Queue/Memory & Batch \\
Loss symmetry & Unidirectional& Unidirectional  & Bidirectional symmetrized \\
Projection head & Linear layer & MLP (no prediction head) & MLP + prediction head (query only) \\
Optimization & SGD & SGD & AdamW/LARS \\
Stability mechanisms & Momentum encoder & Momentum encoder & Frozen patch projection, long warmup \\
Data augmentation & Basic & Enhanced (\xazomara{SimCLR}) & Enhanced (\xazomara{SimCLR}) \\
Batch size & Small (256) & Small (256) & Large (4096) \\
Training duration & 200 epochs & 200/800 epochs & Up to 1000 epochs \\
\hline
\end{tabular}
\end{sidewaystable}

\begin{sidewaystable}[!t]
\centering
\caption{Comparison of various contrastive learning methods based on their projection heads, predictors, momentum encoders, memory usage, type of negatives, views, loss functions, and network architectures.}
\label{tab:contrastive}
\begin{tabular}{lcccccccc}
\hline
Method & Proj. & Pred. & Mom. & Mem. & Neg. & Views & Loss & Networks \\ \hline
\malakia|InstDis| \cite{wu2018instdis} & \checkmark &  &  & \checkmark & mem. & N/A & non-param. softmax & single \\
\malakia|PIRL| \cite{misra2019pirl} & \checkmark &  &  & \checkmark & mem. & Jigsaw & InfoNCE & siamese \\
\malakia|UEL| \cite{ye2019uel} & \checkmark &  &  &  & N/A & N/A & neg. log likelihood & siamese \\
\malakia|CPC| \cite{oord2019cpc, henaff2020cpcv2} &  & AR &  & \checkmark & mem. & image & InfoNCE & single \\
\malakia|CMC| \cite{tian2020cmc} &  &  &  & \checkmark & mem. & channels & InfoNCE & siamese \\ 
\malakia|SimCLR| \cite{chen2020simclr} & \checkmark &  &  &  & batch & aug. & NT-Xent & siamese \\
\malakia|SimCLR-v2| \cite{chen2020simclrv2} & \checkmark &  &  & \checkmark & mem. & aug. & NT-Xent & siamese \\
\malakia|G-SimCLR| \cite{chakraborty2020gsimclr} & \checkmark &  &  &  & batch & aug. & NT-Xent & siamese \\
\malakia|SupCon| \cite{khosla2021supcon} & \checkmark &  &  &  & N/A & aug. & SupCon & siamese \\
\malakia|SimCL| \cite{yang2022simcl} & \checkmark &  &  &  & batch & noise & InfoNCE & single \\ 
\malakia|MoCo| \cite{he2020moco} & \checkmark &  & \checkmark & \checkmark & mem. & aug. & InfoNCE & siamese \\
\malakia|MoCo-v2| \cite{chen2020mocov2} & \checkmark &  & \checkmark & \checkmark & mem. & aug. & InfoNCE & siamese \\
\malakia|MoCo-v3| \cite{chen2021mocov3} & \checkmark &  & \checkmark &  & batch & aug. & InfoNCE & siamese \\
\malakia|MoCHI| \cite{kalantidis2020mochi} & \checkmark &  & \checkmark & \checkmark & mem. & aug. & InfoNCE & siamese \\
\malakia|SimMoCo| \cite{zhang2022simco} & \checkmark &  & \checkmark &  & batch & aug. & DT & siamese \\
\malakia|SimCo| \cite{zhang2022simco} & \checkmark &  &  &  & batch & aug. & DT & siamese \\
\malakia|MOBY| \cite{xie2021moby} & \checkmark & \checkmark & \checkmark & \checkmark & mem. & aug. & InfoNCE & siamese \\
\malakia|LooC| \cite{xiao2021looc} & \checkmark &  & \checkmark & \checkmark & mem. & aug. & InfoNCE var. & siamese \\
\malakia|MoCLR| \cite{tian2021dnc} & \checkmark &  & \checkmark &  & batch & aug. & InfoNCE & siamese \\
\malakia|TiCo| \cite{zhu2022tico} & \checkmark &  & \checkmark & N/A & N/A & aug. & TiCo & siamese \\
\malakia|InfoMin| \cite{tian2020infomin} & \checkmark &  & \checkmark & \checkmark & mem. & aug. & InfoNCE & siamese \\
\malakia|ReSSL| \cite{zheng2021ressl} & \checkmark &  & \checkmark & \checkmark & mem. & aug. & softmax+CE & siamese \\ 
\malakia|NNCLR| \cite{dwibedi2021nnclr} & \checkmark & \checkmark &  & \checkmark & mem.+batch & aug. & Neighbor InfoNCE & siamese \\
\malakia|MSF| \cite{koohpayegani2021msf} & \checkmark & \checkmark & \checkmark & \checkmark & mem.+N/A & aug. & MSE & siamese \\
\malakia|All4One| \cite{estepa2023all4one} & \checkmark & \checkmark & \checkmark & \checkmark & mem.+batch & aug. & NNCLR+centroid+BT & siamese \\
\malakia|MYOW| \cite{azabou2021myow} & \checkmark & \checkmark & \checkmark & \checkmark & mem. & aug. & NCS & siamese \\
\malakia|Mugs| \cite{zhou2022mugs} & \checkmark & \checkmark & \checkmark & \checkmark & mem.+cent. & aug. & InfoNCE+CE & siamese \\ 
\malakia|MMCL| \cite{shah2021mmcl} & \checkmark &  &  &  & batch & aug. & NT-Xent & siamese \\
\malakia|MMCR| \cite{yerxa2023mmcr} & \checkmark &  & \checkmark &  & batch & aug. & nuclear norm & siamese \\ 
\malakia|ReLIC| \cite{mitrovic2020relic} & \checkmark &  &  &  & batch & aug. & InfoNCE+KL div. & siamese \\
\malakia|DCL| \cite{yeh2022dcl} & \checkmark &  &  &  & mem./batch & aug. & DCL/DCLW & siamese \\
\malakia|UniGrad| \cite{tao2022unigrad} & \checkmark &  & \checkmark &  & N/A & aug. & UniGrad & siamese \\
\malakia|CLSA| \cite{wang2022clsa} & \checkmark &  & \checkmark & \checkmark & mem. & aug. & contrastive+DDL & siamese \\ 
\malakia|AdCo| \cite{hu2021adco} & \checkmark &  & \checkmark & \checkmark & mem. & aug. & InfoNCE & siamese \\
\malakia|CaCo| \cite{wang2022caco} & \checkmark &  & \checkmark & \checkmark & mem. & aug. & InfoNCE & siamese \\ \hline
\end{tabular}
\end{sidewaystable}

\begin{sidewaystable}[!t]
\centering
\caption{Comparison of various dense contrastive methods based on their projection heads, predictors, momentum encoders, memory usage, negatives, views, loss functions, and network architectures. \raggedright RA: RoI Align; c-p: copy-paste; PPM: Pixel Propagation Module.}
\label{tab:dense}
\begin{tabular}{lcccccccc}
\hline
Method & Proj. & Pred. & Mom. & Mem. & Negatives & Views & Loss & Networks \\ \hline
\xazomara{DenseCL} \cite{wang2021densecl} & \checkmark &  & \checkmark & \checkmark & memory & aug. & InfoNCE+dense & siamese \\
\xazomara{DetConS} \cite{henaff2021detcon} & \checkmark &  &  &  & batch & aug. & InfoNCE & siamese \\
\xazomara{DetConB} \cite{henaff2021detcon} & \checkmark & \checkmark & \checkmark &  & batch & aug. & InfoNCE & siamese \\
\xazomara{InsLoc} \cite{yang2021insloc} & \checkmark & RA & \checkmark & \checkmark & memory & c-p & InfoNCE & siamese \\
\xazomara{CP2} \cite{wang2022cp2} & \checkmark &  & \checkmark & \checkmark & memory & c-p & InfoNCE & siamese \\
\xazomara{PixContrast} \cite{xie2021pixpro} & \checkmark &  & \checkmark &  & image & aug. & InfoNCE var. & siamese \\
\xazomara{PixPro} \cite{xie2021pixpro} & \checkmark & PPM & \checkmark &  & views & aug. & InfoNCE var.+cos. & siamese \\
\xazomara{UniVIP} \cite{li2022univip} & \checkmark & \checkmark & \checkmark &  & N/A & aug. & NCS+affinity+scene & siamese \\ \hline
\end{tabular}
\end{sidewaystable}

\begin{sidewaystable}[!t]
\centering
\caption{Comparison of various clustering methods based on their projection heads, predictors, momentum encoders, memory usage, online updates, prototypes, and loss functions.}
\label{tab:cluster}
\begin{tabular}{lcccccccc}
\hline
Method & Proj. & Pred. & Mom. & Mem. & Online & Proto & SK & Loss \\ 
\hline
\xazomara{Deep Cluster} \cite{caron2019deepcluster} &  &  &  &  &  &  &  & logistic \\
\xazomara{Deeper Cluster} \cite{caron2019deepercluster} & \checkmark &  &  &  &  &  &  & logistic \\
\malakia|LA| \cite{zhuang2019la} &  &  &  & \checkmark &  &  &  & log-lik. \\
\malakia|SeLa| \cite{asano2020sela} & \checkmark & softmax &  &  &  &  & \checkmark & CE \\
\malakia|PCL| \cite{li2021pcl} &  &  &  &  &  & \checkmark &  & InfoNCE+ProtoNCE \\
\malakia|SwAV| \cite{caron2020swav} & \checkmark & softmax &  &  & \checkmark & \checkmark & \checkmark & CE \\
\malakia|ODC| \cite{zhan2020odc} & \checkmark &  &  & \checkmark & \checkmark &  &  &  \\
\malakia|SMoG| \cite{pang2022smog} & \checkmark & \checkmark & \checkmark &  & \checkmark & \checkmark &  & InfoNCE+var. \\
\malakia|Self-Classifier| \cite{amrani2022selfclassifier} & \checkmark &  &  &  & \checkmark &  &  & CE var. \\ \hline
\end{tabular}
\end{sidewaystable}

\begin{sidewaystable}[!t]
\centering
\caption{Comparison of various self-distillation methods based on their projection heads, predictors, momentum encoders, memory usage, stop gradient operations, loss functions, symmetry, and network architectures.}
\label{tab:selfdistil}
\begin{tabular}{lcccccccc}
\hline
Method & Proj. & Pred. & Mom. & Mem. & \texttt{stop-grad} & Loss & Sym. & Networks \\ \hline
\malakia|BYOL| \cite{grill2020byol} & \checkmark & \checkmark & \checkmark &  & \checkmark & MSE & \checkmark & online/target \\
\malakia|DINO| \cite{caron2021dino} & \checkmark &  & \checkmark &  & \checkmark & softmax+CE & \checkmark & student/teacher \\
\malakia|SimSiam| \cite{chen2020simsiam} & \checkmark & \checkmark &  &  & \checkmark & NCS & \checkmark & siamese \\
\malakia|oBoW| \cite{gidaris2021obow} & \checkmark &  & \checkmark &  & \checkmark & CE &  & student/teacher \\
\malakia|DINO-v2| \cite{oquab2023dinov2} & \checkmark &  & \checkmark &  & \checkmark & softmax+CE & \checkmark & student/teacher \\
\malakia|FastSiam| \cite{pototzky2022fastsiam} & \checkmark & \checkmark &  &  & \checkmark & NCS &  & siamese \\ \hline
\end{tabular}
\end{sidewaystable}

\begin{sidewaystable}[!t]
\centering
\caption{Comparison of various knowledge distillation methods based on their projection heads, predictors, momentum encoders, memory usage, stop gradient operations, loss functions, symmetry, and network architectures.}
\label{tab:knowledgedistill}
\begin{tabular}{lcccccccc}
\hline
Method & Proj. & Pred. & Mom. & Mem. & \xazomara{stop-grad} & Loss & Sym. & Networks \\ \hline
\malakia|SEED| \cite{fang2021seed} & \checkmark &  &  & \checkmark & & softmax+CE &  & student/teacher \\
\malakia|DisCo| \cite{gao2022disco} & \checkmark &  & \checkmark &  & \checkmark & InfoNCE+MSE &  & student/mean/teacher \\
\malakia|BINGO| \cite{xu2022bingo} & \checkmark &  &  & \checkmark & & inter+intra InfoNCE &  & student/teacher \\ \hline
\end{tabular}
\end{sidewaystable}

\begin{sidewaystable}[!t]
\centering
\caption{Comparison of various feature decorrelation methods based on their projection heads, predictors, momentum encoders, memory usage, loss functions, and network architectures.}
\label{tab:cca}
\begin{tabular}{lccccccc}
\hline
Method & Proj. & Pred. & Mom. & Mem. & Loss & Networks \\ \hline
\xazomara{Barlow Twins} \cite{zbontar2021barlowtwins} & \checkmark &  &  &  & Cross Correlation & siamese \\
\xazomara{VICReg} \cite{bardes2022vicreg} & \checkmark &  &  &  & Variance-Invariance-Covariance & siamese \\
\xazomara{W-MSE} \cite{ermolov2021wmse} & \checkmark &  &  &  & MSE & siamese \\
\xazomara{Mixed Barlow Twins} \cite{bandara2023mixedbarlowtwins} & \checkmark &  &  &  & Cross Correlation & siamese \\
\xazomara{TWIST} \cite{wang2021twist} & \checkmark & softmax &  &  & Mutual Information & siamese \\
\xazomara{TLDR} \cite{kalantidis2022tldr} & \checkmark &  &  &  & Cross Correlation & siamese \\
\xazomara{ARB} \cite{zhang2022arb} & \checkmark &  &  &  & MSE & siamese \\
\xazomara{Truncated Triplet} \cite{wang2021truncatedtriplet} & \checkmark &  & \checkmark &  & Truncated Triplet & siamese \\ \hline
\end{tabular}
\end{sidewaystable}

\bibliography{sn-bibliography}


\begin{thebibliography}{200}
\ifx \bisbn   \undefined \def \bisbn  #1{ISBN #1}\fi
\ifx \binits  \undefined \def \binits#1{#1}\fi
\ifx \bauthor  \undefined \def \bauthor#1{#1}\fi
\ifx \batitle  \undefined \def \batitle#1{#1}\fi
\ifx \bjtitle  \undefined \def \bjtitle#1{#1}\fi
\ifx \bvolume  \undefined \def \bvolume#1{\textbf{#1}}\fi
\ifx \byear  \undefined \def \byear#1{#1}\fi
\ifx \bissue  \undefined \def \bissue#1{#1}\fi
\ifx \bfpage  \undefined \def \bfpage#1{#1}\fi
\ifx \blpage  \undefined \def \blpage #1{#1}\fi
\ifx \burl  \undefined \def \burl#1{\textsf{#1}}\fi
\ifx \doiurl  \undefined \def \doiurl#1{\url{https://doi.org/#1}}\fi
\ifx \betal  \undefined \def \betal{\textit{et al.}}\fi
\ifx \binstitute  \undefined \def \binstitute#1{#1}\fi
\ifx \binstitutionaled  \undefined \def \binstitutionaled#1{#1}\fi
\ifx \bctitle  \undefined \def \bctitle#1{#1}\fi
\ifx \beditor  \undefined \def \beditor#1{#1}\fi
\ifx \bpublisher  \undefined \def \bpublisher#1{#1}\fi
\ifx \bbtitle  \undefined \def \bbtitle#1{#1}\fi
\ifx \bedition  \undefined \def \bedition#1{#1}\fi
\ifx \bseriesno  \undefined \def \bseriesno#1{#1}\fi
\ifx \blocation  \undefined \def \blocation#1{#1}\fi
\ifx \bsertitle  \undefined \def \bsertitle#1{#1}\fi
\ifx \bsnm \undefined \def \bsnm#1{#1}\fi
\ifx \bsuffix \undefined \def \bsuffix#1{#1}\fi
\ifx \bparticle \undefined \def \bparticle#1{#1}\fi
\ifx \barticle \undefined \def \barticle#1{#1}\fi
\bibcommenthead
\ifx \bconfdate \undefined \def \bconfdate #1{#1}\fi
\ifx \botherref \undefined \def \botherref #1{#1}\fi
\ifx \url \undefined \def \url#1{\textsf{#1}}\fi
\ifx \bchapter \undefined \def \bchapter#1{#1}\fi
\ifx \bbook \undefined \def \bbook#1{#1}\fi
\ifx \bcomment \undefined \def \bcomment#1{#1}\fi
\ifx \oauthor \undefined \def \oauthor#1{#1}\fi
\ifx \citeauthoryear \undefined \def \citeauthoryear#1{#1}\fi
\ifx \endbibitem  \undefined \def \endbibitem {}\fi
\ifx \bconflocation  \undefined \def \bconflocation#1{#1}\fi
\ifx \arxivurl  \undefined \def \arxivurl#1{\textsf{#1}}\fi
\csname PreBibitemsHook\endcsname

\bibitem[\protect\citeauthoryear{Krizhevsky et~al.}{2012}]{krizhevsky2012alexnet}
\begin{bchapter}
\bauthor{\bsnm{Krizhevsky}, \binits{A.}},
\bauthor{\bsnm{Sutskever}, \binits{I.}},
\bauthor{\bsnm{Hinton}, \binits{G.E.}}:
\bctitle{Imagenet classification with deep convolutional neural networks}.
In: \beditor{\bsnm{Pereira}, \binits{F.}},
\beditor{\bsnm{Burges}, \binits{C.J.}},
\beditor{\bsnm{Bottou}, \binits{L.}},
\beditor{\bsnm{Weinberger}, \binits{K.Q.}} (eds.)
\bbtitle{Advances in Neural Information Processing Systems},
vol. \bseriesno{25}.
\bpublisher{Curran Associates, Inc.}, \blocation{???}
(\byear{2012}).
\burl{https://proceedings.neurips.cc/paper_files/paper/2012/file/c399862d3b9d6b76c8436e924a68c45b-Paper.pdf}
\end{bchapter}
\endbibitem

\bibitem[\protect\citeauthoryear{Dosovitskiy et~al.}{2021}]{dosovitskiy2021vit}
\begin{botherref}
\oauthor{\bsnm{Dosovitskiy}, \binits{A.}},
\oauthor{\bsnm{Beyer}, \binits{L.}},
\oauthor{\bsnm{Kolesnikov}, \binits{A.}},
\oauthor{\bsnm{Weissenborn}, \binits{D.}},
\oauthor{\bsnm{Zhai}, \binits{X.}},
\oauthor{\bsnm{Unterthiner}, \binits{T.}},
\oauthor{\bsnm{Dehghani}, \binits{M.}},
\oauthor{\bsnm{Minderer}, \binits{M.}},
\oauthor{\bsnm{Heigold}, \binits{G.}},
\oauthor{\bsnm{Gelly}, \binits{S.}},
\oauthor{\bsnm{Uszkoreit}, \binits{J.}},
\oauthor{\bsnm{Houlsby}, \binits{N.}}:
An Image is Worth 16x16 Words: Transformers for Image Recognition at Scale
(2021)
\end{botherref}
\endbibitem

\bibitem[\protect\citeauthoryear{Touvron et~al.}{2021}]{touvron2021deit}
\begin{botherref}
\oauthor{\bsnm{Touvron}, \binits{H.}},
\oauthor{\bsnm{Cord}, \binits{M.}},
\oauthor{\bsnm{Douze}, \binits{M.}},
\oauthor{\bsnm{Massa}, \binits{F.}},
\oauthor{\bsnm{Sablayrolles}, \binits{A.}},
\oauthor{\bsnm{Jégou}, \binits{H.}}:
Training data-efficient image transformers \& distillation through attention
(2021).
\url{https://arxiv.org/abs/2012.12877}
\end{botherref}
\endbibitem

\bibitem[\protect\citeauthoryear{Vaswani et~al.}{2023}]{vaswani2023attention}
\begin{botherref}
\oauthor{\bsnm{Vaswani}, \binits{A.}},
\oauthor{\bsnm{Shazeer}, \binits{N.}},
\oauthor{\bsnm{Parmar}, \binits{N.}},
\oauthor{\bsnm{Uszkoreit}, \binits{J.}},
\oauthor{\bsnm{Jones}, \binits{L.}},
\oauthor{\bsnm{Gomez}, \binits{A.N.}},
\oauthor{\bsnm{Kaiser}, \binits{L.}},
\oauthor{\bsnm{Polosukhin}, \binits{I.}}:
Attention Is All You Need
(2023)
\end{botherref}
\endbibitem

\bibitem[\protect\citeauthoryear{Alzubaidi et~al.}{2023}]{alzubaidi2023survey}
\begin{botherref}
\oauthor{\bsnm{Alzubaidi}, \binits{L.}},
\oauthor{\bsnm{Bai}, \binits{J.}},
\oauthor{\bsnm{Al-Sabaawi}, \binits{A.}},
\oauthor{\bsnm{Santamaría}, \binits{J.}},
\oauthor{\bsnm{Albahri}, \binits{A.S.}},
\oauthor{\bsnm{Al-dabbagh}, \binits{B.S.N.}},
\oauthor{\bsnm{Fadhel}, \binits{M.A.}},
\oauthor{\bsnm{Manoufali}, \binits{M.}},
\oauthor{\bsnm{Zhang}, \binits{J.}},
\oauthor{\bsnm{Al-Timemy}, \binits{A.H.}},
\oauthor{\bsnm{Duan}, \binits{Y.}},
\oauthor{\bsnm{Abdullah}, \binits{A.}},
\oauthor{\bsnm{Farhan}, \binits{L.}},
\oauthor{\bsnm{Lu}, \binits{Y.}},
\oauthor{\bsnm{Gupta}, \binits{A.}},
\oauthor{\bsnm{Albu}, \binits{F.}},
\oauthor{\bsnm{Abbosh}, \binits{A.}},
\oauthor{\bsnm{Gu}, \binits{Y.}}:
A survey on deep learning tools dealing with data scarcity: definitions, challenges, solutions, tips, and applications.
Journal of Big Data
\textbf{10}(1)
(2023)
\doiurl{10.1186/s40537-023-00727-2}
\end{botherref}
\endbibitem

\bibitem[\protect\citeauthoryear{Bengio et~al.}{2014}]{bengio2014representationlearningreviewnew}
\begin{botherref}
\oauthor{\bsnm{Bengio}, \binits{Y.}},
\oauthor{\bsnm{Courville}, \binits{A.}},
\oauthor{\bsnm{Vincent}, \binits{P.}}:
Representation Learning: A Review and New Perspectives
(2014).
\url{https://arxiv.org/abs/1206.5538}
\end{botherref}
\endbibitem

\bibitem[\protect\citeauthoryear{Balestriero et~al.}{2023}]{balestriero2023cookbook}
\begin{botherref}
\oauthor{\bsnm{Balestriero}, \binits{R.}},
\oauthor{\bsnm{Ibrahim}, \binits{M.}},
\oauthor{\bsnm{Sobal}, \binits{V.}},
\oauthor{\bsnm{Morcos}, \binits{A.}},
\oauthor{\bsnm{Shekhar}, \binits{S.}},
\oauthor{\bsnm{Goldstein}, \binits{T.}},
\oauthor{\bsnm{Bordes}, \binits{F.}},
\oauthor{\bsnm{Bardes}, \binits{A.}},
\oauthor{\bsnm{Mialon}, \binits{G.}},
\oauthor{\bsnm{Tian}, \binits{Y.}},
\oauthor{\bsnm{Schwarzschild}, \binits{A.}},
\oauthor{\bsnm{Wilson}, \binits{A.G.}},
\oauthor{\bsnm{Geiping}, \binits{J.}},
\oauthor{\bsnm{Garrido}, \binits{Q.}},
\oauthor{\bsnm{Fernandez}, \binits{P.}},
\oauthor{\bsnm{Bar}, \binits{A.}},
\oauthor{\bsnm{Pirsiavash}, \binits{H.}},
\oauthor{\bsnm{LeCun}, \binits{Y.}},
\oauthor{\bsnm{Goldblum}, \binits{M.}}:
A Cookbook of Self-Supervised Learning
(2023)
\end{botherref}
\endbibitem

\bibitem[\protect\citeauthoryear{Ohri and Kumar}{2021}]{ohri2021review}
\begin{barticle}
\bauthor{\bsnm{Ohri}, \binits{K.}},
\bauthor{\bsnm{Kumar}, \binits{M.}}:
\batitle{Review on self-supervised image recognition using deep neural networks}.
\bjtitle{Knowledge-Based Systems}
\bvolume{224},
\bfpage{107090}
(\byear{2021})
\doiurl{10.1016/j.knosys.2021.107090}
\end{barticle}
\endbibitem

\bibitem[\protect\citeauthoryear{Sermanet et~al.}{2018}]{sermanet2018time}
\begin{bchapter}
\bauthor{\bsnm{Sermanet}, \binits{P.}},
\bauthor{\bsnm{Lynch}, \binits{C.}},
\bauthor{\bsnm{Chebotar}, \binits{Y.}},
\bauthor{\bsnm{Hsu}, \binits{J.}},
\bauthor{\bsnm{Jang}, \binits{E.}},
\bauthor{\bsnm{Schaal}, \binits{S.}},
\bauthor{\bsnm{Levine}, \binits{S.}},
\bauthor{\bsnm{Brain}, \binits{G.}}:
\bctitle{Time-contrastive networks: Self-supervised learning from video}.
In: \bbtitle{2018 IEEE International Conference on Robotics and Automation (ICRA)},
pp. \bfpage{1134}--\blpage{1141}
(\byear{2018}).
\doiurl{10.1109/ICRA.2018.8462891}
\end{bchapter}
\endbibitem

\bibitem[\protect\citeauthoryear{Noroozi and Favaro}{2016}]{noroozi2017unsupervised}
\begin{botherref}
\oauthor{\bsnm{Noroozi}, \binits{M.}},
\oauthor{\bsnm{Favaro}, \binits{P.}}:
Unsupervised Learning of Visual Representations by Solving Jigsaw Puzzles.
arXiv
(2016).
\doiurl{10.48550/ARXIV.1603.09246} .
\url{https://arxiv.org/abs/1603.09246}
\end{botherref}
\endbibitem

\bibitem[\protect\citeauthoryear{Lan et~al.}{2020}]{lan2020albert}
\begin{botherref}
\oauthor{\bsnm{Lan}, \binits{Z.}},
\oauthor{\bsnm{Chen}, \binits{M.}},
\oauthor{\bsnm{Goodman}, \binits{S.}},
\oauthor{\bsnm{Gimpel}, \binits{K.}},
\oauthor{\bsnm{Sharma}, \binits{P.}},
\oauthor{\bsnm{Soricut}, \binits{R.}}:
ALBERT: A Lite BERT for Self-supervised Learning of Language Representations
(2020)
\end{botherref}
\endbibitem

\bibitem[\protect\citeauthoryear{Devlin et~al.}{2018}]{devlin2018bert}
\begin{botherref}
\oauthor{\bsnm{Devlin}, \binits{J.}},
\oauthor{\bsnm{Chang}, \binits{M.-W.}},
\oauthor{\bsnm{Lee}, \binits{K.}},
\oauthor{\bsnm{Toutanova}, \binits{K.}}:
BERT: Pre-training of Deep Bidirectional Transformers for Language Understanding
(2018)
\end{botherref}
\endbibitem

\bibitem[\protect\citeauthoryear{Brown et~al.}{2020}]{brown2020language}
\begin{botherref}
\oauthor{\bsnm{Brown}, \binits{T.B.}},
\oauthor{\bsnm{Mann}, \binits{B.}},
\oauthor{\bsnm{Ryder}, \binits{N.}},
\oauthor{\bsnm{Subbiah}, \binits{M.}},
\oauthor{\bsnm{Kaplan}, \binits{J.}},
\oauthor{\bsnm{Dhariwal}, \binits{P.}},
\oauthor{\bsnm{Neelakantan}, \binits{A.}},
\oauthor{\bsnm{Shyam}, \binits{P.}},
\oauthor{\bsnm{Sastry}, \binits{G.}},
\oauthor{\bsnm{Askell}, \binits{A.}},
\oauthor{\bsnm{Agarwal}, \binits{S.}},
\oauthor{\bsnm{Herbert-Voss}, \binits{A.}},
\oauthor{\bsnm{Krueger}, \binits{G.}},
\oauthor{\bsnm{Henighan}, \binits{T.}},
\oauthor{\bsnm{Child}, \binits{R.}},
\oauthor{\bsnm{Ramesh}, \binits{A.}},
\oauthor{\bsnm{Ziegler}, \binits{D.M.}},
\oauthor{\bsnm{Wu}, \binits{J.}},
\oauthor{\bsnm{Winter}, \binits{C.}},
\oauthor{\bsnm{Hesse}, \binits{C.}},
\oauthor{\bsnm{Chen}, \binits{M.}},
\oauthor{\bsnm{Sigler}, \binits{E.}},
\oauthor{\bsnm{Litwin}, \binits{M.}},
\oauthor{\bsnm{Gray}, \binits{S.}},
\oauthor{\bsnm{Chess}, \binits{B.}},
\oauthor{\bsnm{Clark}, \binits{J.}},
\oauthor{\bsnm{Berner}, \binits{C.}},
\oauthor{\bsnm{McCandlish}, \binits{S.}},
\oauthor{\bsnm{Radford}, \binits{A.}},
\oauthor{\bsnm{Sutskever}, \binits{I.}},
\oauthor{\bsnm{Amodei}, \binits{D.}}:
Language Models are Few-Shot Learners
(2020)
\end{botherref}
\endbibitem

\bibitem[\protect\citeauthoryear{Chen et~al.}{2017}]{chen2017reading}
\begin{botherref}
\oauthor{\bsnm{Chen}, \binits{D.}},
\oauthor{\bsnm{Fisch}, \binits{A.}},
\oauthor{\bsnm{Weston}, \binits{J.}},
\oauthor{\bsnm{Bordes}, \binits{A.}}:
Reading Wikipedia to Answer Open-Domain Questions
(2017)
\end{botherref}
\endbibitem

\bibitem[\protect\citeauthoryear{Hamilton et~al.}{2017}]{hamilton2017advances}
\begin{bchapter}
\bauthor{\bsnm{Hamilton}, \binits{W.}},
\bauthor{\bsnm{Ying}, \binits{Z.}},
\bauthor{\bsnm{Leskovec}, \binits{J.}}:
\bctitle{Inductive representation learning on large graphs}.
In: \beditor{\bsnm{Guyon}, \binits{I.}},
\beditor{\bsnm{Luxburg}, \binits{U.V.}},
\beditor{\bsnm{Bengio}, \binits{S.}},
\beditor{\bsnm{Wallach}, \binits{H.}},
\beditor{\bsnm{Fergus}, \binits{R.}},
\beditor{\bsnm{Vishwanathan}, \binits{S.}},
\beditor{\bsnm{Garnett}, \binits{R.}} (eds.)
\bbtitle{Advances in Neural Information Processing Systems},
vol. \bseriesno{30}.
\bpublisher{Curran Associates, Inc.}, \blocation{???}
(\byear{2017}).
\burl{https://proceedings.neurips.cc/paper_files/paper/2017/file/5dd9db5e033da9c6fb5ba83c7a7ebea9-Paper.pdf}
\end{bchapter}
\endbibitem

\bibitem[\protect\citeauthoryear{He et~al.}{2020}]{he2020moco}
\begin{botherref}
\oauthor{\bsnm{He}, \binits{K.}},
\oauthor{\bsnm{Fan}, \binits{H.}},
\oauthor{\bsnm{Wu}, \binits{Y.}},
\oauthor{\bsnm{Xie}, \binits{S.}},
\oauthor{\bsnm{Girshick}, \binits{R.}}:
Momentum Contrast for Unsupervised Visual Representation Learning
(2020)
\end{botherref}
\endbibitem

\bibitem[\protect\citeauthoryear{Goyal et~al.}{2021}]{goyal2021seer}
\begin{botherref}
\oauthor{\bsnm{Goyal}, \binits{P.}},
\oauthor{\bsnm{Caron}, \binits{M.}},
\oauthor{\bsnm{Lefaudeux}, \binits{B.}},
\oauthor{\bsnm{Xu}, \binits{M.}},
\oauthor{\bsnm{Wang}, \binits{P.}},
\oauthor{\bsnm{Pai}, \binits{V.}},
\oauthor{\bsnm{Singh}, \binits{M.}},
\oauthor{\bsnm{Liptchinsky}, \binits{V.}},
\oauthor{\bsnm{Misra}, \binits{I.}},
\oauthor{\bsnm{Joulin}, \binits{A.}},
\oauthor{\bsnm{Bojanowski}, \binits{P.}}:
Self-supervised Pretraining of Visual Features in the Wild
(2021)
\end{botherref}
\endbibitem

\bibitem[\protect\citeauthoryear{Goyal et~al.}{2022}]{goyal2022seer}
\begin{botherref}
\oauthor{\bsnm{Goyal}, \binits{P.}},
\oauthor{\bsnm{Duval}, \binits{Q.}},
\oauthor{\bsnm{Seessel}, \binits{I.}},
\oauthor{\bsnm{Caron}, \binits{M.}},
\oauthor{\bsnm{Misra}, \binits{I.}},
\oauthor{\bsnm{Sagun}, \binits{L.}},
\oauthor{\bsnm{Joulin}, \binits{A.}},
\oauthor{\bsnm{Bojanowski}, \binits{P.}}:
Vision Models Are More Robust And Fair When Pretrained On Uncurated Images Without Supervision
(2022)
\end{botherref}
\endbibitem

\bibitem[\protect\citeauthoryear{Deng et~al.}{2009}]{deng2009imagenet}
\begin{botherref}
\oauthor{\bsnm{Deng}, \binits{J.}},
\oauthor{\bsnm{Dong}, \binits{W.}},
\oauthor{\bsnm{Socher}, \binits{R.}},
\oauthor{\bsnm{Li}, \binits{L.-J.}},
\oauthor{\bsnm{Li}, \binits{K.}},
\oauthor{\bsnm{Fei-Fei}, \binits{L.}}:
Imagenet: A large-scale hierarchical image database.
2009 IEEE Conference on Computer Vision and Pattern Recognition,
248--255
(2009)
\end{botherref}
\endbibitem

\bibitem[\protect\citeauthoryear{Russakovsky et~al.}{2015}]{russakovsky2015imagenet}
\begin{botherref}
\oauthor{\bsnm{Russakovsky}, \binits{O.}},
\oauthor{\bsnm{Deng}, \binits{J.}},
\oauthor{\bsnm{Su}, \binits{H.}},
\oauthor{\bsnm{Krause}, \binits{J.}},
\oauthor{\bsnm{Satheesh}, \binits{S.}},
\oauthor{\bsnm{Ma}, \binits{S.}},
\oauthor{\bsnm{Huang}, \binits{Z.}},
\oauthor{\bsnm{Karpathy}, \binits{A.}},
\oauthor{\bsnm{Khosla}, \binits{A.}},
\oauthor{\bsnm{Bernstein}, \binits{M.}},
\oauthor{\bsnm{Berg}, \binits{A.C.}},
\oauthor{\bsnm{Fei-Fei}, \binits{L.}}:
ImageNet Large Scale Visual Recognition Challenge
(2015)
\end{botherref}
\endbibitem

\bibitem[\protect\citeauthoryear{van~den Oord et~al.}{2019}]{oord2019cpc}
\begin{botherref}
\oauthor{\bsnm{Oord}, \binits{A.}},
\oauthor{\bsnm{Li}, \binits{Y.}},
\oauthor{\bsnm{Vinyals}, \binits{O.}}:
Representation Learning with Contrastive Predictive Coding
(2019)
\end{botherref}
\endbibitem

\bibitem[\protect\citeauthoryear{Zhang et~al.}{2024}]{zhang2024selfsupervised}
\begin{botherref}
\oauthor{\bsnm{Zhang}, \binits{K.}},
\oauthor{\bsnm{Wen}, \binits{Q.}},
\oauthor{\bsnm{Zhang}, \binits{C.}},
\oauthor{\bsnm{Cai}, \binits{R.}},
\oauthor{\bsnm{Jin}, \binits{M.}},
\oauthor{\bsnm{Liu}, \binits{Y.}},
\oauthor{\bsnm{Zhang}, \binits{J.}},
\oauthor{\bsnm{Liang}, \binits{Y.}},
\oauthor{\bsnm{Pang}, \binits{G.}},
\oauthor{\bsnm{Song}, \binits{D.}},
\oauthor{\bsnm{Pan}, \binits{S.}}:
Self-Supervised Learning for Time Series Analysis: Taxonomy, Progress, and Prospects
(2024)
\end{botherref}
\endbibitem

\bibitem[\protect\citeauthoryear{Liu et~al.}{2022}]{liu2022audio}
\begin{barticle}
\bauthor{\bsnm{Liu}, \binits{S.}},
\bauthor{\bsnm{Mallol-Ragolta}, \binits{A.}},
\bauthor{\bsnm{Parada-Cabaleiro}, \binits{E.}},
\bauthor{\bsnm{Qian}, \binits{K.}},
\bauthor{\bsnm{Jing}, \binits{X.}},
\bauthor{\bsnm{Kathan}, \binits{A.}},
\bauthor{\bsnm{Hu}, \binits{B.}},
\bauthor{\bsnm{Schuller}, \binits{B.W.}}:
\batitle{Audio self-supervised learning: A survey}.
\bjtitle{Patterns}
\bvolume{3}(\bissue{12}),
\bfpage{100616}
(\byear{2022})
\doiurl{10.1016/j.patter.2022.100616}
\end{barticle}
\endbibitem

\bibitem[\protect\citeauthoryear{Schiappa et~al.}{2023}]{schiappa2023self}
\begin{barticle}
\bauthor{\bsnm{Schiappa}, \binits{M.C.}},
\bauthor{\bsnm{Rawat}, \binits{Y.S.}},
\bauthor{\bsnm{Shah}, \binits{M.}}:
\batitle{Self-supervised learning for videos: A survey}.
\bjtitle{ACM Computing Surveys}
\bvolume{55}(\bissue{13s}),
\bfpage{1}--\blpage{37}
(\byear{2023})
\doiurl{10.1145/3577925}
\end{barticle}
\endbibitem

\bibitem[\protect\citeauthoryear{Kingma and Welling}{2013}]{kingma2013autoencoding}
\begin{botherref}
\oauthor{\bsnm{Kingma}, \binits{D.P.}},
\oauthor{\bsnm{Welling}, \binits{M.}}:
Auto-Encoding Variational Bayes
(2013)
\end{botherref}
\endbibitem

\bibitem[\protect\citeauthoryear{Vincent et~al.}{2008}]{vincent2008extracting}
\begin{bchapter}
\bauthor{\bsnm{Vincent}, \binits{P.}},
\bauthor{\bsnm{Larochelle}, \binits{H.}},
\bauthor{\bsnm{Bengio}, \binits{Y.}},
\bauthor{\bsnm{Manzagol}, \binits{P.-A.}}:
\bctitle{Extracting and composing robust features with denoising autoencoders},
pp. \bfpage{1096}--\blpage{1103}
(\byear{2008}).
\doiurl{10.1145/1390156.1390294}
\end{bchapter}
\endbibitem

\bibitem[\protect\citeauthoryear{Goodfellow et~al.}{2014}]{goodfellow2014generative}
\begin{botherref}
\oauthor{\bsnm{Goodfellow}, \binits{I.J.}},
\oauthor{\bsnm{Pouget-Abadie}, \binits{J.}},
\oauthor{\bsnm{Mirza}, \binits{M.}},
\oauthor{\bsnm{Xu}, \binits{B.}},
\oauthor{\bsnm{Warde-Farley}, \binits{D.}},
\oauthor{\bsnm{Ozair}, \binits{S.}},
\oauthor{\bsnm{Courville}, \binits{A.}},
\oauthor{\bsnm{Bengio}, \binits{Y.}}:
Generative Adversarial Networks
(2014)
\end{botherref}
\endbibitem

\bibitem[\protect\citeauthoryear{Chen et~al.}{2020}]{chen2020simclr}
\begin{botherref}
\oauthor{\bsnm{Chen}, \binits{T.}},
\oauthor{\bsnm{Kornblith}, \binits{S.}},
\oauthor{\bsnm{Norouzi}, \binits{M.}},
\oauthor{\bsnm{Hinton}, \binits{G.}}:
A Simple Framework for Contrastive Learning of Visual Representations
(2020)
\end{botherref}
\endbibitem

\bibitem[\protect\citeauthoryear{Grill et~al.}{2020}]{grill2020byol}
\begin{botherref}
\oauthor{\bsnm{Grill}, \binits{J.-B.}},
\oauthor{\bsnm{Strub}, \binits{F.}},
\oauthor{\bsnm{Altché}, \binits{F.}},
\oauthor{\bsnm{Tallec}, \binits{C.}},
\oauthor{\bsnm{Richemond}, \binits{P.H.}},
\oauthor{\bsnm{Buchatskaya}, \binits{E.}},
\oauthor{\bsnm{Doersch}, \binits{C.}},
\oauthor{\bsnm{Pires}, \binits{B.A.}},
\oauthor{\bsnm{Guo}, \binits{Z.D.}},
\oauthor{\bsnm{Azar}, \binits{M.G.}},
\oauthor{\bsnm{Piot}, \binits{B.}},
\oauthor{\bsnm{Kavukcuoglu}, \binits{K.}},
\oauthor{\bsnm{Munos}, \binits{R.}},
\oauthor{\bsnm{Valko}, \binits{M.}}:
Bootstrap your own latent: A new approach to self-supervised Learning
(2020)
\end{botherref}
\endbibitem

\bibitem[\protect\citeauthoryear{Doersch et~al.}{2015}]{doersch2015unsupervised}
\begin{botherref}
\oauthor{\bsnm{Doersch}, \binits{C.}},
\oauthor{\bsnm{Gupta}, \binits{A.}},
\oauthor{\bsnm{Efros}, \binits{A.A.}}:
Unsupervised visual representation learning by context prediction.
CoRR
\textbf{abs/1505.05192}
(2015)
{\href{https://arxiv.org/abs/1505.05192}{{1505.05192}}}
\end{botherref}
\endbibitem

\bibitem[\protect\citeauthoryear{Krishnan et~al.}{2022}]{rishnan2022self}
\begin{barticle}
\bauthor{\bsnm{Krishnan}, \binits{R.}},
\bauthor{\bsnm{Rajpurkar}, \binits{P.}},
\bauthor{\bsnm{Topol}, \binits{E.J.}}:
\batitle{Self-supervised learning in medicine and healthcare}.
\bjtitle{Nature Biomedical Engineering}
\bvolume{6}(\bissue{12}),
\bfpage{1346}--\blpage{1352}
(\byear{2022})
\doiurl{10.1038/s41551-022-00914-1}
\end{barticle}
\endbibitem

\bibitem[\protect\citeauthoryear{Wang et~al.}{2022}]{wang2022self}
\begin{barticle}
\bauthor{\bsnm{Wang}, \binits{Y.}},
\bauthor{\bsnm{Albrecht}, \binits{C.M.}},
\bauthor{\bsnm{Braham}, \binits{N.A.A.}},
\bauthor{\bsnm{Mou}, \binits{L.}},
\bauthor{\bsnm{Zhu}, \binits{X.X.}}:
\batitle{Self-supervised learning in remote sensing: A review}.
\bjtitle{IEEE Geoscience and Remote Sensing Magazine}
\bvolume{10}(\bissue{4}),
\bfpage{213}--\blpage{247}
(\byear{2022})
\doiurl{10.1109/MGRS.2022.3198244}
\end{barticle}
\endbibitem

\bibitem[\protect\citeauthoryear{Shurrab and Duwairi}{2022}]{shurrab2022self}
\begin{barticle}
\bauthor{\bsnm{Shurrab}, \binits{S.}},
\bauthor{\bsnm{Duwairi}, \binits{R.}}:
\batitle{Self-supervised learning methods and applications in medical imaging analysis: a survey}.
\bjtitle{PeerJ Computer Science}
\bvolume{8},
\bfpage{1045}
(\byear{2022})
\doiurl{10.7717/peerj-cs.1045}
\end{barticle}
\endbibitem

\bibitem[\protect\citeauthoryear{Chowdhury et~al.}{2021}]{chowdhury2021applying}
\begin{barticle}
\bauthor{\bsnm{Chowdhury}, \binits{A.}},
\bauthor{\bsnm{Rosenthal}, \binits{J.}},
\bauthor{\bsnm{Waring}, \binits{J.}},
\bauthor{\bsnm{Umeton}, \binits{R.}}:
\batitle{Applying self-supervised learning to medicine: Review of the state of the art and medical implementations}.
\bjtitle{Informatics}
\bvolume{8}(\bissue{3}),
\bfpage{59}
(\byear{2021})
\doiurl{10.3390/informatics8030059}
\end{barticle}
\endbibitem

\bibitem[\protect\citeauthoryear{Ozbulak et~al.}{2023}]{ozbulak2023know}
\begin{botherref}
\oauthor{\bsnm{Ozbulak}, \binits{U.}},
\oauthor{\bsnm{Lee}, \binits{H.J.}},
\oauthor{\bsnm{Boga}, \binits{B.}},
\oauthor{\bsnm{Anzaku}, \binits{E.T.}},
\oauthor{\bsnm{Park}, \binits{H.}},
\oauthor{\bsnm{Messem}, \binits{A.V.}},
\oauthor{\bsnm{Neve}, \binits{W.D.}},
\oauthor{\bsnm{Vankerschaver}, \binits{J.}}:
Know Your Self-supervised Learning: A Survey on Image-based Generative and Discriminative Training
(2023)
\end{botherref}
\endbibitem

\bibitem[\protect\citeauthoryear{Rani et~al.}{2023}]{rani2023self}
\begin{barticle}
\bauthor{\bsnm{Rani}, \binits{V.}},
\bauthor{\bsnm{Nabi}, \binits{S.T.}},
\bauthor{\bsnm{Kumar}, \binits{M.}},
\bauthor{\bsnm{Mittal}, \binits{A.}},
\bauthor{\bsnm{Kumar}, \binits{K.}}:
\batitle{Self-supervised learning: A succinct review}.
\bjtitle{Archives of Computational Methods in Engineering}
\bvolume{30}(\bissue{4}),
\bfpage{2761}--\blpage{2775}
(\byear{2023})
\doiurl{10.1007/s11831-023-09884-2}
\end{barticle}
\endbibitem

\bibitem[\protect\citeauthoryear{Jaiswal et~al.}{2020}]{jaiswal2020survey}
\begin{botherref}
\oauthor{\bsnm{Jaiswal}, \binits{A.}},
\oauthor{\bsnm{Babu}, \binits{A.R.}},
\oauthor{\bsnm{Zadeh}, \binits{M.Z.}},
\oauthor{\bsnm{Banerjee}, \binits{D.}},
\oauthor{\bsnm{Makedon}, \binits{F.}}:
A Survey on Contrastive Self-supervised Learning
(2020)
\end{botherref}
\endbibitem

\bibitem[\protect\citeauthoryear{Jing and Tian}{2019}]{jing2019selfsupervised}
\begin{botherref}
\oauthor{\bsnm{Jing}, \binits{L.}},
\oauthor{\bsnm{Tian}, \binits{Y.}}:
Self-supervised Visual Feature Learning with Deep Neural Networks: A Survey
(2019)
\end{botherref}
\endbibitem

\bibitem[\protect\citeauthoryear{Kumar et~al.}{2022}]{kumar2022contrastive}
\begin{barticle}
\bauthor{\bsnm{Kumar}, \binits{P.}},
\bauthor{\bsnm{Rawat}, \binits{P.}},
\bauthor{\bsnm{Chauhan}, \binits{S.}}:
\batitle{Contrastive self-supervised learning: review, progress, challenges and future research directions}.
\bjtitle{International Journal of Multimedia Information Retrieval}
\bvolume{11}(\bissue{4}),
\bfpage{461}--\blpage{488}
(\byear{2022})
\doiurl{10.1007/s13735-022-00245-6}
\end{barticle}
\endbibitem

\bibitem[\protect\citeauthoryear{Khan et~al.}{2022}]{khan2022survey}
\begin{bchapter}
\bauthor{\bsnm{Khan}, \binits{A.}},
\bauthor{\bsnm{AlBarri}, \binits{S.}},
\bauthor{\bsnm{Manzoor}, \binits{M.A.}}:
\bctitle{Contrastive self-supervised learning: A survey on different architectures}.
In: \bbtitle{2022 2nd International Conference on Artificial Intelligence (ICAI)},
pp. \bfpage{1}--\blpage{6}
(\byear{2022}).
\doiurl{10.1109/ICAI55435.2022.9773725}
\end{bchapter}
\endbibitem

\bibitem[\protect\citeauthoryear{Liu et~al.}{2021}]{liu2023survey}
\begin{botherref}
\oauthor{\bsnm{Liu}, \binits{X.}},
\oauthor{\bsnm{Zhang}, \binits{F.}},
\oauthor{\bsnm{Hou}, \binits{Z.}},
\oauthor{\bsnm{Mian}, \binits{L.}},
\oauthor{\bsnm{Wang}, \binits{Z.}},
\oauthor{\bsnm{Zhang}, \binits{J.}},
\oauthor{\bsnm{Tang}, \binits{J.}}:
Self-supervised learning: Generative or contrastive.
IEEE Transactions on Knowledge and Data Engineering,
1--1
(2021)
\doiurl{10.1109/tkde.2021.3090866}
\end{botherref}
\endbibitem

\bibitem[\protect\citeauthoryear{Hu et~al.}{2024}]{hu2024survey}
\begin{barticle}
\bauthor{\bsnm{Hu}, \binits{H.}},
\bauthor{\bsnm{Wang}, \binits{X.}},
\bauthor{\bsnm{Zhang}, \binits{Y.}},
\bauthor{\bsnm{Chen}, \binits{Q.}},
\bauthor{\bsnm{Guan}, \binits{Q.}}:
\batitle{A comprehensive survey on contrastive learning}.
\bjtitle{Neurocomputing}
\bvolume{610},
\bfpage{128645}
(\byear{2024})
\doiurl{10.1016/j.neucom.2024.128645}
\end{barticle}
\endbibitem

\bibitem[\protect\citeauthoryear{Zhang et~al.}{2016}]{zhang2016colorization}
\begin{botherref}
\oauthor{\bsnm{Zhang}, \binits{R.}},
\oauthor{\bsnm{Isola}, \binits{P.}},
\oauthor{\bsnm{Efros}, \binits{A.A.}}:
Colorful Image Colorization
(2016)
\end{botherref}
\endbibitem

\bibitem[\protect\citeauthoryear{Pathak et~al.}{2016}]{pathak2016context}
\begin{botherref}
\oauthor{\bsnm{Pathak}, \binits{D.}},
\oauthor{\bsnm{Kr{\"{a}}henb{\"{u}}hl}, \binits{P.}},
\oauthor{\bsnm{Donahue}, \binits{J.}},
\oauthor{\bsnm{Darrell}, \binits{T.}},
\oauthor{\bsnm{Efros}, \binits{A.A.}}:
Context encoders: Feature learning by inpainting.
CoRR
\textbf{abs/1604.07379}
(2016)
{\href{https://arxiv.org/abs/1604.07379}{{1604.07379}}}
\end{botherref}
\endbibitem

\bibitem[\protect\citeauthoryear{Noroozi et~al.}{2018}]{noroozi2018boosting}
\begin{botherref}
\oauthor{\bsnm{Noroozi}, \binits{M.}},
\oauthor{\bsnm{Vinjimoor}, \binits{A.}},
\oauthor{\bsnm{Favaro}, \binits{P.}},
\oauthor{\bsnm{Pirsiavash}, \binits{H.}}:
Boosting self-supervised learning via knowledge transfer.
CoRR
\textbf{abs/1805.00385}
(2018)
{\href{https://arxiv.org/abs/1805.00385}{{1805.00385}}}
\end{botherref}
\endbibitem

\bibitem[\protect\citeauthoryear{Gidaris et~al.}{2018}]{gidaris2018rotnet}
\begin{botherref}
\oauthor{\bsnm{Gidaris}, \binits{S.}},
\oauthor{\bsnm{Singh}, \binits{P.}},
\oauthor{\bsnm{Komodakis}, \binits{N.}}:
Unsupervised Representation Learning by Predicting Image Rotations
(2018)
\end{botherref}
\endbibitem

\bibitem[\protect\citeauthoryear{Bao et~al.}{2022}]{bao2022beit}
\begin{botherref}
\oauthor{\bsnm{Bao}, \binits{H.}},
\oauthor{\bsnm{Dong}, \binits{L.}},
\oauthor{\bsnm{Piao}, \binits{S.}},
\oauthor{\bsnm{Wei}, \binits{F.}}:
BEiT: BERT Pre-Training of Image Transformers
(2022)
\end{botherref}
\endbibitem

\bibitem[\protect\citeauthoryear{Simonyan and Zisserman}{2015}]{simonyan2015vgg}
\begin{botherref}
\oauthor{\bsnm{Simonyan}, \binits{K.}},
\oauthor{\bsnm{Zisserman}, \binits{A.}}:
Very Deep Convolutional Networks for Large-Scale Image Recognition
(2015)
\end{botherref}
\endbibitem

\bibitem[\protect\citeauthoryear{He et~al.}{2015}]{he2015resnet}
\begin{botherref}
\oauthor{\bsnm{He}, \binits{K.}},
\oauthor{\bsnm{Zhang}, \binits{X.}},
\oauthor{\bsnm{Ren}, \binits{S.}},
\oauthor{\bsnm{Sun}, \binits{J.}}:
Deep Residual Learning for Image Recognition
(2015)
\end{botherref}
\endbibitem

\bibitem[\protect\citeauthoryear{Zagoruyko and Komodakis}{2016}]{zagoruyko2016wide}
\begin{botherref}
\oauthor{\bsnm{Zagoruyko}, \binits{S.}},
\oauthor{\bsnm{Komodakis}, \binits{N.}}:
Wide Residual Networks
(2016)
\end{botherref}
\endbibitem

\bibitem[\protect\citeauthoryear{Liu et~al.}{2021}]{liu2021swin}
\begin{botherref}
\oauthor{\bsnm{Liu}, \binits{Z.}},
\oauthor{\bsnm{Lin}, \binits{Y.}},
\oauthor{\bsnm{Cao}, \binits{Y.}},
\oauthor{\bsnm{Hu}, \binits{H.}},
\oauthor{\bsnm{Wei}, \binits{Y.}},
\oauthor{\bsnm{Zhang}, \binits{Z.}},
\oauthor{\bsnm{Lin}, \binits{S.}},
\oauthor{\bsnm{Guo}, \binits{B.}}:
Swin Transformer: Hierarchical Vision Transformer using Shifted Windows
(2021)
\end{botherref}
\endbibitem

\bibitem[\protect\citeauthoryear{El-Nouby et~al.}{2021}]{elnouby2021xcit}
\begin{botherref}
\oauthor{\bsnm{El-Nouby}, \binits{A.}},
\oauthor{\bsnm{Touvron}, \binits{H.}},
\oauthor{\bsnm{Caron}, \binits{M.}},
\oauthor{\bsnm{Bojanowski}, \binits{P.}},
\oauthor{\bsnm{Douze}, \binits{M.}},
\oauthor{\bsnm{Joulin}, \binits{A.}},
\oauthor{\bsnm{Laptev}, \binits{I.}},
\oauthor{\bsnm{Neverova}, \binits{N.}},
\oauthor{\bsnm{Synnaeve}, \binits{G.}},
\oauthor{\bsnm{Verbeek}, \binits{J.}},
\oauthor{\bsnm{Jegou}, \binits{H.}}:
XCiT: Cross-Covariance Image Transformers
(2021)
\end{botherref}
\endbibitem

\bibitem[\protect\citeauthoryear{Chen and He}{2020}]{chen2020simsiam}
\begin{botherref}
\oauthor{\bsnm{Chen}, \binits{X.}},
\oauthor{\bsnm{He}, \binits{K.}}:
Exploring Simple Siamese Representation Learning
(2020)
\end{botherref}
\endbibitem

\bibitem[\protect\citeauthoryear{Bromley et~al.}{1993}]{bromley1998signature}
\begin{bchapter}
\bauthor{\bsnm{Bromley}, \binits{J.}},
\bauthor{\bsnm{Guyon}, \binits{I.}},
\bauthor{\bsnm{LeCun}, \binits{Y.}},
\bauthor{\bsnm{S\"{a}ckinger}, \binits{E.}},
\bauthor{\bsnm{Shah}, \binits{R.}}:
\bctitle{Signature verification using a "siamese" time delay neural network}.
In: \beditor{\bsnm{Cowan}, \binits{J.}},
\beditor{\bsnm{Tesauro}, \binits{G.}},
\beditor{\bsnm{Alspector}, \binits{J.}} (eds.)
\bbtitle{Advances in Neural Information Processing Systems},
vol. \bseriesno{6}.
\bpublisher{Morgan-Kaufmann}, \blocation{???}
(\byear{1993}).
\burl{https://proceedings.neurips.cc/paper_files/paper/1993/file/288cc0ff022877bd3df94bc9360b9c5d-Paper.pdf}
\end{bchapter}
\endbibitem

\bibitem[\protect\citeauthoryear{Chopra et~al.}{2005}]{chopra2005learning}
\begin{bchapter}
\bauthor{\bsnm{Chopra}, \binits{S.}},
\bauthor{\bsnm{Hadsell}, \binits{R.}},
\bauthor{\bsnm{LeCun}, \binits{Y.}}:
\bctitle{Learning a similarity metric discriminatively, with application to face verification}.
In: \bbtitle{2005 IEEE Computer Society Conference on Computer Vision and Pattern Recognition (CVPR'05)},
vol. \bseriesno{1},
pp. \bfpage{539}--\blpage{5461}
(\byear{2005}).
\doiurl{10.1109/CVPR.2005.202}
\end{bchapter}
\endbibitem

\bibitem[\protect\citeauthoryear{Weinberger and Saul}{2009}]{weinberg2009distance}
\begin{barticle}
\bauthor{\bsnm{Weinberger}, \binits{K.Q.}},
\bauthor{\bsnm{Saul}, \binits{L.K.}}:
\batitle{Distance metric learning for large margin nearest neighbor classification}.
\bjtitle{Journal of Machine Learning Research}
\bvolume{10}(\bissue{9}),
\bfpage{207}--\blpage{244}
(\byear{2009})
\end{barticle}
\endbibitem

\bibitem[\protect\citeauthoryear{Chechik et~al.}{2009}]{chechik2009learning}
\begin{bchapter}
\bauthor{\bsnm{Chechik}, \binits{G.}},
\bauthor{\bsnm{Sharma}, \binits{V.}},
\bauthor{\bsnm{Shalit}, \binits{U.}},
\bauthor{\bsnm{Bengio}, \binits{S.}}:
\bctitle{Large scale online learning of image similarity through ranking}.
In: \beditor{\bsnm{Araujo}, \binits{H.}},
\beditor{\bsnm{Mendon{\c{c}}a}, \binits{A.M.}},
\beditor{\bsnm{Pinho}, \binits{A.J.}},
\beditor{\bsnm{Torres}, \binits{M.I.}} (eds.)
\bbtitle{Pattern Recognition and Image Analysis},
pp. \bfpage{11}--\blpage{14}.
\bpublisher{Springer},
\blocation{Berlin, Heidelberg}
(\byear{2009})
\end{bchapter}
\endbibitem

\bibitem[\protect\citeauthoryear{Sohn}{2016}]{sohn2016improved}
\begin{bchapter}
\bauthor{\bsnm{Sohn}, \binits{K.}}:
\bctitle{Improved deep metric learning with multi-class n-pair loss objective}.
In: \beditor{\bsnm{Lee}, \binits{D.}},
\beditor{\bsnm{Sugiyama}, \binits{M.}},
\beditor{\bsnm{Luxburg}, \binits{U.}},
\beditor{\bsnm{Guyon}, \binits{I.}},
\beditor{\bsnm{Garnett}, \binits{R.}} (eds.)
\bbtitle{Advances in Neural Information Processing Systems},
vol. \bseriesno{29}.
\bpublisher{Curran Associates, Inc.}, \blocation{???}
(\byear{2016}).
\burl{https://proceedings.neurips.cc/paper_files/paper/2016/file/6b180037abbebea991d8b1232f8a8ca9-Paper.pdf}
\end{bchapter}
\endbibitem

\bibitem[\protect\citeauthoryear{Wu et~al.}{2018}]{wu2018instdis}
\begin{botherref}
\oauthor{\bsnm{Wu}, \binits{Z.}},
\oauthor{\bsnm{Xiong}, \binits{Y.}},
\oauthor{\bsnm{Yu}, \binits{S.}},
\oauthor{\bsnm{Lin}, \binits{D.}}:
Unsupervised Feature Learning via Non-Parametric Instance-level Discrimination
(2018)
\end{botherref}
\endbibitem

\bibitem[\protect\citeauthoryear{Yeh et~al.}{2022}]{yeh2022dcl}
\begin{botherref}
\oauthor{\bsnm{Yeh}, \binits{C.-H.}},
\oauthor{\bsnm{Hong}, \binits{C.-Y.}},
\oauthor{\bsnm{Hsu}, \binits{Y.-C.}},
\oauthor{\bsnm{Liu}, \binits{T.-L.}},
\oauthor{\bsnm{Chen}, \binits{Y.}},
\oauthor{\bsnm{LeCun}, \binits{Y.}}:
Decoupled Contrastive Learning
(2022)
\end{botherref}
\endbibitem

\bibitem[\protect\citeauthoryear{Dwibedi et~al.}{2021}]{dwibedi2021nnclr}
\begin{botherref}
\oauthor{\bsnm{Dwibedi}, \binits{D.}},
\oauthor{\bsnm{Aytar}, \binits{Y.}},
\oauthor{\bsnm{Tompson}, \binits{J.}},
\oauthor{\bsnm{Sermanet}, \binits{P.}},
\oauthor{\bsnm{Zisserman}, \binits{A.}}:
With a Little Help from My Friends: Nearest-Neighbor Contrastive Learning of Visual Representations
(2021)
\end{botherref}
\endbibitem

\bibitem[\protect\citeauthoryear{Mitrovic et~al.}{2020}]{mitrovic2020relic}
\begin{botherref}
\oauthor{\bsnm{Mitrovic}, \binits{J.}},
\oauthor{\bsnm{McWilliams}, \binits{B.}},
\oauthor{\bsnm{Walker}, \binits{J.}},
\oauthor{\bsnm{Buesing}, \binits{L.}},
\oauthor{\bsnm{Blundell}, \binits{C.}}:
Representation Learning via Invariant Causal Mechanisms
(2020)
\end{botherref}
\endbibitem

\bibitem[\protect\citeauthoryear{Li et~al.}{2021}]{li2021pcl}
\begin{botherref}
\oauthor{\bsnm{Li}, \binits{J.}},
\oauthor{\bsnm{Zhou}, \binits{P.}},
\oauthor{\bsnm{Xiong}, \binits{C.}},
\oauthor{\bsnm{Hoi}, \binits{S.C.H.}}:
Prototypical Contrastive Learning of Unsupervised Representations
(2021)
\end{botherref}
\endbibitem

\bibitem[\protect\citeauthoryear{Caron et~al.}{2020}]{caron2020swav}
\begin{bchapter}
\bauthor{\bsnm{Caron}, \binits{M.}},
\bauthor{\bsnm{Misra}, \binits{I.}},
\bauthor{\bsnm{Mairal}, \binits{J.}},
\bauthor{\bsnm{Goyal}, \binits{P.}},
\bauthor{\bsnm{Bojanowski}, \binits{P.}},
\bauthor{\bsnm{Joulin}, \binits{A.}}:
\bctitle{Unsupervised learning of visual features by contrasting cluster assignments}.
In: \beditor{\bsnm{Larochelle}, \binits{H.}},
\beditor{\bsnm{Ranzato}, \binits{M.}},
\beditor{\bsnm{Hadsell}, \binits{R.}},
\beditor{\bsnm{Balcan}, \binits{M.F.}},
\beditor{\bsnm{Lin}, \binits{H.}} (eds.)
\bbtitle{Advances in Neural Information Processing Systems},
vol. \bseriesno{33},
pp. \bfpage{9912}--\blpage{9924}.
\bpublisher{Curran Associates, Inc.}, \blocation{???}
(\byear{2020}).
\burl{https://proceedings.neurips.cc/paper_files/paper/2020/file/70feb62b69f16e0238f741fab228fec2-Paper.pdf}
\end{bchapter}
\endbibitem

\bibitem[\protect\citeauthoryear{Cubuk et~al.}{2019a}]{cubuk2019autoaugment}
\begin{botherref}
\oauthor{\bsnm{Cubuk}, \binits{E.D.}},
\oauthor{\bsnm{Zoph}, \binits{B.}},
\oauthor{\bsnm{Mane}, \binits{D.}},
\oauthor{\bsnm{Vasudevan}, \binits{V.}},
\oauthor{\bsnm{Le}, \binits{Q.V.}}:
AutoAugment: Learning Augmentation Policies from Data
(2019)
\end{botherref}
\endbibitem

\bibitem[\protect\citeauthoryear{Cubuk et~al.}{2019b}]{cubuk2019randaugment}
\begin{botherref}
\oauthor{\bsnm{Cubuk}, \binits{E.D.}},
\oauthor{\bsnm{Zoph}, \binits{B.}},
\oauthor{\bsnm{Shlens}, \binits{J.}},
\oauthor{\bsnm{Le}, \binits{Q.V.}}:
RandAugment: Practical automated data augmentation with a reduced search space
(2019)
\end{botherref}
\endbibitem

\bibitem[\protect\citeauthoryear{Caron et~al.}{2019}]{caron2019deepcluster}
\begin{botherref}
\oauthor{\bsnm{Caron}, \binits{M.}},
\oauthor{\bsnm{Bojanowski}, \binits{P.}},
\oauthor{\bsnm{Joulin}, \binits{A.}},
\oauthor{\bsnm{Douze}, \binits{M.}}:
Deep Clustering for Unsupervised Learning of Visual Features
(2019)
\end{botherref}
\endbibitem

\bibitem[\protect\citeauthoryear{Zbontar et~al.}{2021}]{zbontar2021barlowtwins}
\begin{botherref}
\oauthor{\bsnm{Zbontar}, \binits{J.}},
\oauthor{\bsnm{Jing}, \binits{L.}},
\oauthor{\bsnm{Misra}, \binits{I.}},
\oauthor{\bsnm{LeCun}, \binits{Y.}},
\oauthor{\bsnm{Deny}, \binits{S.}}:
Barlow Twins: Self-Supervised Learning via Redundancy Reduction
(2021)
\end{botherref}
\endbibitem

\bibitem[\protect\citeauthoryear{Larsson et~al.}{2016}]{larsson2016learning}
\begin{botherref}
\oauthor{\bsnm{Larsson}, \binits{G.}},
\oauthor{\bsnm{Maire}, \binits{M.}},
\oauthor{\bsnm{Shakhnarovich}, \binits{G.}}:
Learning representations for automatic colorization.
CoRR
\textbf{abs/1603.06668}
(2016)
{\href{https://arxiv.org/abs/1603.06668}{{1603.06668}}}
\end{botherref}
\endbibitem

\bibitem[\protect\citeauthoryear{Caron et~al.}{2019}]{caron2019deepercluster}
\begin{botherref}
\oauthor{\bsnm{Caron}, \binits{M.}},
\oauthor{\bsnm{Bojanowski}, \binits{P.}},
\oauthor{\bsnm{Mairal}, \binits{J.}},
\oauthor{\bsnm{Joulin}, \binits{A.}}:
Unsupervised Pre-Training of Image Features on Non-Curated Data
(2019)
\end{botherref}
\endbibitem

\bibitem[\protect\citeauthoryear{Zhang et~al.}{2017}]{zhang2017splitbrain}
\begin{botherref}
\oauthor{\bsnm{Zhang}, \binits{R.}},
\oauthor{\bsnm{Isola}, \binits{P.}},
\oauthor{\bsnm{Efros}, \binits{A.A.}}:
Split-Brain Autoencoders: Unsupervised Learning by Cross-Channel Prediction
(2017)
\end{botherref}
\endbibitem

\bibitem[\protect\citeauthoryear{Donahue et~al.}{2016}]{donahue2016adversarial}
\begin{botherref}
\oauthor{\bsnm{Donahue}, \binits{J.}},
\oauthor{\bsnm{Kr{\"{a}}henb{\"{u}}hl}, \binits{P.}},
\oauthor{\bsnm{Darrell}, \binits{T.}}:
Adversarial feature learning.
CoRR
\textbf{abs/1605.09782}
(2016)
{\href{https://arxiv.org/abs/1605.09782}{{1605.09782}}}
\end{botherref}
\endbibitem

\bibitem[\protect\citeauthoryear{Jenni et~al.}{2020}]{jenni2020lci}
\begin{botherref}
\oauthor{\bsnm{Jenni}, \binits{S.}},
\oauthor{\bsnm{Jin}, \binits{H.}},
\oauthor{\bsnm{Favaro}, \binits{P.}}:
Steering Self-Supervised Feature Learning Beyond Local Pixel Statistics
(2020)
\end{botherref}
\endbibitem

\bibitem[\protect\citeauthoryear{Khosla et~al.}{2021}]{khosla2021supcon}
\begin{botherref}
\oauthor{\bsnm{Khosla}, \binits{P.}},
\oauthor{\bsnm{Teterwak}, \binits{P.}},
\oauthor{\bsnm{Wang}, \binits{C.}},
\oauthor{\bsnm{Sarna}, \binits{A.}},
\oauthor{\bsnm{Tian}, \binits{Y.}},
\oauthor{\bsnm{Isola}, \binits{P.}},
\oauthor{\bsnm{Maschinot}, \binits{A.}},
\oauthor{\bsnm{Liu}, \binits{C.}},
\oauthor{\bsnm{Krishnan}, \binits{D.}}:
Supervised Contrastive Learning
(2021)
\end{botherref}
\endbibitem

\bibitem[\protect\citeauthoryear{Tian et~al.}{2020a}]{tian2020cmc}
\begin{botherref}
\oauthor{\bsnm{Tian}, \binits{Y.}},
\oauthor{\bsnm{Krishnan}, \binits{D.}},
\oauthor{\bsnm{Isola}, \binits{P.}}:
Contrastive Multiview Coding
(2020)
\end{botherref}
\endbibitem

\bibitem[\protect\citeauthoryear{Tian et~al.}{2020b}]{tian2020infomin}
\begin{bchapter}
\bauthor{\bsnm{Tian}, \binits{Y.}},
\bauthor{\bsnm{Sun}, \binits{C.}},
\bauthor{\bsnm{Poole}, \binits{B.}},
\bauthor{\bsnm{Krishnan}, \binits{D.}},
\bauthor{\bsnm{Schmid}, \binits{C.}},
\bauthor{\bsnm{Isola}, \binits{P.}}:
\bctitle{What makes for good views for contrastive learning?}
In: \beditor{\bsnm{Larochelle}, \binits{H.}},
\beditor{\bsnm{Ranzato}, \binits{M.}},
\beditor{\bsnm{Hadsell}, \binits{R.}},
\beditor{\bsnm{Balcan}, \binits{M.F.}},
\beditor{\bsnm{Lin}, \binits{H.}} (eds.)
\bbtitle{Advances in Neural Information Processing Systems},
vol. \bseriesno{33},
pp. \bfpage{6827}--\blpage{6839}.
\bpublisher{Curran Associates, Inc.}, \blocation{???}
(\byear{2020}).
\burl{https://proceedings.neurips.cc/paper_files/paper/2020/file/4c2e5eaae9152079b9e95845750bb9ab-Paper.pdf}
\end{bchapter}
\endbibitem

\bibitem[\protect\citeauthoryear{Misra and van~der Maaten}{2019}]{misra2019pirl}
\begin{botherref}
\oauthor{\bsnm{Misra}, \binits{I.}},
\oauthor{\bsnm{Maaten}, \binits{L.}}:
Self-Supervised Learning of Pretext-Invariant Representations
(2019)
\end{botherref}
\endbibitem

\bibitem[\protect\citeauthoryear{Ye et~al.}{2019}]{ye2019uel}
\begin{botherref}
\oauthor{\bsnm{Ye}, \binits{M.}},
\oauthor{\bsnm{Zhang}, \binits{X.}},
\oauthor{\bsnm{Yuen}, \binits{P.C.}},
\oauthor{\bsnm{Chang}, \binits{S.-F.}}:
Unsupervised Embedding Learning via Invariant and Spreading Instance Feature
(2019)
\end{botherref}
\endbibitem

\bibitem[\protect\citeauthoryear{Hjelm et~al.}{2019}]{hjelm2019dim}
\begin{botherref}
\oauthor{\bsnm{Hjelm}, \binits{R.D.}},
\oauthor{\bsnm{Fedorov}, \binits{A.}},
\oauthor{\bsnm{Lavoie-Marchildon}, \binits{S.}},
\oauthor{\bsnm{Grewal}, \binits{K.}},
\oauthor{\bsnm{Bachman}, \binits{P.}},
\oauthor{\bsnm{Trischler}, \binits{A.}},
\oauthor{\bsnm{Bengio}, \binits{Y.}}:
Learning deep representations by mutual information estimation and maximization
(2019)
\end{botherref}
\endbibitem

\bibitem[\protect\citeauthoryear{Bachman et~al.}{2019}]{bachman2019amdim}
\begin{botherref}
\oauthor{\bsnm{Bachman}, \binits{P.}},
\oauthor{\bsnm{Hjelm}, \binits{R.D.}},
\oauthor{\bsnm{Buchwalter}, \binits{W.}}:
Learning Representations by Maximizing Mutual Information Across Views
(2019)
\end{botherref}
\endbibitem

\bibitem[\protect\citeauthoryear{Hénaff et~al.}{2020}]{henaff2020cpcv2}
\begin{botherref}
\oauthor{\bsnm{Hénaff}, \binits{O.J.}},
\oauthor{\bsnm{Srinivas}, \binits{A.}},
\oauthor{\bsnm{Fauw}, \binits{J.D.}},
\oauthor{\bsnm{Razavi}, \binits{A.}},
\oauthor{\bsnm{Doersch}, \binits{C.}},
\oauthor{\bsnm{Eslami}, \binits{S.M.A.}},
\oauthor{\bsnm{Oord}, \binits{A.}}:
Data-Efficient Image Recognition with Contrastive Predictive Coding
(2020)
\end{botherref}
\endbibitem

\bibitem[\protect\citeauthoryear{Chen et~al.}{2020}]{chen2020simclrv2}
\begin{botherref}
\oauthor{\bsnm{Chen}, \binits{T.}},
\oauthor{\bsnm{Kornblith}, \binits{S.}},
\oauthor{\bsnm{Swersky}, \binits{K.}},
\oauthor{\bsnm{Norouzi}, \binits{M.}},
\oauthor{\bsnm{Hinton}, \binits{G.}}:
Big Self-Supervised Models are Strong Semi-Supervised Learners
(2020)
\end{botherref}
\endbibitem

\bibitem[\protect\citeauthoryear{Chakraborty et~al.}{2020}]{chakraborty2020gsimclr}
\begin{botherref}
\oauthor{\bsnm{Chakraborty}, \binits{S.}},
\oauthor{\bsnm{Gosthipaty}, \binits{A.R.}},
\oauthor{\bsnm{Paul}, \binits{S.}}:
G-SimCLR : Self-Supervised Contrastive Learning with Guided Projection via Pseudo Labelling
(2020)
\end{botherref}
\endbibitem

\bibitem[\protect\citeauthoryear{Kumar et~al.}{2014}]{kumar2014static}
\begin{botherref}
\oauthor{\bsnm{Kumar}, \binits{V.}},
\oauthor{\bsnm{Nandi}, \binits{G.C.}},
\oauthor{\bsnm{Kala}, \binits{R.}}:
Static hand gesture recognition using stacked denoising sparse autoencoders.
2014 Seventh International Conference on Contemporary Computing (IC3),
99--104
(2014)
\end{botherref}
\endbibitem

\bibitem[\protect\citeauthoryear{Yang et~al.}{2022}]{yang2022simcl}
\begin{bchapter}
\bauthor{\bsnm{Yang}, \binits{H.}},
\bauthor{\bsnm{Ding}, \binits{X.}},
\bauthor{\bsnm{Wang}, \binits{J.}},
\bauthor{\bsnm{Li}, \binits{J.}}:
\bctitle{Simcl:simple contrastive learning for image classification}.
In: \bbtitle{Proceedings of the 5th International Conference on Big Data Technologies}.
\bsertitle{ICBDT 2022}.
\bpublisher{ACM}, \blocation{???}
(\byear{2022}).
\doiurl{10.1145/3565291.3565335} .
\burl{http://dx.doi.org/10.1145/3565291.3565335}
\end{bchapter}
\endbibitem

\bibitem[\protect\citeauthoryear{Lee et~al.}{2021}]{lee2021cbyolcsimclr}
\begin{botherref}
\oauthor{\bsnm{Lee}, \binits{K.-H.}},
\oauthor{\bsnm{Arnab}, \binits{A.}},
\oauthor{\bsnm{Guadarrama}, \binits{S.}},
\oauthor{\bsnm{Canny}, \binits{J.}},
\oauthor{\bsnm{Fischer}, \binits{I.}}:
Compressive Visual Representations
(2021)
\end{botherref}
\endbibitem

\bibitem[\protect\citeauthoryear{Jing et~al.}{2022}]{jing2022directclr}
\begin{botherref}
\oauthor{\bsnm{Jing}, \binits{L.}},
\oauthor{\bsnm{Vincent}, \binits{P.}},
\oauthor{\bsnm{LeCun}, \binits{Y.}},
\oauthor{\bsnm{Tian}, \binits{Y.}}:
Understanding Dimensional Collapse in Contrastive Self-supervised Learning
(2022).
\url{https://arxiv.org/abs/2110.09348}
\end{botherref}
\endbibitem

\bibitem[\protect\citeauthoryear{Chen et~al.}{2020}]{chen2020mocov2}
\begin{botherref}
\oauthor{\bsnm{Chen}, \binits{X.}},
\oauthor{\bsnm{Fan}, \binits{H.}},
\oauthor{\bsnm{Girshick}, \binits{R.}},
\oauthor{\bsnm{He}, \binits{K.}}:
Improved Baselines with Momentum Contrastive Learning
(2020)
\end{botherref}
\endbibitem

\bibitem[\protect\citeauthoryear{Chen et~al.}{2021}]{chen2021mocov3}
\begin{botherref}
\oauthor{\bsnm{Chen}, \binits{X.}},
\oauthor{\bsnm{Xie}, \binits{S.}},
\oauthor{\bsnm{He}, \binits{K.}}:
An Empirical Study of Training Self-Supervised Vision Transformers
(2021)
\end{botherref}
\endbibitem

\bibitem[\protect\citeauthoryear{Loshchilov and Hutter}{2019}]{loshchilov2019adamw}
\begin{botherref}
\oauthor{\bsnm{Loshchilov}, \binits{I.}},
\oauthor{\bsnm{Hutter}, \binits{F.}}:
Decoupled Weight Decay Regularization
(2019)
\end{botherref}
\endbibitem

\bibitem[\protect\citeauthoryear{You et~al.}{2017}]{you2017lars}
\begin{botherref}
\oauthor{\bsnm{You}, \binits{Y.}},
\oauthor{\bsnm{Gitman}, \binits{I.}},
\oauthor{\bsnm{Ginsburg}, \binits{B.}}:
Large Batch Training of Convolutional Networks
(2017)
\end{botherref}
\endbibitem

\bibitem[\protect\citeauthoryear{Kalantidis et~al.}{2020}]{kalantidis2020mochi}
\begin{botherref}
\oauthor{\bsnm{Kalantidis}, \binits{Y.}},
\oauthor{\bsnm{Sariyildiz}, \binits{M.B.}},
\oauthor{\bsnm{Pion}, \binits{N.}},
\oauthor{\bsnm{Weinzaepfel}, \binits{P.}},
\oauthor{\bsnm{Larlus}, \binits{D.}}:
Hard Negative Mixing for Contrastive Learning
(2020)
\end{botherref}
\endbibitem

\bibitem[\protect\citeauthoryear{Giakoumoglou and Stathaki}{2024}]{giakoumoglou2024synco}
\begin{botherref}
\oauthor{\bsnm{Giakoumoglou}, \binits{N.}},
\oauthor{\bsnm{Stathaki}, \binits{T.}}:
SynCo: Synthetic Hard Negatives for Contrastive Visual Representation Learning
(2024).
\url{https://arxiv.org/abs/2410.02401}
\end{botherref}
\endbibitem

\bibitem[\protect\citeauthoryear{Zhang et~al.}{2022}]{zhang2022simco}
\begin{botherref}
\oauthor{\bsnm{Zhang}, \binits{C.}},
\oauthor{\bsnm{Zhang}, \binits{K.}},
\oauthor{\bsnm{Pham}, \binits{T.X.}},
\oauthor{\bsnm{Niu}, \binits{A.}},
\oauthor{\bsnm{Qiao}, \binits{Z.}},
\oauthor{\bsnm{Yoo}, \binits{C.D.}},
\oauthor{\bsnm{Kweon}, \binits{I.S.}}:
Dual Temperature Helps Contrastive Learning Without Many Negative Samples: Towards Understanding and Simplifying MoCo
(2022)
\end{botherref}
\endbibitem

\bibitem[\protect\citeauthoryear{Xiao et~al.}{2021}]{xiao2021looc}
\begin{botherref}
\oauthor{\bsnm{Xiao}, \binits{T.}},
\oauthor{\bsnm{Wang}, \binits{X.}},
\oauthor{\bsnm{Efros}, \binits{A.A.}},
\oauthor{\bsnm{Darrell}, \binits{T.}}:
What Should Not Be Contrastive in Contrastive Learning
(2021)
\end{botherref}
\endbibitem

\bibitem[\protect\citeauthoryear{Zhu et~al.}{2022}]{zhu2022tico}
\begin{botherref}
\oauthor{\bsnm{Zhu}, \binits{J.}},
\oauthor{\bsnm{Moraes}, \binits{R.M.}},
\oauthor{\bsnm{Karakulak}, \binits{S.}},
\oauthor{\bsnm{Sobol}, \binits{V.}},
\oauthor{\bsnm{Canziani}, \binits{A.}},
\oauthor{\bsnm{LeCun}, \binits{Y.}}:
TiCo: Transformation Invariance and Covariance Contrast for Self-Supervised Visual Representation Learning
(2022)
\end{botherref}
\endbibitem

\bibitem[\protect\citeauthoryear{Zheng et~al.}{2021}]{zheng2021ressl}
\begin{botherref}
\oauthor{\bsnm{Zheng}, \binits{M.}},
\oauthor{\bsnm{You}, \binits{S.}},
\oauthor{\bsnm{Wang}, \binits{F.}},
\oauthor{\bsnm{Qian}, \binits{C.}},
\oauthor{\bsnm{Zhang}, \binits{C.}},
\oauthor{\bsnm{Wang}, \binits{X.}},
\oauthor{\bsnm{Xu}, \binits{C.}}:
ReSSL: Relational Self-Supervised Learning with Weak Augmentation
(2021)
\end{botherref}
\endbibitem

\bibitem[\protect\citeauthoryear{Xie et~al.}{2021}]{xie2021moby}
\begin{botherref}
\oauthor{\bsnm{Xie}, \binits{Z.}},
\oauthor{\bsnm{Lin}, \binits{Y.}},
\oauthor{\bsnm{Yao}, \binits{Z.}},
\oauthor{\bsnm{Zhang}, \binits{Z.}},
\oauthor{\bsnm{Dai}, \binits{Q.}},
\oauthor{\bsnm{Cao}, \binits{Y.}},
\oauthor{\bsnm{Hu}, \binits{H.}}:
Self-Supervised Learning with Swin Transformers
(2021)
\end{botherref}
\endbibitem

\bibitem[\protect\citeauthoryear{Tian et~al.}{2021}]{tian2021dnc}
\begin{botherref}
\oauthor{\bsnm{Tian}, \binits{Y.}},
\oauthor{\bsnm{Henaff}, \binits{O.J.}},
\oauthor{\bsnm{Oord}, \binits{A.}}:
Divide and Contrast: Self-supervised Learning from Uncurated Data
(2021)
\end{botherref}
\endbibitem

\bibitem[\protect\citeauthoryear{Koohpayegani et~al.}{2021}]{koohpayegani2021msf}
\begin{botherref}
\oauthor{\bsnm{Koohpayegani}, \binits{S.A.}},
\oauthor{\bsnm{Tejankar}, \binits{A.}},
\oauthor{\bsnm{Pirsiavash}, \binits{H.}}:
Mean Shift for Self-Supervised Learning
(2021)
\end{botherref}
\endbibitem

\bibitem[\protect\citeauthoryear{Estepa et~al.}{2023}]{estepa2023all4one}
\begin{botherref}
\oauthor{\bsnm{Estepa}, \binits{I.G.}},
\oauthor{\bsnm{Sarasúa}, \binits{I.}},
\oauthor{\bsnm{Nagarajan}, \binits{B.}},
\oauthor{\bsnm{Radeva}, \binits{P.}}:
All4One: Symbiotic Neighbour Contrastive Learning via Self-Attention and Redundancy Reduction
(2023)
\end{botherref}
\endbibitem

\bibitem[\protect\citeauthoryear{Azabou et~al.}{2021}]{azabou2021myow}
\begin{botherref}
\oauthor{\bsnm{Azabou}, \binits{M.}},
\oauthor{\bsnm{Azar}, \binits{M.G.}},
\oauthor{\bsnm{Liu}, \binits{R.}},
\oauthor{\bsnm{Lin}, \binits{C.-H.}},
\oauthor{\bsnm{Johnson}, \binits{E.C.}},
\oauthor{\bsnm{Bhaskaran-Nair}, \binits{K.}},
\oauthor{\bsnm{Dabagia}, \binits{M.}},
\oauthor{\bsnm{Avila-Pires}, \binits{B.}},
\oauthor{\bsnm{Kitchell}, \binits{L.}},
\oauthor{\bsnm{Hengen}, \binits{K.B.}},
\oauthor{\bsnm{Gray-Roncal}, \binits{W.}},
\oauthor{\bsnm{Valko}, \binits{M.}},
\oauthor{\bsnm{Dyer}, \binits{E.L.}}:
Mine Your Own vieW: Self-Supervised Learning Through Across-Sample Prediction
(2021)
\end{botherref}
\endbibitem

\bibitem[\protect\citeauthoryear{Zhou et~al.}{2022}]{zhou2022mugs}
\begin{botherref}
\oauthor{\bsnm{Zhou}, \binits{P.}},
\oauthor{\bsnm{Zhou}, \binits{Y.}},
\oauthor{\bsnm{Si}, \binits{C.}},
\oauthor{\bsnm{Yu}, \binits{W.}},
\oauthor{\bsnm{Ng}, \binits{T.K.}},
\oauthor{\bsnm{Yan}, \binits{S.}}:
Mugs: A Multi-Granular Self-Supervised Learning Framework
(2022)
\end{botherref}
\endbibitem

\bibitem[\protect\citeauthoryear{Falcon and Cho}{2020}]{falcon2020yadim}
\begin{botherref}
\oauthor{\bsnm{Falcon}, \binits{W.}},
\oauthor{\bsnm{Cho}, \binits{K.}}:
A Framework For Contrastive Self-Supervised Learning And Designing A New Approach
(2020)
\end{botherref}
\endbibitem

\bibitem[\protect\citeauthoryear{Tomasev et~al.}{2022}]{tomasev2022relicv2}
\begin{botherref}
\oauthor{\bsnm{Tomasev}, \binits{N.}},
\oauthor{\bsnm{Bica}, \binits{I.}},
\oauthor{\bsnm{McWilliams}, \binits{B.}},
\oauthor{\bsnm{Buesing}, \binits{L.}},
\oauthor{\bsnm{Pascanu}, \binits{R.}},
\oauthor{\bsnm{Blundell}, \binits{C.}},
\oauthor{\bsnm{Mitrovic}, \binits{J.}}:
Pushing the limits of self-supervised ResNets: Can we outperform supervised learning without labels on ImageNet?
(2022)
\end{botherref}
\endbibitem

\bibitem[\protect\citeauthoryear{Tao et~al.}{2022}]{tao2022unigrad}
\begin{botherref}
\oauthor{\bsnm{Tao}, \binits{C.}},
\oauthor{\bsnm{Wang}, \binits{H.}},
\oauthor{\bsnm{Zhu}, \binits{X.}},
\oauthor{\bsnm{Dong}, \binits{J.}},
\oauthor{\bsnm{Song}, \binits{S.}},
\oauthor{\bsnm{Huang}, \binits{G.}},
\oauthor{\bsnm{Dai}, \binits{J.}}:
Exploring the Equivalence of Siamese Self-Supervised Learning via A Unified Gradient Framework
(2022)
\end{botherref}
\endbibitem

\bibitem[\protect\citeauthoryear{Ho and Vasconcelos}{2020}]{ho2020clae}
\begin{botherref}
\oauthor{\bsnm{Ho}, \binits{C.-H.}},
\oauthor{\bsnm{Vasconcelos}, \binits{N.}}:
Contrastive Learning with Adversarial Examples
(2020).
\url{https://arxiv.org/abs/2010.12050}
\end{botherref}
\endbibitem

\bibitem[\protect\citeauthoryear{Kim et~al.}{2020}]{kim2020rocl}
\begin{botherref}
\oauthor{\bsnm{Kim}, \binits{M.}},
\oauthor{\bsnm{Tack}, \binits{J.}},
\oauthor{\bsnm{Hwang}, \binits{S.J.}}:
Adversarial Self-Supervised Contrastive Learning
(2020).
\url{https://arxiv.org/abs/2006.07589}
\end{botherref}
\endbibitem

\bibitem[\protect\citeauthoryear{Chen et~al.}{2020}]{chen2020sat}
\begin{botherref}
\oauthor{\bsnm{Chen}, \binits{K.}},
\oauthor{\bsnm{Zhou}, \binits{H.}},
\oauthor{\bsnm{Chen}, \binits{Y.}},
\oauthor{\bsnm{Mao}, \binits{X.}},
\oauthor{\bsnm{Li}, \binits{Y.}},
\oauthor{\bsnm{He}, \binits{Y.}},
\oauthor{\bsnm{Xue}, \binits{H.}},
\oauthor{\bsnm{Zhang}, \binits{W.}},
\oauthor{\bsnm{Yu}, \binits{N.}}:
Self-supervised Adversarial Training
(2020).
\url{https://arxiv.org/abs/1911.06470}
\end{botherref}
\endbibitem

\bibitem[\protect\citeauthoryear{Hu et~al.}{2021}]{hu2021adco}
\begin{bchapter}
\bauthor{\bsnm{Hu}, \binits{Q.}},
\bauthor{\bsnm{Wang}, \binits{X.}},
\bauthor{\bsnm{Hu}, \binits{W.}},
\bauthor{\bsnm{Qi}, \binits{G.-J.}}:
\bctitle{Adco: Adversarial contrast for efficient learning of unsupervised representations from self-trained negative adversaries}.
In: \bbtitle{Proceedings of the IEEE/CVF Conference on Computer Vision and Pattern Recognition},
pp. \bfpage{1074}--\blpage{1083}
(\byear{2021})
\end{bchapter}
\endbibitem

\bibitem[\protect\citeauthoryear{Wang et~al.}{2023}]{wang2022caco}
\begin{botherref}
\oauthor{\bsnm{Wang}, \binits{X.}},
\oauthor{\bsnm{Huang}, \binits{Y.}},
\oauthor{\bsnm{Zeng}, \binits{D.}},
\oauthor{\bsnm{Qi}, \binits{G.-J.}}:
Caco: Both positive and negative samples are directly learnable via cooperative-adversarial contrastive learning.
IEEE Transactions on Pattern Analysis and Machine Intelligence
(2023)
\end{botherref}
\endbibitem

\bibitem[\protect\citeauthoryear{Zhang et~al.}{2022}]{zhang2022deacl}
\begin{botherref}
\oauthor{\bsnm{Zhang}, \binits{C.}},
\oauthor{\bsnm{Zhang}, \binits{K.}},
\oauthor{\bsnm{Zhang}, \binits{C.}},
\oauthor{\bsnm{Niu}, \binits{A.}},
\oauthor{\bsnm{Feng}, \binits{J.}},
\oauthor{\bsnm{Yoo}, \binits{C.D.}},
\oauthor{\bsnm{Kweon}, \binits{I.S.}}:
Decoupled Adversarial Contrastive Learning for Self-supervised Adversarial Robustness
(2022).
\url{https://arxiv.org/abs/2207.10899}
\end{botherref}
\endbibitem

\bibitem[\protect\citeauthoryear{Haghighi et~al.}{2022}]{haghighi2022dira}
\begin{botherref}
\oauthor{\bsnm{Haghighi}, \binits{F.}},
\oauthor{\bsnm{Taher}, \binits{M.R.H.}},
\oauthor{\bsnm{Gotway}, \binits{M.B.}},
\oauthor{\bsnm{Liang}, \binits{J.}}:
DiRA: Discriminative, Restorative, and Adversarial Learning for Self-supervised Medical Image Analysis
(2022).
\url{https://arxiv.org/abs/2204.10437}
\end{botherref}
\endbibitem

\bibitem[\protect\citeauthoryear{Shah et~al.}{2021}]{shah2021mmcl}
\begin{botherref}
\oauthor{\bsnm{Shah}, \binits{A.}},
\oauthor{\bsnm{Sra}, \binits{S.}},
\oauthor{\bsnm{Chellappa}, \binits{R.}},
\oauthor{\bsnm{Cherian}, \binits{A.}}:
Max-Margin Contrastive Learning
(2021)
\end{botherref}
\endbibitem

\bibitem[\protect\citeauthoryear{Yerxa et~al.}{2023}]{yerxa2023mmcr}
\begin{botherref}
\oauthor{\bsnm{Yerxa}, \binits{T.}},
\oauthor{\bsnm{Kuang}, \binits{Y.}},
\oauthor{\bsnm{Simoncelli}, \binits{E.}},
\oauthor{\bsnm{Chung}, \binits{S.}}:
Learning Efficient Coding of Natural Images with Maximum Manifold Capacity Representations
(2023)
\end{botherref}
\endbibitem

\bibitem[\protect\citeauthoryear{Rojas-Gomez et~al.}{2024}]{rojasgomez2024sassl}
\begin{botherref}
\oauthor{\bsnm{Rojas-Gomez}, \binits{R.A.}},
\oauthor{\bsnm{Singhal}, \binits{K.}},
\oauthor{\bsnm{Etemad}, \binits{A.}},
\oauthor{\bsnm{Bijamov}, \binits{A.}},
\oauthor{\bsnm{Morningstar}, \binits{W.R.}},
\oauthor{\bsnm{Mansfield}, \binits{P.A.}}:
SASSL: Enhancing Self-Supervised Learning via Neural Style Transfer
(2024).
\url{https://arxiv.org/abs/2312.01187}
\end{botherref}
\endbibitem

\bibitem[\protect\citeauthoryear{Ren et~al.}{2016}]{ren2016fasterrcnn}
\begin{botherref}
\oauthor{\bsnm{Ren}, \binits{S.}},
\oauthor{\bsnm{He}, \binits{K.}},
\oauthor{\bsnm{Girshick}, \binits{R.}},
\oauthor{\bsnm{Sun}, \binits{J.}}:
Faster R-CNN: Towards Real-Time Object Detection with Region Proposal Networks
(2016)
\end{botherref}
\endbibitem

\bibitem[\protect\citeauthoryear{Wang et~al.}{2021}]{wang2021densecl}
\begin{botherref}
\oauthor{\bsnm{Wang}, \binits{X.}},
\oauthor{\bsnm{Zhang}, \binits{R.}},
\oauthor{\bsnm{Shen}, \binits{C.}},
\oauthor{\bsnm{Kong}, \binits{T.}},
\oauthor{\bsnm{Li}, \binits{L.}}:
Dense Contrastive Learning for Self-Supervised Visual Pre-Training
(2021)
\end{botherref}
\endbibitem

\bibitem[\protect\citeauthoryear{Pinheiro et~al.}{2020}]{pinheiro2020vader}
\begin{botherref}
\oauthor{\bsnm{Pinheiro}, \binits{P.O.}},
\oauthor{\bsnm{Almahairi}, \binits{A.}},
\oauthor{\bsnm{Benmalek}, \binits{R.Y.}},
\oauthor{\bsnm{Golemo}, \binits{F.}},
\oauthor{\bsnm{Courville}, \binits{A.}}:
Unsupervised Learning of Dense Visual Representations
(2020)
\end{botherref}
\endbibitem

\bibitem[\protect\citeauthoryear{Xie et~al.}{2021}]{xie2021pixpro}
\begin{botherref}
\oauthor{\bsnm{Xie}, \binits{Z.}},
\oauthor{\bsnm{Lin}, \binits{Y.}},
\oauthor{\bsnm{Zhang}, \binits{Z.}},
\oauthor{\bsnm{Cao}, \binits{Y.}},
\oauthor{\bsnm{Lin}, \binits{S.}},
\oauthor{\bsnm{Hu}, \binits{H.}}:
Propagate Yourself: Exploring Pixel-Level Consistency for Unsupervised Visual Representation Learning
(2021)
\end{botherref}
\endbibitem

\bibitem[\protect\citeauthoryear{Hénaff et~al.}{2021}]{henaff2021detcon}
\begin{botherref}
\oauthor{\bsnm{Hénaff}, \binits{O.J.}},
\oauthor{\bsnm{Koppula}, \binits{S.}},
\oauthor{\bsnm{Alayrac}, \binits{J.-B.}},
\oauthor{\bsnm{Oord}, \binits{A.}},
\oauthor{\bsnm{Vinyals}, \binits{O.}},
\oauthor{\bsnm{Carreira}, \binits{J.}}:
Efficient Visual Pretraining with Contrastive Detection
(2021)
\end{botherref}
\endbibitem

\bibitem[\protect\citeauthoryear{Xiao et~al.}{2021}]{xiao2021resim}
\begin{botherref}
\oauthor{\bsnm{Xiao}, \binits{T.}},
\oauthor{\bsnm{Reed}, \binits{C.J.}},
\oauthor{\bsnm{Wang}, \binits{X.}},
\oauthor{\bsnm{Keutzer}, \binits{K.}},
\oauthor{\bsnm{Darrell}, \binits{T.}}:
Region Similarity Representation Learning
(2021)
\end{botherref}
\endbibitem

\bibitem[\protect\citeauthoryear{Yang et~al.}{2021}]{yang2021insloc}
\begin{botherref}
\oauthor{\bsnm{Yang}, \binits{C.}},
\oauthor{\bsnm{Wu}, \binits{Z.}},
\oauthor{\bsnm{Zhou}, \binits{B.}},
\oauthor{\bsnm{Lin}, \binits{S.}}:
Instance Localization for Self-supervised Detection Pretraining
(2021)
\end{botherref}
\endbibitem

\bibitem[\protect\citeauthoryear{Girshick}{2015}]{girshick2015fastrcnn}
\begin{botherref}
\oauthor{\bsnm{Girshick}, \binits{R.}}:
Fast R-CNN
(2015)
\end{botherref}
\endbibitem

\bibitem[\protect\citeauthoryear{Buettner and Kovashka}{2022}]{buettner2022insloc}
\begin{botherref}
\oauthor{\bsnm{Buettner}, \binits{K.}},
\oauthor{\bsnm{Kovashka}, \binits{A.}}:
Contrastive View Design Strategies to Enhance Robustness to Domain Shifts in Downstream Object Detection
(2022)
\end{botherref}
\endbibitem

\bibitem[\protect\citeauthoryear{Bardes et~al.}{2022}]{bardes2022vicregl}
\begin{botherref}
\oauthor{\bsnm{Bardes}, \binits{A.}},
\oauthor{\bsnm{Ponce}, \binits{J.}},
\oauthor{\bsnm{LeCun}, \binits{Y.}}:
VICRegL: Self-Supervised Learning of Local Visual Features
(2022)
\end{botherref}
\endbibitem

\bibitem[\protect\citeauthoryear{Wang et~al.}{2022}]{wang2022cp2}
\begin{botherref}
\oauthor{\bsnm{Wang}, \binits{F.}},
\oauthor{\bsnm{Wang}, \binits{H.}},
\oauthor{\bsnm{Wei}, \binits{C.}},
\oauthor{\bsnm{Yuille}, \binits{A.}},
\oauthor{\bsnm{Shen}, \binits{W.}}:
CP2: Copy-Paste Contrastive Pretraining for Semantic Segmentation
(2022)
\end{botherref}
\endbibitem

\bibitem[\protect\citeauthoryear{Wei et~al.}{2021}]{wei2021soco}
\begin{botherref}
\oauthor{\bsnm{Wei}, \binits{F.}},
\oauthor{\bsnm{Gao}, \binits{Y.}},
\oauthor{\bsnm{Wu}, \binits{Z.}},
\oauthor{\bsnm{Hu}, \binits{H.}},
\oauthor{\bsnm{Lin}, \binits{S.}}:
Aligning Pretraining for Detection via Object-Level Contrastive Learning
(2021).
\url{https://arxiv.org/abs/2106.02637}
\end{botherref}
\endbibitem

\bibitem[\protect\citeauthoryear{Xie et~al.}{2021}]{xie2021detco}
\begin{botherref}
\oauthor{\bsnm{Xie}, \binits{E.}},
\oauthor{\bsnm{Ding}, \binits{J.}},
\oauthor{\bsnm{Wang}, \binits{W.}},
\oauthor{\bsnm{Zhan}, \binits{X.}},
\oauthor{\bsnm{Xu}, \binits{H.}},
\oauthor{\bsnm{Sun}, \binits{P.}},
\oauthor{\bsnm{Li}, \binits{Z.}},
\oauthor{\bsnm{Luo}, \binits{P.}}:
DetCo: Unsupervised Contrastive Learning for Object Detection
(2021).
\url{https://arxiv.org/abs/2102.04803}
\end{botherref}
\endbibitem

\bibitem[\protect\citeauthoryear{Yang et~al.}{2022}]{yang2022inscon}
\begin{botherref}
\oauthor{\bsnm{Yang}, \binits{J.}},
\oauthor{\bsnm{Zhang}, \binits{K.}},
\oauthor{\bsnm{Cui}, \binits{Z.}},
\oauthor{\bsnm{Su}, \binits{J.}},
\oauthor{\bsnm{Luo}, \binits{J.}},
\oauthor{\bsnm{Wei}, \binits{X.}}:
InsCon:Instance Consistency Feature Representation via Self-Supervised Learning
(2022).
\url{https://arxiv.org/abs/2203.07688}
\end{botherref}
\endbibitem

\bibitem[\protect\citeauthoryear{Li et~al.}{2022}]{li2022univip}
\begin{botherref}
\oauthor{\bsnm{Li}, \binits{Z.}},
\oauthor{\bsnm{Zhu}, \binits{Y.}},
\oauthor{\bsnm{Yang}, \binits{F.}},
\oauthor{\bsnm{Li}, \binits{W.}},
\oauthor{\bsnm{Zhao}, \binits{C.}},
\oauthor{\bsnm{Chen}, \binits{Y.}},
\oauthor{\bsnm{Chen}, \binits{Z.}},
\oauthor{\bsnm{Xie}, \binits{J.}},
\oauthor{\bsnm{Wu}, \binits{L.}},
\oauthor{\bsnm{Zhao}, \binits{R.}},
\oauthor{\bsnm{Tang}, \binits{M.}},
\oauthor{\bsnm{Wang}, \binits{J.}}:
UniVIP: A Unified Framework for Self-Supervised Visual Pre-training
(2022)
\end{botherref}
\endbibitem

\bibitem[\protect\citeauthoryear{Lin et~al.}{2015}]{li2015coco}
\begin{botherref}
\oauthor{\bsnm{Lin}, \binits{T.-Y.}},
\oauthor{\bsnm{Maire}, \binits{M.}},
\oauthor{\bsnm{Belongie}, \binits{S.}},
\oauthor{\bsnm{Bourdev}, \binits{L.}},
\oauthor{\bsnm{Girshick}, \binits{R.}},
\oauthor{\bsnm{Hays}, \binits{J.}},
\oauthor{\bsnm{Perona}, \binits{P.}},
\oauthor{\bsnm{Ramanan}, \binits{D.}},
\oauthor{\bsnm{Zitnick}, \binits{C.L.}},
\oauthor{\bsnm{Dollár}, \binits{P.}}:
Microsoft COCO: Common Objects in Context
(2015)
\end{botherref}
\endbibitem

\bibitem[\protect\citeauthoryear{Zhuang et~al.}{2019}]{zhuang2019la}
\begin{botherref}
\oauthor{\bsnm{Zhuang}, \binits{C.}},
\oauthor{\bsnm{Zhai}, \binits{A.L.}},
\oauthor{\bsnm{Yamins}, \binits{D.}}:
Local Aggregation for Unsupervised Learning of Visual Embeddings
(2019)
\end{botherref}
\endbibitem

\bibitem[\protect\citeauthoryear{Zhan et~al.}{2020}]{zhan2020odc}
\begin{botherref}
\oauthor{\bsnm{Zhan}, \binits{X.}},
\oauthor{\bsnm{Xie}, \binits{J.}},
\oauthor{\bsnm{Liu}, \binits{Z.}},
\oauthor{\bsnm{Ong}, \binits{Y.S.}},
\oauthor{\bsnm{Loy}, \binits{C.C.}}:
Online Deep Clustering for Unsupervised Representation Learning
(2020)
\end{botherref}
\endbibitem

\bibitem[\protect\citeauthoryear{YM. et~al.}{2020}]{asano2020sela}
\begin{bchapter}
\bauthor{\bsnm{YM.}, \binits{A.}},
\bauthor{\bsnm{C.}, \binits{R.}},
\bauthor{\bsnm{A.}, \binits{V.}}:
\bctitle{Self-labelling via simultaneous clustering and representation learning}.
In: \bbtitle{International Conference on Learning Representations}
(\byear{2020}).
\burl{https://openreview.net/forum?id=Hyx-jyBFPr}
\end{bchapter}
\endbibitem

\bibitem[\protect\citeauthoryear{Cuturi}{2013}]{cuturi2013sinkhorn}
\begin{bchapter}
\bauthor{\bsnm{Cuturi}, \binits{M.}}:
\bctitle{Sinkhorn distances: Lightspeed computation of optimal transport}.
In: \beditor{\bsnm{Burges}, \binits{C.J.}},
\beditor{\bsnm{Bottou}, \binits{L.}},
\beditor{\bsnm{Welling}, \binits{M.}},
\beditor{\bsnm{Ghahramani}, \binits{Z.}},
\beditor{\bsnm{Weinberger}, \binits{K.Q.}} (eds.)
\bbtitle{Advances in Neural Information Processing Systems},
vol. \bseriesno{26}.
\bpublisher{Curran Associates, Inc.}, \blocation{???}
(\byear{2013}).
\burl{https://proceedings.neurips.cc/paper_files/paper/2013/file/af21d0c97db2e27e13572cbf59eb343d-Paper.pdf}
\end{bchapter}
\endbibitem

\bibitem[\protect\citeauthoryear{Qian et~al.}{2022}]{qian2022coke}
\begin{botherref}
\oauthor{\bsnm{Qian}, \binits{Q.}},
\oauthor{\bsnm{Xu}, \binits{Y.}},
\oauthor{\bsnm{Hu}, \binits{J.}},
\oauthor{\bsnm{Li}, \binits{H.}},
\oauthor{\bsnm{Jin}, \binits{R.}}:
Unsupervised Visual Representation Learning by Online Constrained K-Means
(2022)
\end{botherref}
\endbibitem

\bibitem[\protect\citeauthoryear{Radosavovic et~al.}{2020}]{radosavovic2020regnet}
\begin{botherref}
\oauthor{\bsnm{Radosavovic}, \binits{I.}},
\oauthor{\bsnm{Kosaraju}, \binits{R.P.}},
\oauthor{\bsnm{Girshick}, \binits{R.}},
\oauthor{\bsnm{He}, \binits{K.}},
\oauthor{\bsnm{Dollár}, \binits{P.}}:
Designing Network Design Spaces
(2020)
\end{botherref}
\endbibitem

\bibitem[\protect\citeauthoryear{Gansbeke et~al.}{2020}]{vangansbeke2020scan}
\begin{botherref}
\oauthor{\bsnm{Gansbeke}, \binits{W.V.}},
\oauthor{\bsnm{Vandenhende}, \binits{S.}},
\oauthor{\bsnm{Georgoulis}, \binits{S.}},
\oauthor{\bsnm{Proesmans}, \binits{M.}},
\oauthor{\bsnm{Gool}, \binits{L.V.}}:
SCAN: Learning to Classify Images without Labels
(2020)
\end{botherref}
\endbibitem

\bibitem[\protect\citeauthoryear{Pang et~al.}{2022}]{pang2022smog}
\begin{botherref}
\oauthor{\bsnm{Pang}, \binits{B.}},
\oauthor{\bsnm{Zhang}, \binits{Y.}},
\oauthor{\bsnm{Li}, \binits{Y.}},
\oauthor{\bsnm{Cai}, \binits{J.}},
\oauthor{\bsnm{Lu}, \binits{C.}}:
Unsupervised Visual Representation Learning by Synchronous Momentum Grouping
(2022)
\end{botherref}
\endbibitem

\bibitem[\protect\citeauthoryear{Amrani et~al.}{2022}]{amrani2022selfclassifier}
\begin{botherref}
\oauthor{\bsnm{Amrani}, \binits{E.}},
\oauthor{\bsnm{Karlinsky}, \binits{L.}},
\oauthor{\bsnm{Bronstein}, \binits{A.}}:
Self-Supervised Classification Network
(2022)
\end{botherref}
\endbibitem

\bibitem[\protect\citeauthoryear{Caron et~al.}{2021}]{caron2021dino}
\begin{botherref}
\oauthor{\bsnm{Caron}, \binits{M.}},
\oauthor{\bsnm{Touvron}, \binits{H.}},
\oauthor{\bsnm{Misra}, \binits{I.}},
\oauthor{\bsnm{Jégou}, \binits{H.}},
\oauthor{\bsnm{Mairal}, \binits{J.}},
\oauthor{\bsnm{Bojanowski}, \binits{P.}},
\oauthor{\bsnm{Joulin}, \binits{A.}}:
Emerging Properties in Self-Supervised Vision Transformers
(2021)
\end{botherref}
\endbibitem

\bibitem[\protect\citeauthoryear{Oquab et~al.}{2023}]{oquab2023dinov2}
\begin{botherref}
\oauthor{\bsnm{Oquab}, \binits{M.}},
\oauthor{\bsnm{Darcet}, \binits{T.}},
\oauthor{\bsnm{Moutakanni}, \binits{T.}},
\oauthor{\bsnm{Vo}, \binits{H.V.}},
\oauthor{\bsnm{Szafraniec}, \binits{M.}},
\oauthor{\bsnm{Khalidov}, \binits{V.}},
\oauthor{\bsnm{Fernandez}, \binits{P.}},
\oauthor{\bsnm{Haziza}, \binits{D.}},
\oauthor{\bsnm{Massa}, \binits{F.}},
\oauthor{\bsnm{El-Nouby}, \binits{A.}},
\oauthor{\bsnm{Howes}, \binits{R.}},
\oauthor{\bsnm{Huang}, \binits{P.-Y.}},
\oauthor{\bsnm{Xu}, \binits{H.}},
\oauthor{\bsnm{Sharma}, \binits{V.}},
\oauthor{\bsnm{Li}, \binits{S.-W.}},
\oauthor{\bsnm{Galuba}, \binits{W.}},
\oauthor{\bsnm{Rabbat}, \binits{M.}},
\oauthor{\bsnm{Assran}, \binits{M.}},
\oauthor{\bsnm{Ballas}, \binits{N.}},
\oauthor{\bsnm{Synnaeve}, \binits{G.}},
\oauthor{\bsnm{Misra}, \binits{I.}},
\oauthor{\bsnm{Jegou}, \binits{H.}},
\oauthor{\bsnm{Mairal}, \binits{J.}},
\oauthor{\bsnm{Labatut}, \binits{P.}},
\oauthor{\bsnm{Joulin}, \binits{A.}},
\oauthor{\bsnm{Bojanowski}, \binits{P.}}:
DINOv2: Learning Robust Visual Features without Supervision
(2023)
\end{botherref}
\endbibitem

\bibitem[\protect\citeauthoryear{Zhou et~al.}{2022}]{zhou2022ibot}
\begin{botherref}
\oauthor{\bsnm{Zhou}, \binits{J.}},
\oauthor{\bsnm{Wei}, \binits{C.}},
\oauthor{\bsnm{Wang}, \binits{H.}},
\oauthor{\bsnm{Shen}, \binits{W.}},
\oauthor{\bsnm{Xie}, \binits{C.}},
\oauthor{\bsnm{Yuille}, \binits{A.}},
\oauthor{\bsnm{Kong}, \binits{T.}}:
iBOT: Image BERT Pre-Training with Online Tokenizer
(2022)
\end{botherref}
\endbibitem

\bibitem[\protect\citeauthoryear{Sablayrolles et~al.}{2019}]{sablayrolles2019koleo}
\begin{botherref}
\oauthor{\bsnm{Sablayrolles}, \binits{A.}},
\oauthor{\bsnm{Douze}, \binits{M.}},
\oauthor{\bsnm{Schmid}, \binits{C.}},
\oauthor{\bsnm{Jégou}, \binits{H.}}:
Spreading vectors for similarity search
(2019)
\end{botherref}
\endbibitem

\bibitem[\protect\citeauthoryear{Pototzky et~al.}{2022}]{pototzky2022fastsiam}
\begin{bbook}
\bauthor{\bsnm{Pototzky}, \binits{D.}},
\bauthor{\bsnm{Sultan}, \binits{A.}},
\bauthor{\bsnm{Schmidt-Thieme}, \binits{L.}}:
\bbtitle{FastSiam: Resource-Efficient Self-supervised Learning on a Single GPU},
pp. \bfpage{53}--\blpage{67}.
\bpublisher{Springer}, \blocation{???}
(\byear{2022}).
\doiurl{10.1007/978-3-031-16788-1_4} .
\burl{http://dx.doi.org/10.1007/978-3-031-16788-1_4}
\end{bbook}
\endbibitem

\bibitem[\protect\citeauthoryear{Gidaris et~al.}{2021}]{gidaris2021obow}
\begin{bchapter}
\bauthor{\bsnm{Gidaris}, \binits{S.}},
\bauthor{\bsnm{Bursuc}, \binits{A.}},
\bauthor{\bsnm{Puy}, \binits{G.}},
\bauthor{\bsnm{Komodakis}, \binits{N.}},
\bauthor{\bsnm{Cord}, \binits{M.}},
\bauthor{\bsnm{P{\'e}rez}, \binits{P.}}:
\bctitle{Learning representations by predicting bags of visual words}.
In: \bbtitle{CVPR}
(\byear{2021})
\end{bchapter}
\endbibitem

\bibitem[\protect\citeauthoryear{Hinton et~al.}{2015}]{hinton2015distilling}
\begin{botherref}
\oauthor{\bsnm{Hinton}, \binits{G.}},
\oauthor{\bsnm{Vinyals}, \binits{O.}},
\oauthor{\bsnm{Dean}, \binits{J.}}:
Distilling the Knowledge in a Neural Network
(2015)
\end{botherref}
\endbibitem

\bibitem[\protect\citeauthoryear{Tian et~al.}{2022}]{tian2022crd}
\begin{botherref}
\oauthor{\bsnm{Tian}, \binits{Y.}},
\oauthor{\bsnm{Krishnan}, \binits{D.}},
\oauthor{\bsnm{Isola}, \binits{P.}}:
Contrastive Representation Distillation
(2022)
\end{botherref}
\endbibitem

\bibitem[\protect\citeauthoryear{Giakoumoglou and Stathaki}{2024a}]{giakoumoglou2024rrd}
\begin{botherref}
\oauthor{\bsnm{Giakoumoglou}, \binits{N.}},
\oauthor{\bsnm{Stathaki}, \binits{T.}}:
Relational Representation Distillation
(2024).
\url{https://arxiv.org/abs/2407.12073}
\end{botherref}
\endbibitem

\bibitem[\protect\citeauthoryear{Giakoumoglou and Stathaki}{2024b}]{giakoumoglou2024dcd}
\begin{botherref}
\oauthor{\bsnm{Giakoumoglou}, \binits{N.}},
\oauthor{\bsnm{Stathaki}, \binits{T.}}:
Discriminative and Consistent Representation Distillation
(2024).
\url{https://arxiv.org/abs/2407.11802}
\end{botherref}
\endbibitem

\bibitem[\protect\citeauthoryear{Song et~al.}{2023}]{song2023mmkd}
\begin{botherref}
\oauthor{\bsnm{Song}, \binits{K.}},
\oauthor{\bsnm{Xie}, \binits{J.}},
\oauthor{\bsnm{Zhang}, \binits{S.}},
\oauthor{\bsnm{Luo}, \binits{Z.}}:
Multi-Mode Online Knowledge Distillation for Self-Supervised Visual Representation Learning
(2023).
\url{https://arxiv.org/abs/2304.06461}
\end{botherref}
\endbibitem

\bibitem[\protect\citeauthoryear{Xu et~al.}{2020}]{xu2020sskd}
\begin{botherref}
\oauthor{\bsnm{Xu}, \binits{G.}},
\oauthor{\bsnm{Liu}, \binits{Z.}},
\oauthor{\bsnm{Li}, \binits{X.}},
\oauthor{\bsnm{Loy}, \binits{C.C.}}:
Knowledge Distillation Meets Self-Supervision
(2020).
\url{https://arxiv.org/abs/2006.07114}
\end{botherref}
\endbibitem

\bibitem[\protect\citeauthoryear{Dosovitskiy et~al.}{2015}]{dosovitskiy2015exemplar}
\begin{botherref}
\oauthor{\bsnm{Dosovitskiy}, \binits{A.}},
\oauthor{\bsnm{Fischer}, \binits{P.}},
\oauthor{\bsnm{Springenberg}, \binits{J.T.}},
\oauthor{\bsnm{Riedmiller}, \binits{M.}},
\oauthor{\bsnm{Brox}, \binits{T.}}:
Discriminative Unsupervised Feature Learning with Exemplar Convolutional Neural Networks
(2015).
\url{https://arxiv.org/abs/1406.6909}
\end{botherref}
\endbibitem

\bibitem[\protect\citeauthoryear{Bhat et~al.}{2021}]{bhat2021dogo}
\begin{botherref}
\oauthor{\bsnm{Bhat}, \binits{P.}},
\oauthor{\bsnm{Arani}, \binits{E.}},
\oauthor{\bsnm{Zonooz}, \binits{B.}}:
Distill on the Go: Online knowledge distillation in self-supervised learning
(2021).
\url{https://arxiv.org/abs/2104.09866}
\end{botherref}
\endbibitem

\bibitem[\protect\citeauthoryear{Fang et~al.}{2021}]{fang2021seed}
\begin{botherref}
\oauthor{\bsnm{Fang}, \binits{Z.}},
\oauthor{\bsnm{Wang}, \binits{J.}},
\oauthor{\bsnm{Wang}, \binits{L.}},
\oauthor{\bsnm{Zhang}, \binits{L.}},
\oauthor{\bsnm{Yang}, \binits{Y.}},
\oauthor{\bsnm{Liu}, \binits{Z.}}:
SEED: Self-supervised Distillation For Visual Representation
(2021)
\end{botherref}
\endbibitem

\bibitem[\protect\citeauthoryear{Gao et~al.}{2022}]{gao2022disco}
\begin{botherref}
\oauthor{\bsnm{Gao}, \binits{Y.}},
\oauthor{\bsnm{Zhuang}, \binits{J.-X.}},
\oauthor{\bsnm{Lin}, \binits{S.}},
\oauthor{\bsnm{Cheng}, \binits{H.}},
\oauthor{\bsnm{Sun}, \binits{X.}},
\oauthor{\bsnm{Li}, \binits{K.}},
\oauthor{\bsnm{Shen}, \binits{C.}}:
DisCo: Remedy Self-supervised Learning on Lightweight Models with Distilled Contrastive Learning
(2022)
\end{botherref}
\endbibitem

\bibitem[\protect\citeauthoryear{Xu et~al.}{2022}]{xu2022bingo}
\begin{botherref}
\oauthor{\bsnm{Xu}, \binits{H.}},
\oauthor{\bsnm{Fang}, \binits{J.}},
\oauthor{\bsnm{Zhang}, \binits{X.}},
\oauthor{\bsnm{Xie}, \binits{L.}},
\oauthor{\bsnm{Wang}, \binits{X.}},
\oauthor{\bsnm{Dai}, \binits{W.}},
\oauthor{\bsnm{Xiong}, \binits{H.}},
\oauthor{\bsnm{Tian}, \binits{Q.}}:
Bag of Instances Aggregation Boosts Self-supervised Distillation
(2022)
\end{botherref}
\endbibitem

\bibitem[\protect\citeauthoryear{Hotelling}{1992}]{hotelling1992relations}
\begin{bbook}
\bauthor{\bsnm{Hotelling}, \binits{H.}}:
In: \beditor{\bsnm{Kotz}, \binits{S.}},
\beditor{\bsnm{Johnson}, \binits{N.L.}} (eds.)
\bbtitle{Relations Between Two Sets of Variates},
pp. \bfpage{162}--\blpage{190}.
\bpublisher{Springer},
\blocation{New York, NY}
(\byear{1992}).
\doiurl{10.1007/978-1-4612-4380-9_14} .
\burl{https://doi.org/10.1007/978-1-4612-4380-9_14}
\end{bbook}
\endbibitem

\bibitem[\protect\citeauthoryear{Bardes et~al.}{2022}]{bardes2022vicreg}
\begin{botherref}
\oauthor{\bsnm{Bardes}, \binits{A.}},
\oauthor{\bsnm{Ponce}, \binits{J.}},
\oauthor{\bsnm{LeCun}, \binits{Y.}}:
VICReg: Variance-Invariance-Covariance Regularization for Self-Supervised Learning
(2022)
\end{botherref}
\endbibitem

\bibitem[\protect\citeauthoryear{Ermolov et~al.}{2021}]{ermolov2021wmse}
\begin{botherref}
\oauthor{\bsnm{Ermolov}, \binits{A.}},
\oauthor{\bsnm{Siarohin}, \binits{A.}},
\oauthor{\bsnm{Sangineto}, \binits{E.}},
\oauthor{\bsnm{Sebe}, \binits{N.}}:
Whitening for Self-Supervised Representation Learning
(2021)
\end{botherref}
\endbibitem

\bibitem[\protect\citeauthoryear{Bandara et~al.}{2023}]{bandara2023mixedbarlowtwins}
\begin{botherref}
\oauthor{\bsnm{Bandara}, \binits{W.G.C.}},
\oauthor{\bsnm{Melo}, \binits{C.M.D.}},
\oauthor{\bsnm{Patel}, \binits{V.M.}}:
Guarding Barlow Twins Against Overfitting with Mixed Samples
(2023)
\end{botherref}
\endbibitem

\bibitem[\protect\citeauthoryear{Zhang et~al.}{2022}]{zhang2022arb}
\begin{bchapter}
\bauthor{\bsnm{Zhang}, \binits{S.}},
\bauthor{\bsnm{Qiu}, \binits{L.}},
\bauthor{\bsnm{Zhu}, \binits{F.}},
\bauthor{\bsnm{Yan}, \binits{J.}},
\bauthor{\bsnm{Zhang}, \binits{H.}},
\bauthor{\bsnm{Zhao}, \binits{R.}},
\bauthor{\bsnm{Li}, \binits{H.}},
\bauthor{\bsnm{Yang}, \binits{X.}}:
\bctitle{Align representations with base: A new approach to self-supervised learning}.
In: \bbtitle{The IEEE / CVF Computer Vision and Pattern Recognition Conference}
(\byear{2022})
\end{bchapter}
\endbibitem

\bibitem[\protect\citeauthoryear{Strang}{2006}]{strang2006linear}
\begin{bbook}
\bauthor{\bsnm{Strang}, \binits{G.}}:
\bbtitle{Linear Algebra and Its Applications}.
\bpublisher{Thomson, Brooks/Cole},
\blocation{Belmont, CA}
(\byear{2006}).
\burl{http://www.amazon.com/Linear-Algebra-Its-Applications-Edition/dp/0030105676}
\end{bbook}
\endbibitem

\bibitem[\protect\citeauthoryear{Kalantidis et~al.}{2022}]{kalantidis2022tldr}
\begin{botherref}
\oauthor{\bsnm{Kalantidis}, \binits{Y.}},
\oauthor{\bsnm{Lassance}, \binits{C.}},
\oauthor{\bsnm{Almazan}, \binits{J.}},
\oauthor{\bsnm{Larlus}, \binits{D.}}:
TLDR: Twin Learning for Dimensionality Reduction
(2022)
\end{botherref}
\endbibitem

\bibitem[\protect\citeauthoryear{Wang et~al.}{2021a}]{wang2021twist}
\begin{botherref}
\oauthor{\bsnm{Wang}, \binits{F.}},
\oauthor{\bsnm{Kong}, \binits{T.}},
\oauthor{\bsnm{Zhang}, \binits{R.}},
\oauthor{\bsnm{Liu}, \binits{H.}},
\oauthor{\bsnm{Li}, \binits{H.}}:
Self-Supervised Learning by Estimating Twin Class Distributions
(2021)
\end{botherref}
\endbibitem

\bibitem[\protect\citeauthoryear{Wang et~al.}{2021b}]{wang2021truncatedtriplet}
\begin{botherref}
\oauthor{\bsnm{Wang}, \binits{G.}},
\oauthor{\bsnm{Wang}, \binits{K.}},
\oauthor{\bsnm{Wang}, \binits{G.}},
\oauthor{\bsnm{Torr}, \binits{P.H.S.}},
\oauthor{\bsnm{Lin}, \binits{L.}}:
Solving Inefficiency of Self-supervised Representation Learning
(2021)
\end{botherref}
\endbibitem

\bibitem[\protect\citeauthoryear{He et~al.}{2024}]{he2024or}
\begin{botherref}
\oauthor{\bsnm{He}, \binits{J.}},
\oauthor{\bsnm{Du}, \binits{J.}},
\oauthor{\bsnm{Ma}, \binits{W.}}:
Preventing Dimensional Collapse in Self-Supervised Learning via Orthogonality Regularization
(2024).
\url{https://arxiv.org/abs/2411.00392}
\end{botherref}
\endbibitem

\bibitem[\protect\citeauthoryear{Kolesnikov et~al.}{2019}]{kolesnikov2019revisiting}
\begin{botherref}
\oauthor{\bsnm{Kolesnikov}, \binits{A.}},
\oauthor{\bsnm{Zhai}, \binits{X.}},
\oauthor{\bsnm{Beyer}, \binits{L.}}:
Revisiting Self-Supervised Visual Representation Learning
(2019)
\end{botherref}
\endbibitem

\bibitem[\protect\citeauthoryear{Ruder}{2017}]{ruder2017sgd}
\begin{botherref}
\oauthor{\bsnm{Ruder}, \binits{S.}}:
An overview of gradient descent optimization algorithms
(2017)
\end{botherref}
\endbibitem

\bibitem[\protect\citeauthoryear{Kornblith et~al.}{2019}]{kornblith2019do}
\begin{botherref}
\oauthor{\bsnm{Kornblith}, \binits{S.}},
\oauthor{\bsnm{Shlens}, \binits{J.}},
\oauthor{\bsnm{Le}, \binits{Q.V.}}:
Do Better ImageNet Models Transfer Better?
(2019)
\end{botherref}
\endbibitem

\bibitem[\protect\citeauthoryear{Zhai et~al.}{2020}]{zhai2020large}
\begin{botherref}
\oauthor{\bsnm{Zhai}, \binits{X.}},
\oauthor{\bsnm{Puigcerver}, \binits{J.}},
\oauthor{\bsnm{Kolesnikov}, \binits{A.}},
\oauthor{\bsnm{Ruyssen}, \binits{P.}},
\oauthor{\bsnm{Riquelme}, \binits{C.}},
\oauthor{\bsnm{Lucic}, \binits{M.}},
\oauthor{\bsnm{Djolonga}, \binits{J.}},
\oauthor{\bsnm{Pinto}, \binits{A.S.}},
\oauthor{\bsnm{Neumann}, \binits{M.}},
\oauthor{\bsnm{Dosovitskiy}, \binits{A.}},
\oauthor{\bsnm{Beyer}, \binits{L.}},
\oauthor{\bsnm{Bachem}, \binits{O.}},
\oauthor{\bsnm{Tschannen}, \binits{M.}},
\oauthor{\bsnm{Michalski}, \binits{M.}},
\oauthor{\bsnm{Bousquet}, \binits{O.}},
\oauthor{\bsnm{Gelly}, \binits{S.}},
\oauthor{\bsnm{Houlsby}, \binits{N.}}:
A Large-scale Study of Representation Learning with the Visual Task Adaptation Benchmark
(2020)
\end{botherref}
\endbibitem

\bibitem[\protect\citeauthoryear{Krizhevsky}{2009}]{krizhevsky2009cifar}
\begin{botherref}
\oauthor{\bsnm{Krizhevsky}, \binits{A.}}:
Learning multiple layers of features from tiny images,
32--33
(2009)
\end{botherref}
\endbibitem

\bibitem[\protect\citeauthoryear{Zhou et~al.}{2018}]{zhou2018places365}
\begin{barticle}
\bauthor{\bsnm{Zhou}, \binits{B.}},
\bauthor{\bsnm{Lapedriza}, \binits{A.}},
\bauthor{\bsnm{Khosla}, \binits{A.}},
\bauthor{\bsnm{Oliva}, \binits{A.}},
\bauthor{\bsnm{Torralba}, \binits{A.}}:
\batitle{Places: A 10 million image database for scene recognition}.
\bjtitle{IEEE Transactions on Pattern Analysis and Machine Intelligence}
\bvolume{40}(\bissue{6}),
\bfpage{1452}--\blpage{1464}
(\byear{2018})
\doiurl{10.1109/TPAMI.2017.2723009}
\end{barticle}
\endbibitem

\bibitem[\protect\citeauthoryear{Yalniz et~al.}{2019}]{yalniz2019instagram}
\begin{botherref}
\oauthor{\bsnm{Yalniz}, \binits{I.Z.}},
\oauthor{\bsnm{Jégou}, \binits{H.}},
\oauthor{\bsnm{Chen}, \binits{K.}},
\oauthor{\bsnm{Paluri}, \binits{M.}},
\oauthor{\bsnm{Mahajan}, \binits{D.}}:
Billion-scale semi-supervised learning for image classification
(2019)
\end{botherref}
\endbibitem

\bibitem[\protect\citeauthoryear{}{}]{iNaturalist}
\begin{botherref}
iNaturalist.
\url{https://www.inaturalist.org}.
Accessed: 20 November 2024
\end{botherref}
\endbibitem

\bibitem[\protect\citeauthoryear{Wang and Qi}{2022}]{wang2022clsa}
\begin{botherref}
\oauthor{\bsnm{Wang}, \binits{X.}},
\oauthor{\bsnm{Qi}, \binits{G.-J.}}:
Contrastive Learning with Stronger Augmentations
(2022)
\end{botherref}
\endbibitem

\bibitem[\protect\citeauthoryear{Wang et~al.}{2022}]{wang2022usb}
\begin{botherref}
\oauthor{\bsnm{Wang}, \binits{Y.}},
\oauthor{\bsnm{Chen}, \binits{H.}},
\oauthor{\bsnm{Fan}, \binits{Y.}},
\oauthor{\bsnm{Sun}, \binits{W.}},
\oauthor{\bsnm{Tao}, \binits{R.}},
\oauthor{\bsnm{Hou}, \binits{W.}},
\oauthor{\bsnm{Wang}, \binits{R.}},
\oauthor{\bsnm{Yang}, \binits{L.}},
\oauthor{\bsnm{Zhou}, \binits{Z.}},
\oauthor{\bsnm{Guo}, \binits{L.-Z.}},
\oauthor{\bsnm{Qi}, \binits{H.}},
\oauthor{\bsnm{Wu}, \binits{Z.}},
\oauthor{\bsnm{Li}, \binits{Y.-F.}},
\oauthor{\bsnm{Nakamura}, \binits{S.}},
\oauthor{\bsnm{Ye}, \binits{W.}},
\oauthor{\bsnm{Savvides}, \binits{M.}},
\oauthor{\bsnm{Raj}, \binits{B.}},
\oauthor{\bsnm{Shinozaki}, \binits{T.}},
\oauthor{\bsnm{Schiele}, \binits{B.}},
\oauthor{\bsnm{Wang}, \binits{J.}},
\oauthor{\bsnm{Xie}, \binits{X.}},
\oauthor{\bsnm{Zhang}, \binits{Y.}}:
USB: A Unified Semi-supervised Learning Benchmark for Classification
(2022).
\url{https://arxiv.org/abs/2208.07204}
\end{botherref}
\endbibitem

\bibitem[\protect\citeauthoryear{Bossard et~al.}{2014}]{bossard2014food101}
\begin{bchapter}
\bauthor{\bsnm{Bossard}, \binits{L.}},
\bauthor{\bsnm{Guillaumin}, \binits{M.}},
\bauthor{\bsnm{Gool}, \binits{L.V.}}:
\bctitle{Food-101 {\textendash} mining discriminative components with random forests}.
In: \bbtitle{Computer Vision {\textendash} {ECCV} 2014},
pp. \bfpage{446}--\blpage{461}.
\bpublisher{Springer}, \blocation{???}
(\byear{2014}).
\doiurl{10.1007/978-3-319-10599-4_29} .
\burl{https://doi.org/10.1007/978-3-319-10599-4_29}
\end{bchapter}
\endbibitem

\bibitem[\protect\citeauthoryear{Berg et~al.}{2014}]{berg2014birdsnap}
\begin{bchapter}
\bauthor{\bsnm{Berg}, \binits{T.}},
\bauthor{\bsnm{Liu}, \binits{J.}},
\bauthor{\bsnm{Lee}, \binits{S.W.}},
\bauthor{\bsnm{Alexander}, \binits{M.L.}},
\bauthor{\bsnm{Jacobs}, \binits{D.W.}},
\bauthor{\bsnm{Belhumeur}, \binits{P.N.}}:
\bctitle{Birdsnap: Large-scale fine-grained visual categorization of birds}.
In: \bbtitle{2014 IEEE Conference on Computer Vision and Pattern Recognition},
pp. \bfpage{2019}--\blpage{2026}
(\byear{2014}).
\doiurl{10.1109/CVPR.2014.259}
\end{bchapter}
\endbibitem

\bibitem[\protect\citeauthoryear{Xiao et~al.}{2010}]{xiao2010sun397}
\begin{bchapter}
\bauthor{\bsnm{Xiao}, \binits{J.}},
\bauthor{\bsnm{Hays}, \binits{J.}},
\bauthor{\bsnm{Ehinger}, \binits{K.A.}},
\bauthor{\bsnm{Oliva}, \binits{A.}},
\bauthor{\bsnm{Torralba}, \binits{A.}}:
\bctitle{Sun database: Large-scale scene recognition from abbey to zoo}.
In: \bbtitle{2010 IEEE Computer Society Conference on Computer Vision and Pattern Recognition},
pp. \bfpage{3485}--\blpage{3492}
(\byear{2010}).
\doiurl{10.1109/CVPR.2010.5539970}
\end{bchapter}
\endbibitem

\bibitem[\protect\citeauthoryear{Krause et~al.}{2013}]{krause2013cars}
\begin{bchapter}
\bauthor{\bsnm{Krause}, \binits{J.}},
\bauthor{\bsnm{Stark}, \binits{M.}},
\bauthor{\bsnm{Deng}, \binits{J.}},
\bauthor{\bsnm{Fei-Fei}, \binits{L.}}:
\bctitle{3d object representations for fine-grained categorization}.
In: \bbtitle{2013 IEEE International Conference on Computer Vision Workshops},
pp. \bfpage{554}--\blpage{561}
(\byear{2013}).
\doiurl{10.1109/ICCVW.2013.77}
\end{bchapter}
\endbibitem

\bibitem[\protect\citeauthoryear{Maji et~al.}{2013}]{maji2013aircraft}
\begin{botherref}
\oauthor{\bsnm{Maji}, \binits{S.}},
\oauthor{\bsnm{Rahtu}, \binits{E.}},
\oauthor{\bsnm{Kannala}, \binits{J.}},
\oauthor{\bsnm{Blaschko}, \binits{M.}},
\oauthor{\bsnm{Vedaldi}, \binits{A.}}:
Fine-Grained Visual Classification of Aircraft
(2013)
\end{botherref}
\endbibitem

\bibitem[\protect\citeauthoryear{Everingham et~al.}{2009}]{everingham2009voc}
\begin{barticle}
\bauthor{\bsnm{Everingham}, \binits{M.}},
\bauthor{\bsnm{Gool}, \binits{L.V.}},
\bauthor{\bsnm{Williams}, \binits{C.K.I.}},
\bauthor{\bsnm{Winn}, \binits{J.}},
\bauthor{\bsnm{Zisserman}, \binits{A.}}:
\batitle{The pascal visual object classes ({VOC}) challenge}.
\bjtitle{International Journal of Computer Vision}
\bvolume{88}(\bissue{2}),
\bfpage{303}--\blpage{338}
(\byear{2009})
\doiurl{10.1007/s11263-009-0275-4}
\end{barticle}
\endbibitem

\bibitem[\protect\citeauthoryear{Cimpoi et~al.}{2014}]{cimpoi2014dtd}
\begin{bchapter}
\bauthor{\bsnm{Cimpoi}, \binits{M.}},
\bauthor{\bsnm{Maji}, \binits{S.}},
\bauthor{\bsnm{Kokkinos}, \binits{I.}},
\bauthor{\bsnm{Mohamed}, \binits{S.}},
\bauthor{\bsnm{Vedaldi}, \binits{A.}}:
\bctitle{Describing textures in the wild}.
In: \bbtitle{2014 IEEE Conference on Computer Vision and Pattern Recognition},
pp. \bfpage{3606}--\blpage{3613}
(\byear{2014}).
\doiurl{10.1109/CVPR.2014.461}
\end{bchapter}
\endbibitem

\bibitem[\protect\citeauthoryear{Parkhi et~al.}{2012}]{parkhi2012pets}
\begin{bchapter}
\bauthor{\bsnm{Parkhi}, \binits{O.M.}},
\bauthor{\bsnm{Vedaldi}, \binits{A.}},
\bauthor{\bsnm{Zisserman}, \binits{A.}},
\bauthor{\bsnm{Jawahar}, \binits{C.V.}}:
\bctitle{Cats and dogs}.
In: \bbtitle{2012 IEEE Conference on Computer Vision and Pattern Recognition},
pp. \bfpage{3498}--\blpage{3505}
(\byear{2012}).
\doiurl{10.1109/CVPR.2012.6248092}
\end{bchapter}
\endbibitem

\bibitem[\protect\citeauthoryear{Fei-Fei et~al.}{2007}]{feifei2007caltech101}
\begin{barticle}
\bauthor{\bsnm{Fei-Fei}, \binits{L.}},
\bauthor{\bsnm{Fergus}, \binits{R.}},
\bauthor{\bsnm{Perona}, \binits{P.}}:
\batitle{Learning generative visual models from few training examples: An incremental bayesian approach tested on 101 object categories}.
\bjtitle{Computer Vision and Image Understanding}
\bvolume{106}(\bissue{1}),
\bfpage{59}--\blpage{70}
(\byear{2007})
\doiurl{10.1016/j.cviu.2005.09.012} .
\bcomment{Special issue on Generative Model Based Vision}
\end{barticle}
\endbibitem

\bibitem[\protect\citeauthoryear{Nilsback and Zisserman}{2008}]{nilsback2008flowers102}
\begin{bchapter}
\bauthor{\bsnm{Nilsback}, \binits{M.-E.}},
\bauthor{\bsnm{Zisserman}, \binits{A.}}:
\bctitle{Automated flower classification over a large number of classes}.
In: \bbtitle{2008 Sixth Indian Conference on Computer Vision, Graphics and Image Processing},
pp. \bfpage{722}--\blpage{729}
(\byear{2008}).
\doiurl{10.1109/ICVGIP.2008.47}
\end{bchapter}
\endbibitem

\bibitem[\protect\citeauthoryear{Chen et~al.}{2017}]{chen2017deeplabv3}
\begin{botherref}
\oauthor{\bsnm{Chen}, \binits{L.-C.}},
\oauthor{\bsnm{Papandreou}, \binits{G.}},
\oauthor{\bsnm{Schroff}, \binits{F.}},
\oauthor{\bsnm{Adam}, \binits{H.}}:
Rethinking Atrous Convolution for Semantic Image Segmentation
(2017)
\end{botherref}
\endbibitem

\bibitem[\protect\citeauthoryear{Wu et~al.}{2019}]{wu2019detectron2}
\begin{botherref}
\oauthor{\bsnm{Wu}, \binits{Y.}},
\oauthor{\bsnm{Kirillov}, \binits{A.}},
\oauthor{\bsnm{Massa}, \binits{F.}},
\oauthor{\bsnm{Lo}, \binits{W.-Y.}},
\oauthor{\bsnm{Girshick}, \binits{R.}}:
Detectron2.
\url{https://github.com/facebookresearch/detectron2}
(2019)
\end{botherref}
\endbibitem

\bibitem[\protect\citeauthoryear{He et~al.}{2018}]{he2018maskrcnn}
\begin{botherref}
\oauthor{\bsnm{He}, \binits{K.}},
\oauthor{\bsnm{Gkioxari}, \binits{G.}},
\oauthor{\bsnm{Dollár}, \binits{P.}},
\oauthor{\bsnm{Girshick}, \binits{R.}}:
Mask R-CNN
(2018)
\end{botherref}
\endbibitem

\bibitem[\protect\citeauthoryear{Gupta et~al.}{2019}]{gupta2019lvis}
\begin{botherref}
\oauthor{\bsnm{Gupta}, \binits{A.}},
\oauthor{\bsnm{Dollár}, \binits{P.}},
\oauthor{\bsnm{Girshick}, \binits{R.}}:
LVIS: A Dataset for Large Vocabulary Instance Segmentation
(2019)
\end{botherref}
\endbibitem

\bibitem[\protect\citeauthoryear{Zhou et~al.}{2018}]{zhou2018ade20k}
\begin{botherref}
\oauthor{\bsnm{Zhou}, \binits{B.}},
\oauthor{\bsnm{Zhao}, \binits{H.}},
\oauthor{\bsnm{Puig}, \binits{X.}},
\oauthor{\bsnm{Xiao}, \binits{T.}},
\oauthor{\bsnm{Fidler}, \binits{S.}},
\oauthor{\bsnm{Barriuso}, \binits{A.}},
\oauthor{\bsnm{Torralba}, \binits{A.}}:
Semantic Understanding of Scenes through the ADE20K Dataset
(2018)
\end{botherref}
\endbibitem

\bibitem[\protect\citeauthoryear{Cordts et~al.}{2016}]{cordts2016cityscapes}
\begin{botherref}
\oauthor{\bsnm{Cordts}, \binits{M.}},
\oauthor{\bsnm{Omran}, \binits{M.}},
\oauthor{\bsnm{Ramos}, \binits{S.}},
\oauthor{\bsnm{Rehfeld}, \binits{T.}},
\oauthor{\bsnm{Enzweiler}, \binits{M.}},
\oauthor{\bsnm{Benenson}, \binits{R.}},
\oauthor{\bsnm{Franke}, \binits{U.}},
\oauthor{\bsnm{Roth}, \binits{S.}},
\oauthor{\bsnm{Schiele}, \binits{B.}}:
The Cityscapes Dataset for Semantic Urban Scene Understanding
(2016)
\end{botherref}
\endbibitem

\bibitem[\protect\citeauthoryear{Nathan~Silberman and Fergus}{2012}]{silberman2012nyu}
\begin{bchapter}
\bauthor{\bsnm{Nathan~Silberman}, \binits{P.K.} \bsuffix{Derek~Hoiem}},
\bauthor{\bsnm{Fergus}, \binits{R.}}:
\bctitle{Indoor segmentation and support inference from rgbd images}.
In: \bbtitle{ECCV}
(\byear{2012})
\end{bchapter}
\endbibitem

\bibitem[\protect\citeauthoryear{Long et~al.}{2015}]{long2015fcn}
\begin{botherref}
\oauthor{\bsnm{Long}, \binits{J.}},
\oauthor{\bsnm{Shelhamer}, \binits{E.}},
\oauthor{\bsnm{Darrell}, \binits{T.}}:
Fully Convolutional Networks for Semantic Segmentation
(2015)
\end{botherref}
\endbibitem

\bibitem[\protect\citeauthoryear{Laina et~al.}{2016}]{laina2016deep}
\begin{botherref}
\oauthor{\bsnm{Laina}, \binits{I.}},
\oauthor{\bsnm{Rupprecht}, \binits{C.}},
\oauthor{\bsnm{Belagiannis}, \binits{V.}},
\oauthor{\bsnm{Tombari}, \binits{F.}},
\oauthor{\bsnm{Navab}, \binits{N.}}:
Deeper Depth Prediction with Fully Convolutional Residual Networks
(2016)
\end{botherref}
\endbibitem

\bibitem[\protect\citeauthoryear{Barrett et~al.}{2023}]{barrett2023evolutionary}
\begin{barticle}
\bauthor{\bsnm{Barrett}, \binits{N.}},
\bauthor{\bsnm{Sadeghi}, \binits{Z.}},
\bauthor{\bsnm{Matwin}, \binits{S.}}:
\batitle{Evolutionary augmentation policy optimization for self-supervised learning}.
\bjtitle{Advances in Artificial Intelligence and Machine Learning}
\bvolume{03}(\bissue{02}),
\bfpage{1135}--\blpage{1164}
(\byear{2023})
\doiurl{10.54364/aaiml.2023.1167}
\end{barticle}
\endbibitem

\bibitem[\protect\citeauthoryear{He et~al.}{2021}]{he2021mae}
\begin{botherref}
\oauthor{\bsnm{He}, \binits{K.}},
\oauthor{\bsnm{Chen}, \binits{X.}},
\oauthor{\bsnm{Xie}, \binits{S.}},
\oauthor{\bsnm{Li}, \binits{Y.}},
\oauthor{\bsnm{Dollár}, \binits{P.}},
\oauthor{\bsnm{Girshick}, \binits{R.}}:
Masked Autoencoders Are Scalable Vision Learners
(2021)
\end{botherref}
\endbibitem

\bibitem[\protect\citeauthoryear{Xie et~al.}{2022}]{xie2022simmim}
\begin{botherref}
\oauthor{\bsnm{Xie}, \binits{Z.}},
\oauthor{\bsnm{Zhang}, \binits{Z.}},
\oauthor{\bsnm{Cao}, \binits{Y.}},
\oauthor{\bsnm{Lin}, \binits{Y.}},
\oauthor{\bsnm{Bao}, \binits{J.}},
\oauthor{\bsnm{Yao}, \binits{Z.}},
\oauthor{\bsnm{Dai}, \binits{Q.}},
\oauthor{\bsnm{Hu}, \binits{H.}}:
SimMIM: A Simple Framework for Masked Image Modeling
(2022)
\end{botherref}
\endbibitem

\end{thebibliography}

\end{document}